\documentclass[manuscript,screen]{acmart}

\PassOptionsToPackage{dvipsnames}{xcolor}
\usepackage{lscape}
\usepackage{longtable}
\usepackage[T1]{fontenc}
\usepackage[most]{tcolorbox}
\usepackage{lmodern} 
\usepackage{lipsum}
\usepackage{xcolor, soul}
\usepackage{tikz}
\usepackage{color, colortbl}
\usepackage{tablefootnote}
\usepackage{hhline}
\usepackage{microtype}
\usepackage{booktabs}
\usepackage{array}
\usepackage{hhline}
\usepackage{tabularray}

\AtBeginDocument{
  \providecommand\BibTeX{{
    \normalfont B\kern-0.5em{\scshape i\kern-0.25em b}\kern-0.8em\TeX}}}

\setcopyright{acmcopyright}
\copyrightyear{2025}
\acmYear{2025}
\acmDOI{XXXXXXX.XXXXXXX}

\begin{document}

\title[A Framework for Dynamic Situational Awareness in Human Robot Teams]{A Framework for Dynamic Situational Awareness in Human Robot Teams:\\An Interview Study}

\author{Hashini Senaratne}
\email{hashini.senaratne@csiro.au}
\orcid{0000-0001-5203-3793}
\author{Leimin Tian}
\email{leimin.tian@data61.csiro.au}
\orcid{0000-0001-8559-5610}
\author{Pavan Sikka}
\email{pavan.sikka@data61.csiro.au}
\orcid{0009-0006-4333-4838}
\author{Jason Williams}
\email{jason.williams@data61.csiro.au}
\orcid{0000-0002-2416-075X}
\author{David Howard}
\email{david.howard@data61.csiro.au}
\orcid{}
\affiliation{
  \institution{CSIRO Robotics, Data61}
  \streetaddress{}
  \city{Pullenvale}
  \state{QLD}
  \country{Australia}
  \postcode{4069}
}

\author{Dana Kuli\'c}
\email{dana.kulic@monash.edu}
\orcid{0000-0002-4169-2141}
\affiliation{
  \institution{Monash University and CSIRO Robotics, Data61}
  \streetaddress{}
  \city{Clayton}
  \state{Vic}
  \country{Australia}
  \postcode{3168}
}

\author{C\'ecile Paris}
\email{cecile.paris@data61.csiro.au}
\orcid{0000-0003-3816-0176}
\affiliation{
  \institution{Collaborative Intelligence Future Science Platform, CSIRO Data61}
  \streetaddress{}
  \state{NSW}
  \country{Australia}
  \postcode{}
}

\renewcommand{\shortauthors}{Senaratne, et al.}

\begin{abstract}
  In human-robot teams, human situational awareness is the operator's conscious knowledge of the team's states, actions, plans and their environment. Appropriate human situational awareness is critical to successful human-robot collaboration. 
  In human-robot teaming, it is often assumed that the best and required level of situational awareness is knowing everything at all times. This view is problematic, because what a human needs to know for optimal team performance varies given the dynamic environmental conditions, task context and roles and capabilities of team members. We explore this topic by interviewing 16 participants with active and repeated experience in diverse human-robot teaming applications. Based on analysis of these interviews, we derive a framework explaining the dynamic nature of required situational awareness in human-robot teaming. In addition, we identify a range of factors affecting the dynamic nature of required and actual levels of situational awareness (i.e., dynamic situational awareness), types of situational awareness inefficiencies resulting from gaps between actual and required situational awareness, and their main consequences. We also reveal various strategies, initiated by humans and robots, that assist in maintaining the required situational awareness. Our findings inform the implementation of accurate estimates of dynamic situational awareness and the design of user-adaptive human-robot interfaces. Therefore, this work contributes to the future design of more collaborative and effective human-robot teams.
\end{abstract}


\maketitle

\section{Introduction}

Human-robot collaboration aims to leverage the complementary capabilities of humans and robots in completing joint missions, where robots augment human capabilities rather than replace humans~\cite{MADDIKUNTA2022100257,ajoudani2018progress,bauer2008human,paris2021s}. Human-Robot Teaming (HRT) is particularly challenging in dynamic environments, where the environmental state and the goals of the shared mission may vary over time. Examples of HRT in dynamic environments include  search and rescue missions~\cite{arnold2018search}, manufacturing~\cite{cormier2015situational} and surgical~\cite{kunkes2022influence} applications. To effectively collaborate, team members must maintain \emph{Situational Awareness} (SA) of the evolving contexts~\cite{shayesteh2022enhanced,yonga2023research}. Endsley (1988) defined situational awareness as ``the perception of the elements in the environment within a volume of time and space, the comprehension of their meaning and a projection of their status in the near future''~\cite{endsley1988design}. Her model recognised SA as a distinct information processing stage and a precursor to decision-making and performance~\cite{endsley2000situation}.  Maintaining SA enables humans to successfully adapt to unanticipated dynamic situations and make informed decisions that improve team performance and safety~\cite{doi:10.1177/1541931218621034}. 

In HRT, SA has been viewed as an attribute of both humans and robots; i.e., humans' SA of robots and the environment as well as robots' ability to establish appropriate SA about humans, other robots, and the environment~\cite{yonga2023research}. In this work, we study the SA of humans about robots and the environment.

Human operators' SA in HRTs has been studied at different levels corresponding to Endsley's definition: (1) perception, (2) comprehension, and (3) projection \cite{sen2020effects,muller2023self}. Several works have highlighted the importance for human collaborators to maintain proper SA of robots and the environment~\cite{shayesteh2022enhanced,begum2015collaboration}. However, the required level of SA is not clearly defined in the literature. Many existing works assume that the required level of SA is constant throughout a mission and that the higher the SA of human collaborators, the more efficient and effective the mission outcomes would be~\cite{riley2010situation,jalaleddine2023evaluating,domova2020improving} (also see Section{~\ref{relatedwork:strategies}}). Some prior work provides context-specific examples of the dynamic nature of required SA, e.g., changing levels of autonomy requires changing levels of transparency, but this topic has yet to be studied in detail{~\cite{10.1145/3645090}}. While there is existing evidence for the dynamic nature of humans' actual SA~\cite{riley2010situation}, there is insufficient knowledge about what constitutes required SA in practical HRTs, factors that affect SA and strategies for maintaining SA.  

In this paper, we postulate that the level of SA required by a human collaborator \emph{to achieve optimal task performance} is dynamic, and depends on the environmental conditions, task context and the roles and capabilities of the other team members. For example, if the environment is easy and/or the other team members are capable, even very low levels of SA from a human operator will not detract from team performance. On the contrary, if the environment is challenging or the robot has a very low level of autonomy, a high level of human SA is required to achieve good team performance. Given that environment conditions and robot capability may change during a shared HRT mission, we hypothesise that the required optimal SA is dynamic.

As our first contribution, we propose a conceptual framework for dynamic situational awareness in HRTs. It posits that inefficiencies in human SA, namely SA losses, SA latencies, SA inaccuracies, incomplete SA and excess SA (as defined in Section~\ref{sec:results_DSA}), occur due to gaps between the required and actual levels of SA. These SA inefficiencies contribute to missed actions, delayed actions, unnecessary actions, faulty actions, confusion and unnecessary overload, affecting HRT performance. We developed the dynamic SA framework following an analysis of 16 semi-structured interviews with people who had active and repeated experience in diverse HRT applications. Secondly, we present a list of factors influencing the dynamic nature of actual and required levels of SA, as well as evidence for SA inefficiencies, based on examples from diverse HRT contexts. Thirdly, we identify a range of strategies, operator- or human-robot interface-initiated, for maintaining the required SA within dynamic human-robot collaborative missions. Finally, this paper concludes with a discussion of how the proposed framework and strategies can inform the future design of more collaborative and effective HRTs. We emphasise the need of accurate estimates of dynamic SA and user-adaptive interfaces.

\section{Related Work}\label{sec:bg}

Theoretical models of SA have been developed for human-human teams or human-computer interfaces more broadly~\cite{STANTON2001189}. Recently, some of these models have been adapted to the HRT context. This section briefly reviews these models and analyses their suitability for identifying SA requirements within HRT applications. Further, this section reviews current approaches for estimating and improving SA within HRTs, while identifying challenges of these existing approaches given the hypothesised dynamic nature of SA requirements.

\subsection{Models of Situational Awareness for Human-Robot Teaming}

\subsubsection{Individual Situational Awareness}

Endsley's 3-level model of SA~\cite{endsley1988situation} was first introduced as a property held by a human about the world. Recently, it has been adopted in the context of HRT, considering SA as a property held by a human \cite{riley2010situation, senaratne2023roman} or a robot \cite{fischer2007shaping,muller2023self}. 
For example, human SA was used to analyse an operator's ability to promptly perceive objects detected by robots and comprehend non-optimal robot actions through a remote interface \cite{senaratne2023roman}. In another example, human SA was adopted to describe the person's ability to project future locations of robots and predict task progress \cite{riley2010situation}. 
Leveraging Endsley's SA model and other related theories, Chen et al.{~\cite{chen2014situation}} developed the Situation Awareness-based agent transparency (SAT) model for understanding an operator's SA of mission environments involving a robot-like agent. The SAT model describes three levels of information that agents need to convey about their decision-making process to facilitate the operator's perception, comprehension and projection required to perform effective human-agent teamwork. These levels include the agent's current goals and actions, reasoning process and predictions along with uncertainty. 

Beyond Endsley's model, another representative theory that defines SA as a property of an individual is the theory of activity approach~\cite{bedny1999theory}. It emphasises three stages of activity: orientational (developing an internal representation of the world or current situation), executive (proceeding towards a desired goal via decision-making and action execution), and evaluative (assessing the situation via information feedback, influencing the executive and orientational stages)~\cite{bedny1999theory,salmon2008really}.  
In addition, \citet{smith1995situation} proposed the perceptual cycle model, which views SA as a generative process of knowledge creation and informed action-taking, where internally held schemata direct one's interactions with the world, and the outcome of interaction modifies the original schemata~\cite{smith1995situation,salmon2008really}. 

There have been limited application of the activity approach model and the perceptual cycle model in HRT. When applied, they have been used to define humans' SA~\cite{lynch2023maritime}. Further, the activity approach model and the perceptual cycle model focus on SA processes. Endsley's model focuses on SA constructs, while the SAT model focuses on information useful for SA levels defined by Endsley. Therefore, in this paper, we follow Endsley's 3-level SA model when defining humans' SA within HRT settings.

\subsubsection{Team Situational Awareness}

Beyond defining SA as an individual's mental state, SA theories have also evolved to describe a team-level property~\cite{MANSIKKA2021103473}, introducing the concept of \textit{team SA}. Team SA comprises SA of individual members (individual SA as described above), SA that overlaps between members (i.e., shared SA) and the combined SA of the whole team (i.e., distributed SA). 
Both shared SA and distributed SA have been theorised for HRTs~\cite{ADRIAENSEN2022103320}, with an emphasis on the importance of communication and coordination between humans and robots~\cite{schuster2011measurement}. Team SA offers new approaches to improve team performance through appropriate SA overlaps and distribution among members. Its application in HRT contexts, however, is still emerging. 
Understanding SA requirements at the individual level is necessary before investigating appropriate team SA within dynamic HRT settings. Thus, we focus on individual SA in this work.

\subsubsection{Gaps in Existing Situational Awareness Models for HRT}

Given the dynamic nature of real-world HRT, the SA of an individual may change or be rendered obsolete~\cite{salmon2008representing, riley2010situation}. 
However, existing SA models do not contain a sufficiently detailed representation of these SA dynamics ~\cite{salmon2008really}. 

Furthermore, current models do not adequately distinguish between the \textbf{required} level of SA and the actual SA attained by an individual within dynamic HRT contexts. An individual does not need to know everything about every team member and all aspects of the environment at a given time. The amount of SA required depends on the context, roles, and capabilities of the rest of the team. An individual may not need to perceive, comprehend or predict much at all in some scenarios, e.g., when a robot is operating autonomously and successfully accomplishing its task~\cite{9981165}.
Also, SA requirements of different individuals within a team may differ significantly at a given time. The SAT model addresses this concept to a certain extent by providing context-specific examples; e.g., the operator may only need to know the agent's proposed actions and the projected outcome to make a sufficiently informed decision in a time-sensitive situation{~\cite{chen2014situation}}. Chen et al. also argued that information provided to humans must be dynamic and sensitive to the operator's role{~\cite{Chen2018}}. Nonetheless, there is a lack of individual SA models which can capture the fluctuation in required SA considering varying HRT contexts and task progression, as well as the consequences of such fluctuations{~\cite{salmon2008really}}. Thus, there is a need to understand the required SA of a human operator during a dynamic mission to derive useful models that improve collaboration in HRTs.

Due to the above gaps, current approaches for measuring and improving SA in HRTs are limited, specifically in terms of quantifying and maintaining dynamically varying levels of SA.

\subsection{Approaches for Measuring Situational Awareness in Human-Robot Teams}

There are four main approaches to assess human's SA in HRT: direct, performance, behavioural measures, and process indices, which we review in this section. The majority of SA measures assume that a constant level of SA is desirable and the best SA is knowing everything~\cite{riley2010situation}. 

\subsubsection{Direct Measures}
Direct measures use questions focused on the three SA levels (perception, comprehension and projection) as defined in Ensley's individual SA model~\cite{endsley1988situation}. The Situational Awareness Rating Technique (SART)~\cite{hopko2021effect,dini2017measurement} and Situation Awareness Global Assessment Technique (SAGAT)~\cite{dini2017measurement,endsley1988situation} are two widely administered direct SA measures. SART is a self-rated post-trial subjective measure. It uses human responses to a series of questions associated with attentional demands, attentional supply, and understanding to calculate SA. 
SAGAT uses a freeze probe technique (freezing the simulation at randomly selected times), so the human can be queried during the trial. Their answers are then compared with the ground truth answers from the simulator to collect data corresponding to the three SA levels. 
Neither measure captures fine-grained fluctuations of actual SA: SART suffers heavily from recall and overgeneralisation biases~\cite{endsley2000situation} and is affected by individual differences. 
SART compares an individual's SA against what an individual perceives as maximum or minimum SA based on their past experiences. What is needed in HRT is to compare an individual's SA against the SA level required for the considered scenario. SAGAT addresses this drawback to a certain extent, by querying information that the human should know at a given timepoint and measuring SA at different times during the mission. However, the freeze probe technique necessary for SAGAT is not suitable for real-world scenarios that cannot be interrupted randomly. Further, queries used could also cue the subject to alter their attention for the rest of the mission~\cite{endsley2000situation}. 

\subsubsection{Performance Measures}
Performance measures of SA include global performance measures and embedded task measures. An example global performance measure is the disparity between the final number of target objects found and the number of objects that should have been found in a search and rescue mission~\cite{kanyok2022novel}. An example embedded task measure is the completion times and the number of mistakes associated with each executed task~\cite{tanke2013improving} . Global performance measures suffer from diagnosticity and sensitivity biases, as they only capture the end result of a long string of cognitive processes~\cite{endsley1995measurement}.
Comparatively, embedded task measures work better, as those are more frequent. In selecting embedded task measures, it is crucial to ensure that they rely heavily on SA, but less on other confounds such as lack of information, poor sampling strategies or improper data integration~\cite{endsley2000situation}. 

\subsubsection{Behavioural Measures}
Behavioural measures of SA are often observer-rated. Situation Awareness Behaviorally Anchored Rating Scale (SABARS) is an example measure applied in contexts similar to HRTs (e.g., military and aviation simulations) ~\cite{doi:10.1037/h0095998}. It involves an expert evaluating operator behaviours for consistency with SA by rating a series of questions 
during or subsequent to observing an operator. SABARS can be biased when observers only have access to the operator's behaviours (e.g., actions on the interface and verbalisations), as the operator's actions may not reflect things that are not known due to attentional narrowing~\cite{endsley1995measurement}. This measure outputs a single SA estimate over the entire mission trajectory. Therefore, it does not capture the dynamics of SA. Another example of behavioural measures involves comparing critical or off-nominal incidents noted by observers of robots in the field with behaviours of the operator noted by observers at a control room~\cite{doi:10.1177/0278364916688254}. This approach heavily relies on observers' expertise. However, it enables comparisons between operator actions and observers' knowledge about what is really happening to identify SA failures~\cite{drury2007lassoing,doi:10.1177/0278364916688254}, reflecting instances of gaps in (actual vs. required) SA. Given its benefits, some recent work has attempted to improve this approach with the use of process indices ~\cite{drury2007lassoing,senaratne2023roman}.

\subsubsection{Process Indices}
Process indices of SA are the measures that reflect the processes used for attaining SA. An example is assessing eye movements on remote visual displays, which provide information related to an operator's attention allocation ~\cite{ratwani2010single,dini2017measurement}. Recent work also uses process indices in conjunction with behavioural measures. Examples include analysing the operator's verbalisations collected through a think-aloud protocol (a process index), which are later verified using videotapes of the robot in the field and the operator's interface, and collisions or off-nominal robot actions noted by an observer during the run~\cite{drury2007lassoing,senaratne2023roman}. Related work defines a more fine-grained measure of SA, namely SA latency, defined as the delay between the time the robot requires human assistance and the time the human identifies that need~\cite{senaratne2023roman}. This is a measurement of the SA gap between actual and required SA. The same work also correlates two types of SA latency (associated with perception and comprehension levels) with eye movement metrics collected during a remote HRT experiment to achieve more objective and continuous SA estimates. 

Currently, SA latency is the closest match to the requirement of defining and assessing SA at a fine-grained level to suit the dynamic nature of HRT missions. However, a framework for systematic analysis of required SA levels during dynamic HRT missions is still missing, which can inspire developing useful SA measures. 

\subsection{Current Approaches to Improve SA in HRT}\label{relatedwork:strategies}

Existing work to enhance SA within HRT contexts mainly focuses on providing additional information to the human. Examples include providing extra information about robot states and goals using augmented reality~\cite{tanke2013improving, huuskonen2019augmented}, improving the remote environment visualisation through wide-angle cameras on robots~\cite{sumigray2021improving}, and including virtual annotations (i.e., additional text, sign, image, or video based information displayed on a user interface) to assist the operator in avoiding safety critical incidents when controlling robots under adverse visual conditions~\cite{hong2020effect}. Other studies focus on providing already available information through alternative modalities such as haptics~\cite{fink2023expanded} and mixed reality~\cite{green2008human, allenspach2023design} to reduce the chances of missing important information. Another approach is allowing the operator to manipulate the level of robot autonomy
~\cite{kidwell2012adaptable} considering multiple factors, including the operator's level of trust, workload, fatigue and confidence about the robot's capacity to handle certain tasks or environments. Despite their benefits, these approaches tend to increase operator cognitive load. 
There have also been alternative approaches which focus on reducing the operator's cognitive load. For example, \citet{pitman2007picture} modified the user interfaces to allow the operator to see images from both the ground robot’s camera and from the overhead camera simultaneously as a single image (i.e., picture-in-picture format), rather than in split views.

Overall, these approaches focus on increasing SA rather than assisting humans to achieve required SA levels. To provide such assistance, interfaces may need to be adaptive to the context. For example, when the actual SA levels drop below the required SA, the interface assistance needs to focus on improving SA by directing the user's attention to important pieces of information. Different types of assistance may be required when the actual SA levels are above required SA.  A few studies have designed and evaluated adaptive interfaces with the aim of improving SA, but without paying much attention to the concept of required SA. Examples include modifying the level of automation as a function of participants’ performance~\cite{kidwell2012adaptable} and providing information discovered as relevant from the mission data through the interface~\cite{roldan2018adaptive}. Informed by an improved understanding of required SA in dynamic HRT contexts, future work can improve such adaptive interface designs and implement effective assistance.

\section{Methods}

As reviewed in Section~\ref{sec:bg}, an individual's required SA in dynamic HRT is 
yet largely uncharted in current research. Therefore, to further understand SA  within dynamic HRT contexts, we conducted interviews with human operators in diverse HRTs. Specifically, based on their prior and ongoing experience, our study aimed to explore: (1) how the required and actual human SA level fluctuates within an HRT mission, (2) which factors contribute to the misalignment between the required and actual SA levels and the consequences of such misalignment, and (3) insights for designing HRT interfaces to estimate such misalignment and to assist human in reaching the required SA levels. 

We conducted semi-structured interviews with human members of HRTs to allow us to capture participants' reflections from multiple instances of using the same HRT interface over the long-term in different conditions. We invited participants engaged in different HRT applications representing remote, distant and co-located teams and varying degrees of time  and safety criticality, aiming to identify generalisable as well as domain-specific considerations for SA in HRTs.

The inclusion criteria for participation were: (a) human member 
of an HRT that works towards a shared mission, with interleaved or overlapped tasks distributed between humans and robots, with interdependencies between those tasks, and (b) having repeated experience in working within the associated HRT team. 
Given criterion (a), the selected HRT applications represented collaborative relationships between humans and robots, i.e., they have a shared understanding of the mission and context, they work interdependently and adapt to each other over a sustained period to meet the shared mission \cite{schleiger2023collaborative}. Given criterion (b), in answering our questions, one participant could provide several examples collated from multiple instances of engaging with the same HRT application. Given both these criteria, we excluded those who had experience with HRTs that do not represent collaborative relationships, e.g., restaurant delivery robot applications. 

\subsection{Participants}

Sixteen participants took part in the study (male = 15, female = 1; age: [M = 34.59 (7.10)]; years of experience with their HRT application: [M = 3.27 (2.70)]). See Table 8 in Appendix B for a detailed summary of participants' demographics and HRT experience.
With regard to the self-identified expertise level in associated HRT applications, 7 identified themselves as expert users, 5 as proficient users and 4 as competent users, according to Benner’s novice to expert model~\cite{thomas2017benner}. None of the participants identified themselves as a novice or advanced beginner. Notably, 14 participants were involved in robotics research or engineering-related professions, one participant was a medical practitioner and one was an artist.

\subsection{Human-Robot Teaming Applications}

Table~\ref{tab:context} categorises the HRT applications of our participants in terms of application domains, physical co-presence, time and safety criticality, realistic nature, and number of humans and robots in the team.

\begin{table}[h!]
    \centering \small
    \setlength{\tabcolsep}{4pt}
    \begin{tabular}{lcccccccccc}
        \toprule
        \rowcolor{gray!10!white}
        {\textbf{Application}} &  \textbf{Parti-} & \textbf{Physical} & \textbf{Time} & \textbf{Safety} & \textbf{Realistic} & \textbf{\# of} & \textbf{\# of}\\
        \rowcolor{gray!10!white}
         \textbf{Domain}
         & \textbf{cipants} & \textbf{Co-presence} & \textbf{Criticality} & \textbf{Criticality} & \textbf{Nature} & \textbf{Humans} & \textbf{Robots}\\
 \midrule
        Disaster response &  &  &  & & & & \\
        \hspace{10pt}search \& rescue (S\&R1) 
        & P1-3, P5  & Remote & High  & High & Realistic & 1 & 1-6 \\
        \hspace{10pt}search \& rescue (S\&R2) 
        & P12-13 & Remote & High  & High & Simulated & 1 & 1 \\
        \hspace{10pt}nuclear accident (NuAcc)
        & P12-13 & Remote & High  & High & Realistic & 2 & 2 \\

        Agriculture &  &  &   &  & & &\\
        \hspace{10pt}fruit inspection (FruInsp)
        & P1-3, P5 & Remote & Low & Low & Realistic & 1 & 1-2\\
        \hspace{10pt}fruit harvesting (FruHar)
        & P4, P6 & Distant & Low & Low & Real & 1-2 & 1\\

        Crafting (Craft)
        & P7-8 & Distant & Low & Low & Semi-realistic* & 1 & 1\\

        Mine inspection (MineInsp)
        & P9-11 & Distant,  & Low & Low & Real & 1 & 1\\
        & & Remote & & & &  & \\

        Manufacturing &  &  &   & & & & \\
        \hspace{10pt}aircraft parts (AirManf) 
        & P14 & Co-located & Low & High & Real & 1 & 1\\
        \hspace{10pt}large 3D art (ArtManf)
        & P16 & Co-located & Low & High & Real & 1 & 1\\

        Medical: surgery (Surg)
        & P15 & Co-located & High & High & Semi-realistic\textsuperscript{+} & 1 & 1\\

        \bottomrule
    \end{tabular}
    \caption{Summary of Human-Robot Teaming Applications of The Participants}
    \footnotesize{* Crafting HRT is marked as semi-realistic because the robot did not have the ability to identify which crafting materials were needed, and an experimenter made this decision. \textsuperscript{+} Surgery HRT is marked semi-realistic because surgeons used it to practice laparoscopy surgery training skills using a physical simulator, but not on real patients.}
    \label{tab:context}
    \vspace{-8pt}

\end{table}

Physical co-presence identifies whether the operator and robots are co-located (in direct line of sight of each other and sharing the same physical space), distant (in direct line of sight of each other but not sharing space), or remote (neither seeing each other directly nor sharing the same space).  Time and safety criticality of applications are categorised as high or low, considering the cost of delays and severity of damage that failures could cause. For example, in search and rescue applications, if the robots are not guided promptly, delays in finding casualties can put lives at risk (i.e., high time criticality); and in manufacturing applications where humans are co-located with high payload carrying robots, failures in robots could pose a risk to humans' safety (i.e., high safety criticality). The realistic nature is categorised as ``real'' when robots are operating in the physical target environment (e.g., actual fruit orchards); ``realistic'' when operating in a physical world closely approximating the target environment (e.g., a tunnel with manually placed signs of casualties); ``semi-realistic'' when operating in a physical simulator or a physical world that uses Wizard of Oz prototyping{~\cite{riek2012wizard}}; or ``simulated'' when a virtual representation of the physical world is used. 

In terms of the HRT tasks, P1-3 and P5 used a remote visual interface and a joystick to provide high- to low-level guidance to single to multiple robots. The HRT goal was to explore physically made-up disaster (S\&R1) or farm (FruInsp) sites and locate signs of casualties or fruits, respectively. During search and rescue missions, the operator had to strategically drop WiFi nodes carried by robots on the field to maintain the robots' connectivity to the remote station. P12-13 used a similar visual interface and joystick in a simulated search and rescue application with mixed action initiatives and variable robot autonomy (S\&R2). In this HRT, the operator had the authority to switch between three autonomy levels (pure teleoperation, pure autonomy and semi-autonomy), and the robot could also decide when to change the autonomy level based on task performance metrics. In another HRT task, P12-13 used a visual interface integrated with physical buttons to remotely navigate two robots assigned to two operators in a physically made-up nuclear disaster site (NuAcc). The main robot's goal was to locate and dispose of a contaminated piece while assisted by a secondary view provided by the other robot. P4-6 represented an apple harvesting application (FruHar). They could drive and position a distant robot closer to the canopy using a joystick. Then during the robot's apple-picking, they can override its decisions on which apples to pick and the order of picking via a visual interface. P9-11 operated a semi-autonomous drone in distant and remote settings using a joystick and tablet-based controls to create 3D models of mines (MineInsp). P7-8 were human collaborators of a robot who handed over materials to people to perform crafting activities in a Wizard of Oz experimental setting replicating a shared art studio (Craft). P14 shared jobs with a co-located robot performing processes such as drilling and sanding to manufacture aircraft parts (AirManf). P16, an artist, was involved in programming a robot and collaborating with it to perform large-scale 3D artwork (ArtManf). P15, a trainee surgeon, used a physical simulator replicating the abdomen of a patient to practice laparoscopic surgery activities, assisted by a robot arm that provided an auto-zoomed-in view of the surgery area (Surg). P15 could override the robot's actions through a foot interface. Table 9-18 and Figures 10-19 in Appendix B provide detailed descriptions and illustrations of these HRT applications.

\subsection{Data Collection}

Calls for participation were emailed to Australian research and industrial groups identified to have HRT applications. We invited the volunteers who met the inclusion criteria to participate, and asked them to assist us in identifying other potential participants through snowballing. 

The principal researcher conducted 16 face-to-face interviews over four months in compliance with institutional ethical requirements. We conducted eight interviews in person, and eight as remote video interviews. Interviews were 1.5 hours long and conducted in a semi-structured manner, allowing follow-up questions to improve the clarity of responses. We audio-recorded all the interviews. We encouraged the participants to use images, videos or actual demonstrations of their user interfaces or HRT application to explain certain examples clearly during or after the interviews.

Prior to the interviews, we requested participants to fill out an online questionnaire collecting demographics and background details of their HRT experience (see Appendix A.1)

To guide the interviews, we explained to the participants the concept of SA using Endsley's SA model~\cite{endsley1988design} and asked questions covering four main areas: (1) their HRT context, (2) challenges and failures they face due to gaps in SA, (3) how the required SA levels fluctuate during an HRT mission and which strategies are used to maintain SA at suitable levels, and (4) how the user interface could be improved to support maintaining required levels of SA (see Appendix A.2). The questions used to explore challenges and failures due to SA gaps were motivated by the gulf of execution concept and the gulf of evaluation concept in Norman's stages-of-action model~\cite{norman1986cognitive}.

\subsection{Data Analysis}

The audio-recorded interviews were fully transcribed. We analysed these transcriptions using a thematic analysis method similar to the one suggested by Nowell et al.~\cite{nowell2017thematic}. This method is recognised as useful for examining various perspectives, capturing similarities and differences between interviews and generating unanticipated insights. A set of initial codes were generated deductively based on the interview questions motivated by the gulf of execution and the gulf of evaluation concepts{~\cite{norman1986cognitive}} (e.g., delayed actions, missed actions, unnecessary actions, misinterpretations/ confusions). As our familiarisation with the data improved, we also performed inductive coding that resulted in novel themes and codes in addition to the initial deductive codes. These themes contributed to the main outcome of this paper: the framework detailed in Section {\ref{sec:results_DSA}}. The principal researcher performed the analysis, while other authors engaged in several iterations of code verification and revisions during weekly team meetings. 

\section{Results}\label{sec:results}

\subsection{Framework for Dynamic Situational Awareness in Human-Robot Teaming}\label{sec:results_DSA}

\begin{figure}[b!]
  \centering
  \includegraphics[width=0.95\linewidth]{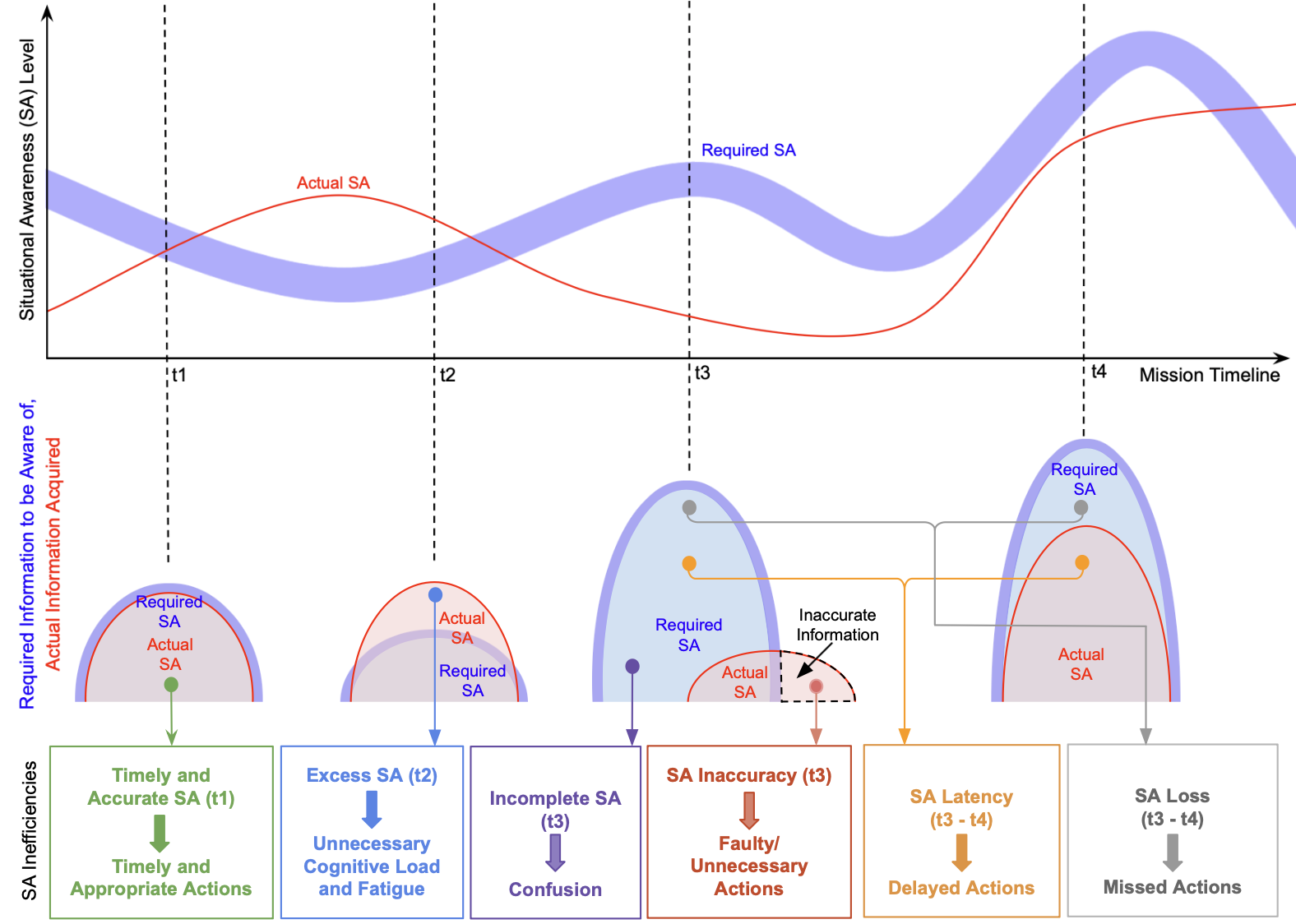}
  \caption{Proposed framework demonstrating the gaps between required and actual situational awareness (SA) within dynamic mission contexts}
  \footnotesize{Note: Dots in same color (grey and orange dots) represent the same information at two different time points: t3 and t4.}
  \label{fig:DSAFramework}
\end{figure}

Informed by the interview data, we developed a framework that synthesises the concept of dynamic and required human SA for maximising task performance within an HRT mission (see Figure \ref{fig:DSAFramework}). The proposed framework captures the notion that the required SA can vary during the mission. For example, during critical periods, high levels of SA are needed, while during more routine periods, lower levels of SA are needed. Further, the human operator's actual SA can deviate from the required SA and is also dynamic. 

In Figure {\ref{fig:DSAFramework}}, the trajectory of the actual SA represents the SA level experienced by an individual during an HRT mission. The required SA is expressed as a band to illustrate that optimal team performance can be achieved by maintaining actual SA within a range of SA levels (i.e., one who becomes aware of some required information a little earlier or later than a given time point could still finish the mission on time). 

As shown in Figure~\ref{fig:DSAFramework}, we define timely and accurate SA when there is an alignment between actual and required SA (t1). When the actual SA at a given time does not align with the required SA, it can result in excess SA, incomplete SA, SA inaccuracy, SA latency, or SA loss.
Actual SA can be higher than the required SA, causing excess SA (t2). In this case, there may not be an immediate negative impact on the task performance. However, excess SA may increase the human's cognitive load unnecessarily, which can result in cognitive fatigue over a longer term or inability of the human to perform additional tasks. 
Incomplete SA is being partially aware of information related to HRT mission (i.e., required SA is higher than actual SA, as shown at t3). Incomplete SA can lead to an operator being unable to explain why the robot behaves in certain ways or to confusion, leading to inability to identify required actions. 
Actual SA may also contain inaccurate information due to an operator misinterpreting the data, resulting in SA inaccuracy (t3). Due to SA inaccuracies, the operator may perform faulty or unnecessary actions, such as inappropriate intervention into a robot's autonomous actions. 
SA latency is the delay between the time the robot member(s) of the HRT require the human member's assistance and the time the human identifies that need~\cite{senaratne2023roman}. As shown in Figure~\ref{fig:DSAFramework}, at t3 the actual SA does not yet contain information required to perform the action, until t4 when the actual SA has increased sufficiently (orange dots). SA latency results in delayed decisions, which often lead to delayed actions. 
SA loss is the failure to identify that the robot(s) requires the human member's assistance. As shown in Figure~\ref{fig:DSAFramework}, the actual SA does not contain information required to perform the action at either t3 or t4 (grey dots), and thus the operator may fail to perform the required actions due to this SA loss.

As shown in our framework, excess SA, incomplete SA, SA inaccuracy, SA latency and SA loss can deteriorate the human-robot collaboration and mission performance. Therefore, attempts should be made to minimise the gap between required SA and actual SA at a given time point. 

\begin{figure}[h!]
  \centering
  \includegraphics[width=0.6\linewidth]{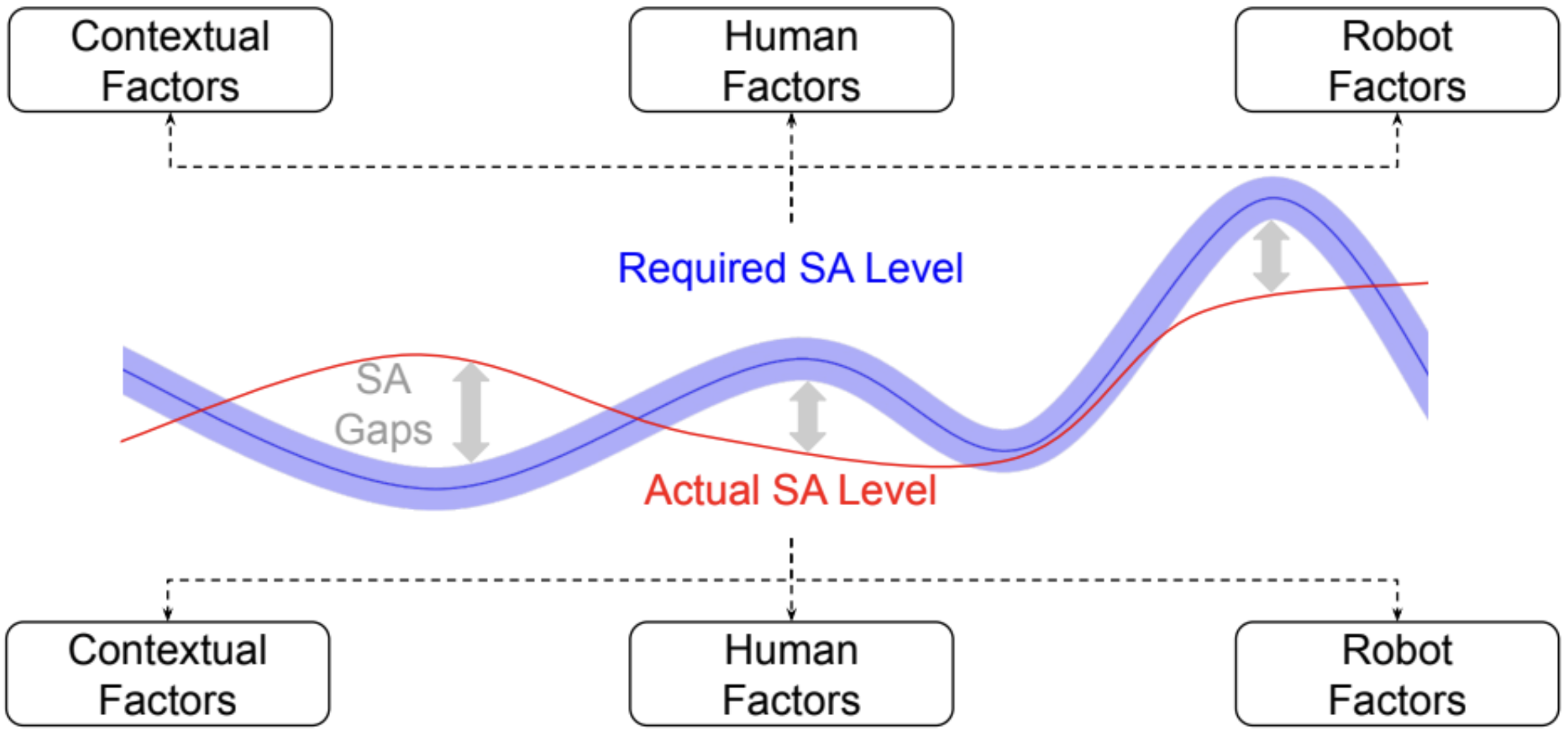}
  \caption{Contextual, human and robot-related factors affect the dynamic nature of required and actual situational awareness (SA), contributing to dynamic situational awareness gaps}
  \label{fig:Factors}
  \vspace{-10pt}

\end{figure}

Our interview analysis provided evidence for the dynamic nature of both the required and actual SA levels within HRT missions, revealing three types of factors contributing to these fluctuations: contextual, human and robot factors (see Figure{~\ref{fig:Factors}}). 
Furthermore, our participants reported multiple anecdotes of delayed, missed, unnecessary and faulty actions as well as confusion resulting from the gaps between required vs. actual SA. Finally, we found several operator- and interface-initiated strategies to support maintaining actual SA levels closer to required SA levels within dynamic HRT missions (see Table~\ref{tab:strategies}). These findings are presented below.

\begin{table}[h!]
\centering\small
\begin{tabular}{
>{\raggedright}p{.12\linewidth}
>{\raggedright}p{.38\linewidth}
>{\raggedright}p{.4\linewidth}
}
\toprule
&\textbf{Human-Initiated}
&\textbf{Interface-Initiated}
\tabularnewline
\midrule
Currently Used
&
Frequent visual scans to update big-picture awareness\par\smallskip
Manage workload demands through prioritisation  or slowing robots down\par
&
Bring selected information to human’s attention through audio alerts
\tabularnewline
Suggested for Future Use
&
Check robot and task status using natural language queries without changing visual attention\par\smallskip
Request explanations about robot actions and status to reach common ground\par
&
Predict incidents requiring  human intervention and bring priority information to human’s attention\par\smallskip
Track user’s attention to assist in dynamically updating big-picture awareness\par\smallskip
Track user’s state to modify robot behaviours or autonomy
\tabularnewline
\bottomrule
\end{tabular}
\caption{Adaptive strategies to maintain alignment between required and actual SA in dynamic HRT missions}\label{tab:strategies}
  \vspace{-10pt}

\end{table}

\subsection{Evidence of Dynamic Required Situational Awareness Level}\label{sec:requiredSA}

All participants (16/16) agreed that the SA levels required during HRT missions vary with time (Interview Question 3a, Appendix A.2). An increase in required SA was also found to increase the likelihood of SA gaps and vice versa. Given that our participants represented HRT applications with co-located, distant and remote settings, single to multiple robots, high to low time and safety criticality, this finding supports the idea that required SA levels are dynamic within diverse HRT contexts. Participants articulated that required SA is dependent on several factors: task context, human and robot-related factors, and that these factors change over the short- and/ or long-term. Below we summarise each of these factors, in the order of strongest to weakest evidence. Refer to Tables 19 and 20 in Appendix C for anecdotes reported by the participants detailing how these factors affect the SA requirements within different HRC settings.

\subsubsection{\textbf{Contextual Factors Influencing Required Situational Awareness}\nopunct}\label{sec:contexual_required}
\hfill 

\begin{figure}[h]
  \centering
  \includegraphics[width=\linewidth]{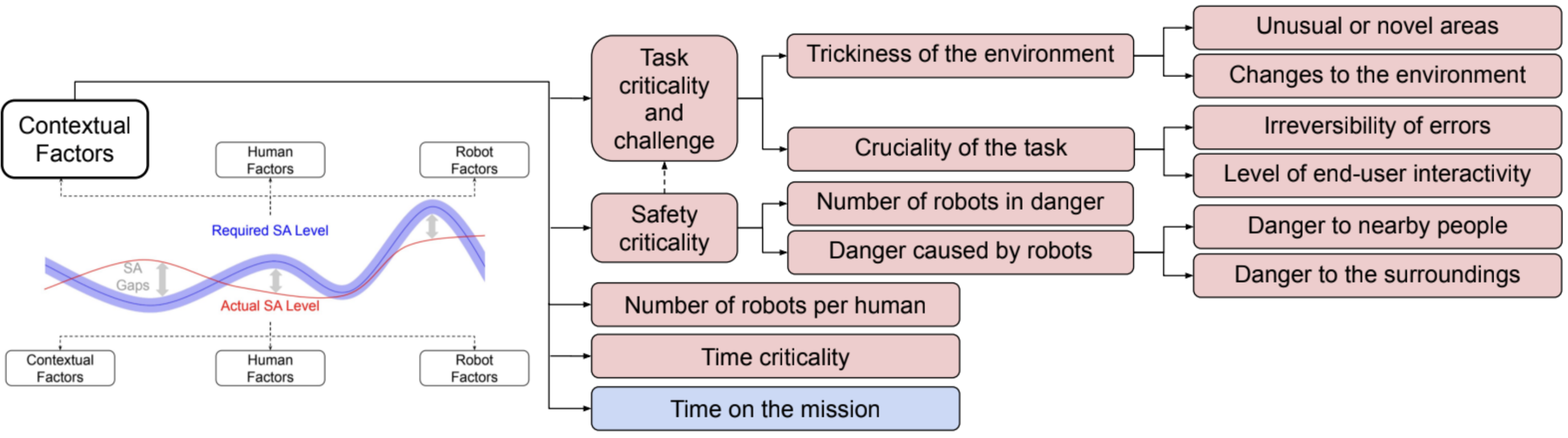}
  \caption{Contextual factors affecting the \textit{required} SA. Factors in red contribute to an increase in required SA when their value increases, while the factor in blue contributes to a decrease in required SA when its value increases. The dashed arrow illustrates that ``safety criticality'' depends on ``task criticality and challenge''.}
  \label{fig:Contexual-RequiredSA}
\end{figure}

Our participants referred to 5 main contextual factors affecting the required SA level, as illustrated in Figure{~\ref{fig:Contexual-RequiredSA}}: task criticality and challenge (14/16), safety criticality of tasks (9/16), the number of robots per human (8/16), time criticality of the mission (5/16), and time spent on the mission (5/16). All participants agreed whether a contextual factor contributed to an increase (factors in red) or decrease (factor in blue) in required SA.

\textit{\textbf{Task criticality and challenge:}} The majority of our participants related the task criticality to the \textit{trickiness of the environment} (11/16). When the environment is ``demanding and challenging'', the ``risk of failure is high'', and humans may need to frequently attend to and intervene robots in a timely manner (P10). Two types of environment trickiness emerged from our interview data: unusual or novel areas to the robots (P1-2, P4-5, P9-10, P12, P15) and dynamic changes that the robots are not programmed to handle (P6, P11). Required SA is low when robots act in simple and known environments  (P2-4, P6). 

A few participants also related this factor to \textit{how crucial a particular task is to the mission} (4/16), considering irreversible errors (P14-15) and level of end-user interactivity associated with the tasks (P7-8). In addition, it was identified that when critical tasks with high required SA are performed in a ``repetitive'' (P15, P16) manner, SA demands decrease over time; following the ``same pattern <means> the situation is less unpredictable and <it requires> less mental resources''.

\textit{\textbf{Safety criticality of tasks:}} 
Safety critical scenarios, i.e., situations where the robots are in danger (P2-3, P5-6, P10) or causing danger to the surrounding environment (P1, P6, P15-16) or nearby people (P2, P9, P16), increase the levels of required SA (9/16). In these scenarios, human collaborators need to closely monitor the robots to quickly recognise those instances and override robot actions to mitigate any damage. The timely detection and avoidance of dangers related to one robot can ``take up a lot of mental space'', but at the same time, they also need to ``keep an eye on'' their own tasks and other robots so as not to miss anything (P1). The likelihood of safety criticality is high when the task criticality and challenge are high. However, we list them as two factors given that not all tricky or crucial tasks are safety critical; e.g., a robot trying to extract an apple from a location where there is nothing after a change to the canopy structure (P6), and a robot milling at a slightly different coordinate of a large 3D art (P16).

\textit{\textbf{The number of robots per human:}} The number of robots per human was identified to positively correlate with SA demands (8/16). Participants in multi-robot HRTs (P3, P5, P12-13), as well as those with plans to monitor multiple robots (P4, P6, P14, P16) contributed to this factor. 
When providing further insights for the type of SA to maintain when the number of robots is high, some highlighted that humans need to comprehend an ``abstract'' (P5) or a ``high level'' (P3) SA of the mission rather than trying to ``comprehend detailed SA on each robot'' (P5), which is not feasible when a human supervises many robots.

\textit{\textbf{Time criticality of the mission:}} Time criticality affects the required SA level (6/16). Participants who worked both in the disaster response and agriculture domains compared the two domains, claiming that because the search and rescue mission ``is time constrained'', they ``have to maintain higher levels of SA'' compared to the fruit inspection mission where there are ``allowances for delays'' (P1-P3, P5). The aircraft manufacturing mission was also identified to be less time critical, therefore, robots are designed to ``wait for human input'' (P14).

\textit{\textbf{Time spent on the mission:}} An increase in time spent on the mission was reported to decrease the required SA level (5/16). 
Higher SA demands usually occur  ``at the beginning of a mission'', given that the humans ``have less information'' about the context and they need to ``figure out what is going on'' and ``make predictions'' (P4-5, P8). In contrast, ``towards the end of the mission'', required SA is low, as they have developed a good understanding of what to expect (P7, P8).

\subsubsection{\textbf{Robot-related Factors Influencing Required Situational Awareness}\nopunct}\label{sec:robot_required}
\hfill 

\begin{figure}[h]
  \centering
  \includegraphics[width=\linewidth]{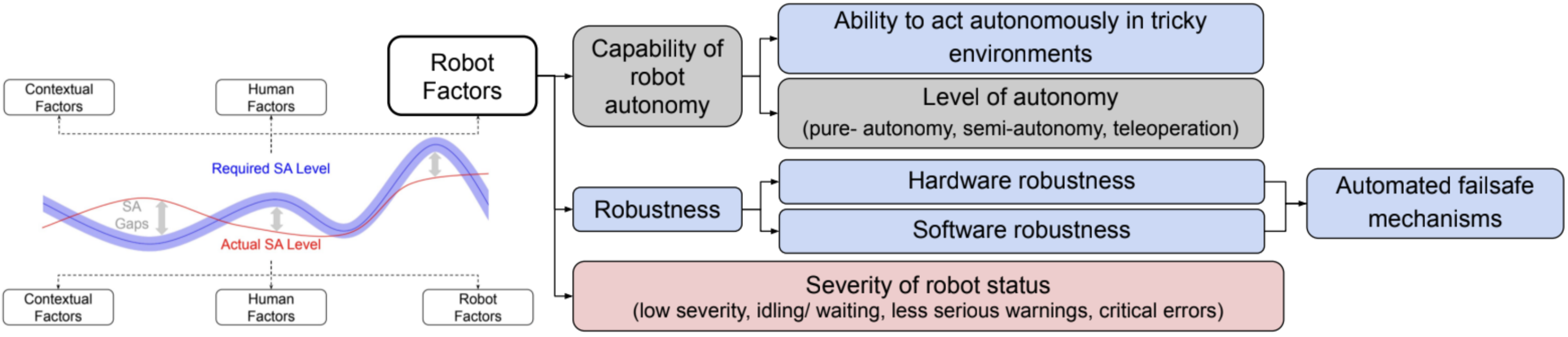}
  \caption{Robot-related factors affecting the \textit{required} SA. The factor in red increases the required SA, while the factors in blue decrease the required SA. Factors in grey indicate mixed trends.}
  \label{fig:Robotic-RequiredSA}
      \vspace{-10pt}
\end{figure}

Three robot-related factors were found to affect the required levels of SA (see Figure{~\ref{fig:Robotic-RequiredSA}}): the capability of robot autonomy (15/16), software and hardware robustness (10/16), and the severity level of the robot’s status (7/16). For factors shown in blue in Figure{~\ref{fig:Robotic-RequiredSA}}, all participants who discussed them agreed that an increase in the factor value decreased the required SA; for the factor in red, all participants who discussed it agreed that an increase in its value increased the required SA. Regarding the capability of robot autonomy, we obtained mixed results depending on whether the HRT is a mixed-initiative variable autonomy application.

\textit{\textbf{Capability of robot autonomy:}} Robot capability may or may not be fixed during the mission. In addition to relating how the human's SA demands increase when robots are not capable of independently acting in tricky environments (as detailed under ``Task criticality and challenge'' in Section~\ref{sec:contexual_required}) (11/16), our participants related how SA demands vary depending on the robot's level of autonomy (10/16). Some participants related robot capability to stages of product research and development process, where ``early stages'' represent lower capability ``than in the end product'' (3/16). 

In general, lower autonomy levels increased SA demands (P4-5, P9, P11-13, P15). The examples we received mostly related to HRTs where the human was in control of switching the autonomy level (see Tables 9, 10, 12, 13, 15 and 18 in Appendix B for details on distinct autonomy levels in disaster response, agriculture, mining, and surgical applications). In contrast, in a ``mixed-initiative variable-autonomy'' application, 
sometimes the opposite trend was observed (P12-13). There, the human needs to additionally perceive when robots switch autonomy levels and evaluate whether or not its chosen autonomy level is appropriate. Furthermore, when ``the operator does not agree with the robot's decision'', the human needs to ``fight for control by spending time back and forth switching between the teleoperation and autonomy'' (P13). 
This additional SA demand is associated with the reliability of the robot's decision. Thus, it can also relate to Software Robustness (see below). In addition, it can be related to an additional factor, Locus of Control (i.e., who decides when robot capability levels should be changed){~\cite{chiou2021trust}}, which we do not report separately due to lack of evidence. 

\textit{\textbf{Software and hardware robustness:}} SA demands change with the robustness of robots (10/16). Our participants felt the need for less SA with ``the robots consistently achieving their goals'' (P5) and when ``the autonomy itself has some failsafes'' (P3, P10-11). In contrast, when there are errors resulting from software or hardware robustness issues (e.g., localisation errors due to issues with SLAM (P1, P13), failing to park the robot at the correct time due to wear and tear of wheels (P4)), humans need to put extra effort into comprehending errors and identifying workarounds to recover and reduce the undesirable impacts (P7, P11).

\textit{\textbf{Severity level of robot status:}}
When the ``severity level'' (P9) of the robot's status is accessible to the human collaborator through the interface (directly or indirectly), it is considered a factor that contributes to the fluctuating SA demands (7/16). The expectation is for human operators to prioritise attending to any ``critical errors'' or warnings that will cause immediate bad outcomes to the mission (e.g., damages or risks to the robot, users or environment) over ``less-serious errors'' or warnings that may not cause an immediate problem (P9-11). When a robot is idling or waiting for human input, there is no demand to immediately attend to the robots (P2-3, P14, P16), compared to attending to robots that are reporting an error status.

\subsubsection{\textbf{Human-related Factors Influencing Required Situational Awareness}\nopunct}
\hfill

\begin{figure}[h]
  \centering
  \includegraphics[width=\linewidth]{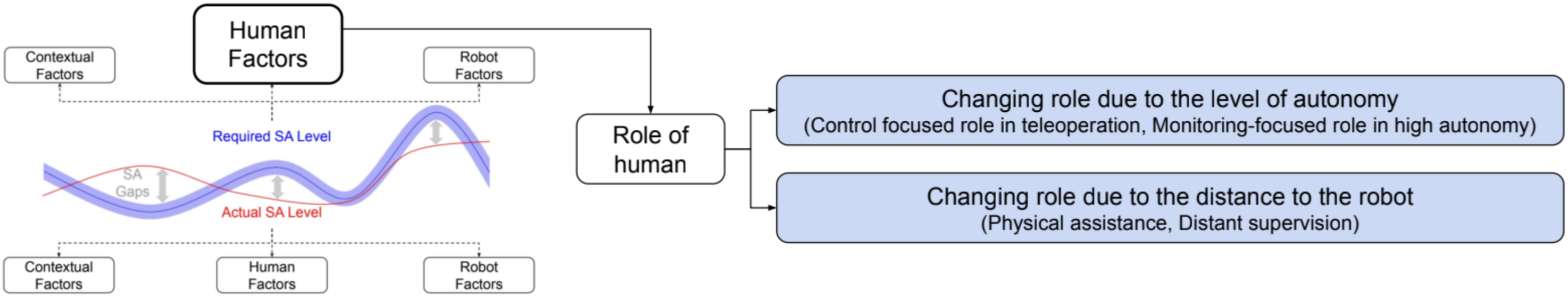}
  \caption{Human factors affecting the \textit{required} SA. Both factors in blue decrease the required SA. The factor in white is not quantifiable, thus cannot be associated with increased or decreased required SA.}
  \label{fig:Human-RequiredSA}
      \vspace{-10pt}

\end{figure}

The ``human role'' was found to affect the required SA (8/16), as detailed below and shown in Figure{~\ref{fig:Human-RequiredSA}}.

\textit{\textbf{Human role:}} Depending on the chosen level of autonomy (7/16) and proximity to the robot (1/16), the operator's role changes within an HRT application, requiring different levels of SA (8/16). In variable autonomy applications that allow the human to  choose the level of robot autonomy, the SA demands often decreased as robots reached higher levels of autonomy, given that humans' roles changed from control-focused roles to monitoring-focused roles (7/16).
Similarly, as the human's distance to the robot increases, SA demands decrease, given that the human role changes from active physical assistance of the robot to distant supervision of unexpected behaviours of robots (P14).

\subsection{Evidence of Dynamic Actual Situational Awareness Level}\label{sec:actualSA}

All participants (16/16) agreed that the actual SA levels experienced during HRT missions vary over time relative to required SA levels (Interview Questions 2 and 3e, Appendix A.2), regardless of the HRT application. Actual SA was reported to depend on contextual, human and robot-related factors, which modulate over time, both in the short- and/or long-term. Below we summarise each of these factors, in the order of strongest to weakest evidence. Please see Tables 21 to 23 in the Appendix C for the full list of examples provided by our participants supporting each factor.

\subsubsection{\textbf{Human-related Factors Influencing Actual Situational Awareness}\nopunct}\label{sec:actualSA_human}
\hfill 

\begin{figure}[h]
  \centering
  \includegraphics[width=\linewidth]{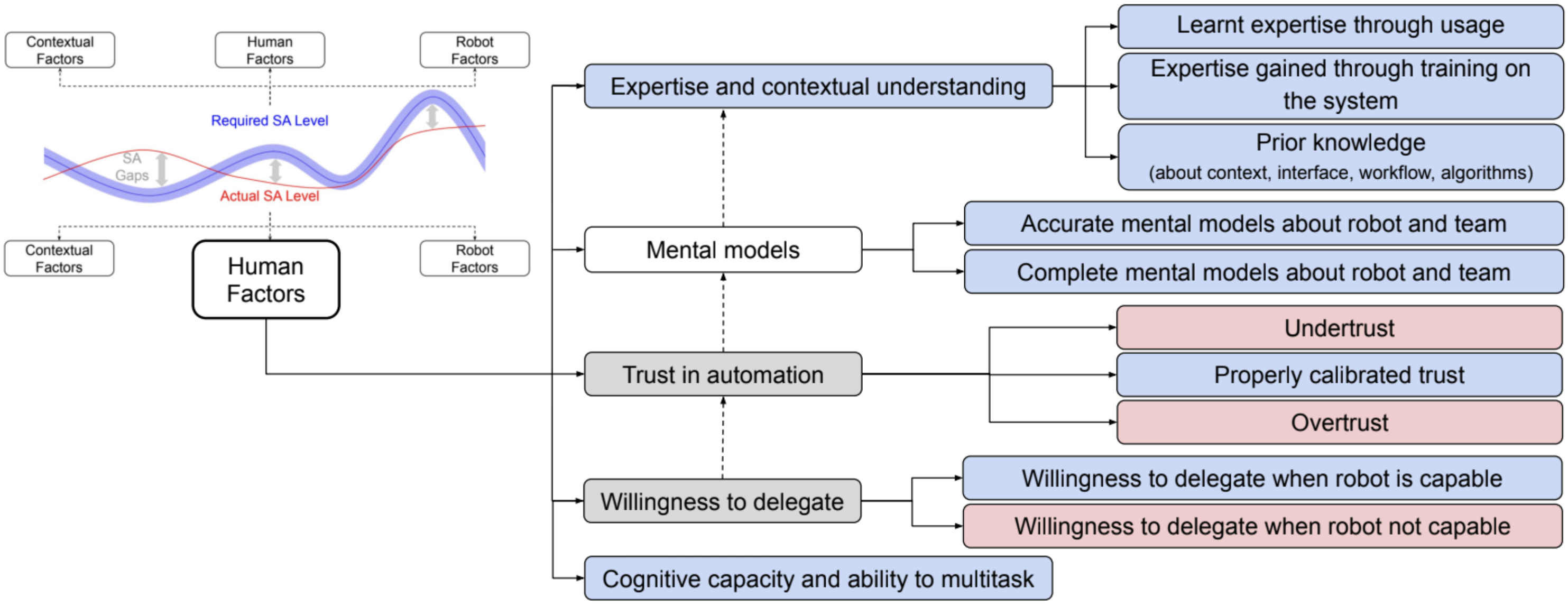}
  \caption{Human factors affecting \textit{actual} SA. Factors in red decrease the actual SA, while factors in blue increase the actual SA. Dashed arrows illustrate the dependencies between factors. The factor in white is not quantifiable, thus cannot be associated with increased or decreased actual SA. Factors in grey indicate mixed trends in their components.}
  \label{fig:Human-ActualSA}
\end{figure}

Five human factors were found to affect the actual levels of SA: expertise and contextual understanding (11/16), mental models about robots (8/16), trust in robot automation (7/16),  cognitive capacity and ability to multitask (6/16) and willingness to delegate (3/16). See Figure{~\ref{fig:Human-ActualSA}}.
Participants agreed on how these factors or their components contribute to gaps between actual and required SA.

\textit{\textbf{Expertise and contextual understanding:}}  Several examples were shared to support the notion that expertise and/or contextual understanding based on prior experience or knowledge assists humans in maintaining actual SA at required levels (11/16). Expertise and/or contextual understanding gained through learning over time (P3, P12, P8), ``training on the system'' (P9), and prior knowledge about the ``workflow'' (P15), ``interface'' (P10) or ``context'' (P2, P10, P12 P15), assisted our participants in ``faster perceiving'' (P12), comprehending the need and correct times to intervene (P2-3, P8) and making correct predictions about robots' actions (P3, P15-16). The expertise and contextual understanding factor also relates to the mental models, trust in automation, and willingness to delegate, as shown with the dashed arrows in Figure{~\ref{fig:Human-ActualSA}}, which we further discuss below. 

\textit{\textbf{Mental models about robots:}} Mental models of robots' behaviours and capabilities, based on diverse experiences, knowledge about underlying decision-making processes of robots or through assumptions, were identified as a contributor to changes in actual SA in the long term (8/16). Our participants provided examples indicating that when humans have accurate and complete mental models, they are able to proactively identify situations where the robots will not perform well, resulting in actual SA closer to the required SA levels and proactive assistance to the robot team (P5, P9). In contrast, incorrect mental models have led to actual SA falling below the required SA, demonstrated by ``failures to intervene at the right times, intervening too early, or not intervening at all when they need to'' (P5). Our participants related these instances to ``a clash between <operators'> mental model <and the reality>'' (P5) or difficulties to develop mental models about robots' actions (P6, P9, P14), robots' decisions (P1, P9), type of robot autonomy (P8) or skills and capabilities of robots or human-robot team (P16). Additional examples were shared by participants who related mental models to trust (see examples under ``Trust in robots' automation'').

\textit{\textbf{Trust in robot automation:}} The human operators' level of trust in robots varies (7/16) depending on various factors, such as their mental models about robots (P3),  ``prior experience'' (P13) and ``understanding of how the robots work'' (P6), making them sometimes under or overtrust the robots. Undertrust often leads to cognitive overload by paying ``close attention'' to unnecessary elements (P16). It can make humans attempt to ``babysit robots unnecessarily'' (P3), resulting in attention tunnelling (i.e., allocation of attention to a particular channel of information or a task for a duration that is longer than optimal) {\cite{wickens2009attentional}}, decreasing their actual SA on the ``big picture of the mission'' (P13). Overtrust may result in ``disregarding what is happening'' and missing aspects that should be noticed (P16) because of ``overconfidence'' (P2, P16) in robots' ability, ``wrong assumptions'' (P2) and ``insufficient situational awareness''  (P16). These changes in actual SA create gaps with the required SA. Proper calibration of trust given the robot capabilities was identified to support maintaining actual SA at required levels, enabling humans to ``make correct predictions and decisions'' (P5).

\textit{\textbf{Cognitive capacity and ability to multitask:}}  Some of our participants supported the idea that actual SA is limited by the operator's cognitive capacity, making it difficult to reach the required SA at a higher level (6/16). This is because during ``multitasking, part of <their> attention is divided'' (P13) and it ``takes away the focus from <one of the tasks>'' (P15), causing them to ``miss the cue from the robot, or what is happening'' (P13). P4, P12 and P13 further shared some strategies that they practice when it is challenging to achieve the required SA given a human's cognitive capacity, which are detailed in Section~\ref{sec:human-initiated-current}.

\textit{\textbf{Willingness to delegate:}} When the robot's autonomy is capable, but the operator is unwilling to delegate tasks to autonomy due to various factors (``lack of trust'' (P5), personal traits (P1, P13), context-based ``preferences'' such as preferring to be relaxed (P13)), they experience a higher cognitive load unnecessarily by trying to be aware of everything (3/16). They experience higher actual SA than required SA (excess SA).

\subsubsection{\textbf{Contextual Factors Influencing Actual Situational Awareness}\nopunct}\label{sec:actualSA_context}
\hfill 

\begin{figure}[h]
  \centering
  \includegraphics[width=\linewidth]{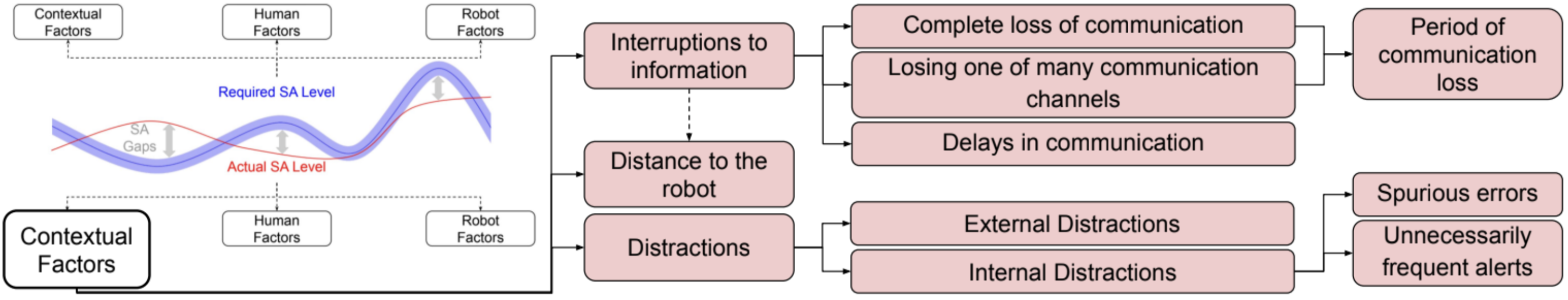}
  \caption{Contextual factors affecting the \textit{actual} SA. All factors in red decrease the actual SA. The dashed arrow illustrates the dependency between factors.}
  \label{fig:Contexual-ActualSA}
\end{figure}

Three contextual factors were found to affect the actual SA levels (see Figure{~\ref{fig:Contexual-ActualSA}}): interruptions to information (8/16), distractions (4/16), and distance to the robot (3/16). Participants agreed that an increase in each of these factors decreases the actual SA, resulting in a widened gap to the required SA.

\textit{\textbf{Interruptions to information:}} Unreliable and high-latency communication with robots that occurs in remote or distant HRTs is reported to decrease actual SA (8/16), causing SA latency until the communication is restored, incomplete SA  or SA loss (refer to definitions in Section{~\ref{sec:results_DSA}}).
Our participants provided examples of short- and long-term interruptions to communication between humans and robots (P1, P9-10) causing SA loss, where humans needed to rely on automated recovery behaviours (P1, P9) or work on bridging or restoring communication (P5, P7-8). P10 referred to the resulting SA loss as: ``my SA of what is occurring with the robot is null and void''. In the event of  experiencing a communication``degradation'' with the robot (P2) or only losing one information channel among many others (P7), operators often experienced delayed or limited access to information about the robot, creating challenges to make decisions because of incomplete SA. In contrast, the operator's SA about robots ``was much better'' when the remote robots were ``within communication range and have steady communication <with the human>'' (P2).

\textit{\textbf{Distractions}} A few participants pointed out that distractions arise due to external factors (e.g., external noises), as well as internal factors (e.g., unnecessarily frequent alerts or spurious errors of the HRT system), contributing to reduced actual SA about the HRT mission (6/16). These distractions took away humans' attention away from what they should be doing (P7, P12), resulting in important information going unnoticed, making it ``difficult to concentrate'' or making them ``overwhelmed'', ``stressed'' or ``confused'' (P9-10, P12).

\textit{\textbf{Distance to the robot:}} A few participants, who worked with robots at different distances within the same HRT application, related how ``proximity to the actual robot'' (P14) negatively correlates with actual SA (3/16). When the robot and the collaborator were co-located, the collaborator usually had high actual SA. When they were at a distance from each other but within the line of sight, although humans' actual SA was lower than the co-located scenario, it was better compared to when they were remote to robots.

\subsubsection{\textbf{Robot-related Factors Influencing Actual Situational Awareness}\nopunct}\label{sec:actualSA_robot}
\hfill 

\begin{figure}[h]
  \centering
  \includegraphics[width=\linewidth]{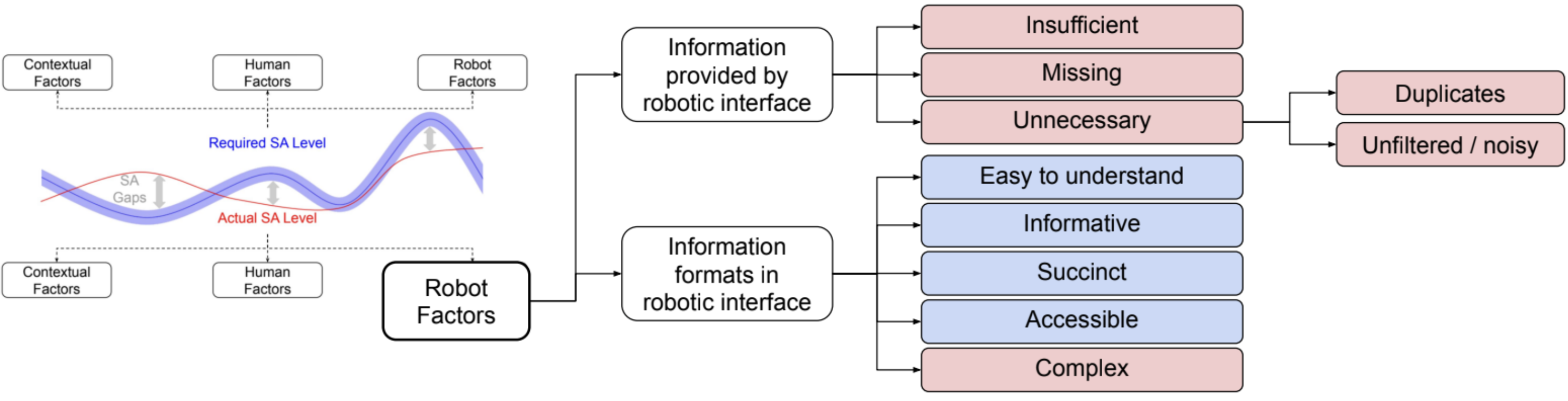}
  \caption{Robot-related factors affecting the \textit{actual} SA. Factors in red decrease actual SA. Factors in blue increase actual SA. Factors in white are not quantifiable, thus cannot be associated with increased or decreased actual SA.}
  \label{fig:Robotic-ActualSA}
\end{figure}

Two robot-related factors were found to affect the actual levels of SA compared to the required SA: information provided by the robotic interface (9/16) and the formats of how information is presented in the robotic interface (8/16), as shown in Figure{~\ref{fig:Robotic-ActualSA}}. Participants agreed that insufficient, missing, or unnecessary information provided by the interface decreases the actual SA. They also reported how better information presentation format increased the actual SA.

\textbf{\textit{Information provided by the robotic interface:}} In addition to limitations to information availability arising from unreliable and high-latency communication channels (detailed above as a contextual factor affecting actual SA), information availability can be impacted by robotic interface design, affecting actual SA (13/16). The lack of required information available through the robotic interface limits the actual SA of human operators (9/16). This ``limited perception that is coming back to <human collaborators>'' results in ``limited SA'' (P3). Participants provided various examples of missing (P2, P6, P8, P14-15, P16) and insufficient (P2, P11, P13) information related to interface design (P11, P13-14, P16), sensor availability (P2, P15) and sensor position and orientation (P6, P8). Unnecessary information provided by the ``busy'' interfaces or ``duplicate'' interface components increases the ``mental effort <of operators> to understand what's going on'' (P1), lowering their actual SA compared to required SA levels (8/16). 

\textit{\textbf{Information formats in the robotic interface:}}
Participants described several inefficient information presentations in robotics interfaces that result in decreased actual SA (8/16). For instance, ``<not providing> the most succinct presentation of information'' (P1), difficult-to-comprehend visualisations and textual messages (P1-2, P9, P12), less informative notifications (P10-11), and inaccessible language (P8). When using complex interfaces, human need to put extra effort in perceiving and comprehending software or hardware controls, as well as analysing information from robots. This additional mental load increases the complexity of collaborating with robots. In contrast, when the information is presented in easy to understand formats (factors in blue in Figure~\ref{fig:Robotic-ActualSA}), it increases human's actual SA.

\subsection{Evidence Indicating the Gaps between Required and Actual Situational Awareness}\label{sec:SAgap}

Given the dynamic required and actual SA levels, gaps where the actual SA is higher or lower than required SA levels can occur. These gaps contribute to consequences such as delayed actions (e.g., ``inability to intervene at the right times'' (P5)), missed actions (e.g., ``not intervening at all, when need to'' (P5)),  unnecessary actions (e.g., ``intervene too early or without needing to'' (P5)), faulty actions, or confusion (leading to difficulties in deciding whether to intervene or not), affecting the human-robot team performance,  described as the SA inefficiencies in Figure{~\ref{fig:DSAFramework}}. This section presents selected examples from our interviews highlighting SA gaps and their outcomes, referring to the human, robot, and contextual factors that influence a human's actual SA as discussed in Section{~\ref{sec:actualSA}}. Specifically, we found delayed actions due to SA latency to be the most prevalent in our interviews, followed by missed actions due to SA loss, faulty actions due to inaccurate SA, unnecessary actions due to inaccurate SA, confusion due to incomplete SA, and unnecessary load and fatigue due to excess SA.

\subsubsection{\textbf{Delayed Actions Due to SA Latency}\nopunct}

Many participants (11/16) reported delayed actions caused mainly by SA latencies arising from various factors. The examples shared are categorised as latencies in perception SA, comprehension SA and prediction SA, as shown in Table{~\ref{tab:delayedActions}}.

\begin{table}[h!]
    \centering\small
    \begin{tabular}{|p{.13\linewidth}|p{.39\linewidth}|p{.41\linewidth}|}
        \hline
        \textbf{Type of SA}  & \textbf{Example delayed action} & \textbf{Reason for SA Gap} \\
        \textbf{latency} & & \\
 \hline
        Perception & Late to review objects sent by robots (P1-2) & Tunnel vision ``on a robot'' or ``<another> panel''\\ 
        \hhline{~|-|-}
        
        & Late to ``pick up a visual warning'' indicating low battery (P9) & The visual attention was elsewhere\\ 
        \hhline{~|-|-}
                
        & Late to assist a ``robot waiting <for human to> change a sandpaper'' (P16) & Delays in noticing a pop-up alert while attending another task \\ 

        \hline

        Comprehension & Late to ``park the robot in a position'', so ``there was not enough workspace to do path planning and pick the apples'' (P4) & Delays to comprehend that ``uneven ground'' and ``wear and tear'' condition of robot's wheels affect braking speed\\
        \hhline{~|-|-}
        
        & Late to ``take over'' the drone's control as soon as it ``lost track of where it was'' (P11) & Delays in understanding that a SLAM drift has happened\\
        \hline
        
        Perception and comprehension & ``Slower'' responses to signals from the crafter requesting to hand over some material (P7-8) & A lag in a video stream from the robot, causing delays in receiving signals and making it challenging to ``decide the best time to do the handover''\\

        \hhline{~|-|-}

        & Late to guide ``two robots trying to cross paths at a <narrow> tunnel'' (P5) & Late to perceive cues and comprehend the need to intervene due to activity on another screen\\
        \hhline{~|-|-}
        
        \hline
        
        Prediction & Late at dropping the communication nodes, while the nodes ``needed to drop back where there is good communication'' (P3) & Delayed predictions about when to drop a node associated with a lack of ``experience''\\

        \hhline{~|-|-}
        
        & Late to ``bring a <drone> back to a safe landing spot'' after noticing that the drone was running out of battery (P9) & An inaccurate mental model assumed that operators could always ``rely on robot's fail-safes,'' which delayed arriving at correct prediction\\

        \hline
    \end{tabular}
    \caption{Examples of delayed action due to perception, comprehension and prediction SA latencies} 
    \label{tab:delayedActions}
    \vspace{-10pt}
\end{table}

As discussed in Section{~\ref{sec:actualSA_human}}, human-related factors can decrease actual SA, such as lack of ability to multitask (causing tunnel vision) and distractions (making warnings or important cues go unnoticed). Moreover, as detailed in Section{~\ref{sec:actualSA_robot}}, robot-related factors can decrease actual SA, such as poor information formats (e.g., use of unimodal alerts). Both contributed to perception SA latency. In addition, human-related factors, such as lack of expertise and contextual understanding and inaccurate mental models, as well as contextual factors that reduce actual SA (see Section{~\ref{sec:actualSA_context}}), such as interruptions to information, were among reasons behind comprehension and prediction SA latencies.

\subsubsection{\textbf{Missed Actions Due to SA Loss}\nopunct}

Our participants identified that missed actions mainly resulted from missing cues from the robot. This can often occur due to human factors, namely multitasking causing attention division and high workload (P1, P5, P13), inexperience (P7-9), and lack of prior knowledge (P6), as well as the contextual factor of information unavailability (P1, P8, P15) (9/16). Participants shared examples of missed actions caused by a loss in perception, comprehension and prediction SA (see Table{~\ref{tab:missedActions}}). Similar to SA latency, perception SA losses are mainly due to the contextual factor of information unavailability decreasing the actual SA (examples shared by P1, P8 and P15 in Table{~\ref{tab:missedActions}}).

\begin{table}[h!]
    \centering\small
    \begin{tabular}{|p{.13\linewidth}|p{.42\linewidth}|p{.38\linewidth}|}
        \hline
        \textbf{Type of SA loss} & \textbf{Example missed action} & \textbf{Reason for SA Gap} \\
 \hline
         Perception & Failing to take action to avoid the drone ``running into an object'', when operators had to ``disable the collision avoidance'' mode, as the drone was struggling to move ``because of dust''(P9) & Failure to perceive objects as different from dust \\
         \hhline{~|-|-}
    
         & Not being able to perceive robots' needs and guide them as operators were ``pushed back into a regime that takes tools <to maintain required levels of SA> away from'' them (P1) & Loss of communication with robots for extended periods\\
         \hhline{~|-|-}

         & Failing to perceive a crafter's signal; therefore to handover requested materials (P8) & Obstructed camera view \\
         \hhline{~|-|-}
        
         & Failing to ``avoid a collision'' between the robot-operated camera and the surgical instruments that a trainee surgeon was holding (P15) & No sensors to view ``where the camera shaft was in the surgical workspace''\\

        \hline

         Comprehension & Failing to comprehend ``very hard to recover from'' terrain conditions and avoid robots operating in autonomous mode ``going into a train track'' and ``exploring a concrete ditch pit'' (P1, P5)  & Operator ``attention was <mostly> on a different agent'' ``for too long''\\
         \hhline{~|-|-}
        
         & Failing to comprehend and respond to crafters' signals  during initial usage (P7, P8)& ``Not being familiar with the robot controls and spending too much time on  controls'' or ``the window showed the video stream <of the user> is pretty small''\\
        
        \hline
        
         Prediction & Failing to filter out certain apples before the robotic gripper failing to ``extract <those> apples from the canopy'' due to ``cable entanglement'' or ``too much slippage'' (P6) & Failure to predict those incidents, knowing the robot's limitations\\
        
        \hline
    \end{tabular}
    \caption{Examples of missed action due to the loss in perception, comprehension and prediction SA} 
    \label{tab:missedActions}
    \vspace{-8pt}
\end{table}

\subsubsection{\textbf{Faulty Actions due to Inaccurate SA}\nopunct}

Due to SA inaccuracies in perception, comprehension and prediction, our participants reported performing faulty actions (11/16). See examples in Table {\ref{tab:faultyActions}}. These faulty actions have contributed to actively impeding task performance. Many of these  relate to failing to accurately perceive or comprehend the status of the robot or its environment (P1-6, P9, P12, P14), sometimes leading to inaccurate predictions. These SA inaccuracies stemmed from reduced actual SA caused by both human factors of lack of expertise and contextual knowledge, and the contextual factor of information unavailability.

\begin{table}[h!]
    \centering\small
    \begin{tabular}{|p{.13\linewidth}|p{.38\linewidth}|p{.42\linewidth}|}
        \hline
        \textbf{Type of SA} & \textbf{Example faulty action} & \textbf{Reason for SA Gap} \\
        \textbf{inaccuracy} & & \\
 \hline
         Perception & ``Multiple <unhelpful> movements with the robot'' when guiding it to open the door of the nuclear disaster site (P12) & Incorrect depth perception \\
        
        \hline

        Comprehension &``Doing things the robot did not expect'', disrupting the manufacturing process (P14) & Misinterpreting the robot's state of ``planning'' as ``broken'', as the robot was not moving \\
         \hhline{~|-|-}
        
        & ``Deciding to land the drone in a place where it is not viable either to retrieve the aircraft or can be dangerous to land (P9),'' and selecting poor locations to drop the communication nodes (e.g., ``over a hill, far too low to the ground level, in water''), resulting in ``breaking the comm backbone'' (P3) & Inaccurate comprehension of the robot's ``underlying terrain'' \\
        \hhline{~|-|-}

        & ``Taking over control and force <a robot> up a <rocky> slope'', when the robot ``was quite slow'', which made ``the robot roll over and breaks the drone on its back''. (P1-3, P5) & Inaccurate understanding about robot's terrain due to difficulty ``to gauge <it> from the graphical user interface'' 
         \\
         \hhline{~|-|-}

        & Repeatedly trying to assist a robot to ``go through a narrow doorway'' (P5) & Inaccurately comprehending that robot is not stuck, while ``the robot was caught up on parts of the doorway that prevented it moving forward''\\
        
        \hline

         Comprehension and potential prediction & ``Getting a robot back to <a previously opened> door, and found that it was closed (P3) 
        & Inaccurately assuming that dynamic obstacles would not present in that environment, therefore, not paying enough attention to``the point cloud data'' and just relying on their memory 
\\
        \hhline{~|-|-}

         &``Stopping the robot halfway, and later on realising that ``probably 30\% of the time, it would be okay for the robot to continue'' and their intervention was not necessary or timely (P4, P6) & Failure to comprehend the robot's environment correctly; e.g., whether ``the apples <were> in a cluster <or not>''; leading to failures in accurately predicting whether the ``gripper would knock down the adjacent apples'', the ``wires along the gripper could get tangled'', or the ``gripper could be damaged because of twisting a branch''  \\
        \hhline{~|-|-}

                 &Using an ``incorrect waypoint choice'' (a planar waypoint instead of a 2D waypoint), which made the drone go backwards, given that the provided ``planar waypoint intersects the drive at the back'' (P10) & Incorrect comprehension of the robot's environment, resulting in inaccurate predictions about how the robot would act\\

         \hline
        Prediction &Making the robot hand over materials to the crafters at specific locations, which made ``the arm collide with an object on the table'' (P7) & Not accurately predicting whether the trajectory of the arm overlaps with those objects \\

        \hline
    \end{tabular}
    \caption{Examples of faulty action due to perception, comprehension and prediction SA inaccuracies} 
    \label{tab:faultyActions}
    \vspace{-14pt}
\end{table}

A participant also shared an instance of faulty action due to SA loss: failing to perceive that their milling robot's tip was feeding; therefore, they went to do a tool change, which resulted in potentially dangerous sparks (P16).

\subsubsection{\textbf{Unnecessary Actions due to Inaccurate SA}\nopunct}

\begin{table}[h!]
    \centering\small
    \begin{tabular}{|p{.13\linewidth}|p{.38\linewidth}|p{.42\linewidth}|}
        \hline
        \textbf{Type of SA} & \textbf{Example unnecessary action} & \textbf{Reason for SA Gap} \\
        \textbf{inaccuracy} & & \\
 \hline
         
        Comprehension & 
        ``Taking control of the drone in situations'', when it would have been perfectly fine for it to progress on its own (P9, P11)&  Not correctly understanding what the drone was doing and what decisions it was making, due to lack of experience with the system 
         \\
        \hhline{~|-|-}

        &``Making the robot hand over <some crafting material> before the crafter signalled'' (P7) & Incorrectly comprehending a gesture 
        as signalling a handover\\
        \hhline{~|-|-}

        \hline

         Comprehension and potential prediction & During ``early <usage stages>, be very hands-on'' and frequently overriding a lot of the autonomy unnecessarily (P1, P5) 
         & Assuming that the robot was not capable or ``<having> a very clear view in mind about'' how exactly the robot should proceed with a task\\
         \hhline{~|-|-}
         
        \hline
    \end{tabular}
    \caption{Examples of unnecessary action due to perception, comprehension and prediction SA inaccuracies} 
    \label{tab:unnecessaryActions}
    \vspace{-14pt}
\end{table}

Some participants related unnecessary actions as an outcome of incorrect SA (5/16) (see Table{~\ref{tab:unnecessaryActions}}). These actions induce additional burden on the operator and/or robots, but without a considerable effect on task performance. Some unnecessary actions stemmed from inaccurate comprehension of the capabilities of robot autonomy (P1, P5, P9, P11). These can be related to human factors that contribute to SA gaps (Section{~\ref{sec:actualSA_human}}), namely poor mental models, lack of trust, and lack of willingness to delegate. Other unnecessary actions could be related to SA gaps arising due to lack of expertise and contextual understanding.

\subsubsection{\textbf{Confusion due to Incomplete SA}\nopunct}

Our participants shared several examples of confusion while working with the HRT applications (8/16), making them unsure whether to take action or not, eventually leading to delayed, unnecessary or faulty actions. See examples in Table{~\ref{tab:confusions}}. Reasons for these confusions are related to their actual SA being impacted by human factors, namely lack of expertise and contextual understanding, as well as the contextual factors, namely insufficient information and poor information formats.

\begin{table}[h!]
    \centering\small
    \begin{tabular}{|p{.12\linewidth}|p{.37\linewidth}|p{.44\linewidth}|}
        \hline
        \textbf{Type of} & \textbf{Example confusion} & \textbf{Reason for SA Gap} \\
        \textbf{Incomplete SA} & & \\
 \hline
         Perception &Wondering ``why the robot kept trying to go in a certain direction'' (P3) &  Incomplete perception of the robot's acting environment; ``<not perceiving> that the area was bigger than it looked''\\
         \hhline{~|-|-}
        
         &``Confused as to why the drone flew backward until <they finished the run and> looked at the scan'' (P10) &  Incomplete perception of the environment's layout\\
         \hhline{~|-|-}
        
         &Confused about the meaning of specific ``errors'', ``warnings'' and ``controls'' (P9-11) & The ``ambiguity <of> wording'', ``not telling the full story'' or ``not <having> translation of <controls>''\\
        
        \hline

         Comprehension &Not understanding and being confused about ``<why a certain> accident happened'' (P16), ``<why> the robot arm followed a weird path
        '' (P4), ``why the robot seem stuck'' (P13), ``<why> the drone is behaving <in a certain> way'' (P11), ``why that task allocation system might try to distribute robots in a certain way'' (P5) & Not being aware of certain information about underlying system (e.g., a weird behaviour of the robot arm occurred because the system calculates ``8 solutions for the path planning'' and uses a ``random'' choice (P4); the ``the underlying calculations <of task allocation process> are a bit hard to do on the fly in <one's> head'' (P5)), or when the system does not provide sufficient information to understand why the robot behaves in the way it does\\

        \hline

    \end{tabular}
    \caption{Examples of confusions due to the loss in incomplete perception, comprehension and prediction SA} 
    \label{tab:confusions}
    \vspace{-14pt}

\end{table}

\subsubsection{\textbf{Unnecessary Load and Fatigue due to Excess SA}}

We did not receive specific examples relating to excessive SA. However, our participants agreed that ``being able to comprehend detailed SA for every single robot'' or perceive or predict every detail related to the mission by ``overloading'' themselves is not a requirement, and it is expected ``to have holes in <individual SA>'' (P5).

\subsection{Strategies for Achieving Required Situational Awareness}

Delayed, missed, unnecessary and faulty actions and confusion that occurred due to gaps between actual and required SA levels contributed to poor mission performance. This was reflected in increased mission duration (9/16) and increased number of failed task completions (5/16). These actions have also contributed to other negative consequences, such as damage to the robot (8/16), damage or risks to the environment or humans in the environment (12/16), and end-user or operator frustration and stress (7/16). Our participants shared some human- and interface-initiated strategies that they currently employ to maintain their actual SA closer to the required levels to reduce these negative consequences. Further, they suggested several interface- and human-initiated strategies that  could assist them in reaching dynamic required SA levels.

\subsubsection{\textbf{Human-initiated Strategies Practiced within Current Applications}\nopunct}\label{sec:human-initiated-current}
\hfill

Within their current HRT applications, human operators employed two main self-initiated strategies to maintain the required SA: (1) performing frequent visual scans to update their big-picture awareness (10/16) and (2) reducing the required workload by either slowing things down or sequencing parallel activities (6/16). 
These strategies were identified to require human expertise and either increase cognitive and memory load or compromise mission performance.

\textit{\textbf{Frequent visual scans to update big-picture awareness:}} Many participants who could view robots with a visual interface reported performing frequent visual scans to update their big-picture awareness (10/16).

In remote HRTs with multiple robots or a single robot, human operators switch their attention between robots and mission-related tasks through regular ``scanning'' of different interface elements (P1, P9, P12). They ran this as a ``background strategy in the head'' (P5). In doing so, they frequently questioned themselves ``how much time has it been since they have looked at each robot'' and reminded themselves not to ``completely neglect a robot for more than <x> minutes and have at least a bird's eye understanding of where the robots are and the environment even when attending completely different tasks'' (P3, P5). Similarly, they made sure not to be ``stuck on one screen'' or panel (P3, P13) by intentionally switching their attention ``between the panels'', screens or ``indicators on the screen'' (P2, P11) ``to get a better perception of what was going on'' (P12). For example, operators performed visual scans to perceive ``battery status and position'' of the robot, ``whether or not all of the sensors are operational and any errors'' (P9). When robots are in teleoperation mode, this is done by ``stopping the controlling <task> every now and then'' (P12).

A similar strategy was used by operators in within-view HRTs with a single robot. ``Watching the robot'' at a distance in the environment provided them with big-picture awareness (P6). Therefore, when they are performing an action on the interface (e.g., providing input to the robot, examining logs), they ``constantly switch back and forth'' between the interface and view of the robot (P6). When ``the robot does what it is meant to do'', they claimed to ``focus more on the robot'' in the environment and ``less on the interface'' (P6). P7 also used this strategy during their crafting HRT application by attending the ``real view'' of the robot.

This ``habit of performing a scan'' (P9) ``on what is happening in the broader environment'' (P1) without ``spending too long on any one thing'' (P1) or falling into a ``tunnel vision'' (P5) was used to dynamically update their awareness about ``the global state'' (P5). This strategy involved maintaining ``a consciousness of how long they have been blind to the bigger picture'', so, if they ``are doing something that is narrowing the focus'', they remind themselves ``to look at the broader picture'' (P1). This act of remembering to ``systematically look at everything, instead of waiting for the interface to make something obvious'' (P9) is a secondary cognitive load. Also, it requires ``sufficient training to be cognisant of how much time roughly to spend on <something>'', given the application context (P2). 

\textit{\textbf{Manage workload demands through prioritisation or slowing robots down:}} When our participants could not obtain the required SA in situations where there were multiple robots requiring intervention, two parallel tasks or a task requiring increasing mental load, they prioritised one robot or task (P2-4, P12) or slowed down the robots (P6, P16) at the cost of mission performance (6/16).

Within remote multi-robot HRTs, when our participants comprehend that there is more than one robot requiring intervention, they reported selecting and attending to one robot first based on their expertise, while temporarily stopping other robots (P3) or assigning them alternative tasks (P3). This is because when two robots are ``in critical locations, trying to switch between both'' can lead to ``losing both'' (P3). In such a scenario, it was important to ``take a second to stop the robots'', so that the human ``can have a quick think and prioritise the robot in more danger'' (P2). In time-critical applications, before attending to the prioritised robot, some operators decided to ``let the <other> robot do something else for the time being'' until they are ``ready to deal with it'' (P3), by configuring ``a de-prioritisation region around dangerous area''. 

Our participants who had experience with remote and within-view HRTs with a single robot also related to this strategy. They practised arranging two tasks that can be done in parallel e.g., ``control of the mobile base and the initialisation of visualisation process'', in a way that those tasks ``can be done in sequence'' considering priorities (P4). This approach was useful when there is one human instead of a couple of humans. Similarly, relating to tasks of teleoperating and acquiring big picture knowledge, P12 shared that they prioritised ``scaning all the screens to get a better perception of what was going on'' by stopping ``either the driving or the controlling of the arm'' every now and then during their nuclear disaster response task.

Participants who had experience with within-view and co-located HRTs practised a slightly different but related strategy when the SA demands were associated with a single task. They found that when ``apples are really close to the robot arm'', increasing the risk of collision with the canopy, they need to maintain ``a heightened sense of awareness for the robot'' (P6). In this scenario, to manage the associated workload, they ``slowed the robot down a little bit'' to obtain ``a bit more time to analyse'' and to think whether they ``actually need to emergency-stop or not'' (P6). Similarly, P16, who worked with a co-located art manufacturing robot, shared an instance where they ran the robot at ``25\% speed'' in a situation where they realised that they ``do not have enough SA''.

Overall, rather than increasing their actual SA to achieve the required SA, these strategies were used to lower the required SA at the cost of mission performance. Further, these strategies rely on operators' expertise in timely recognition and prioritisation of interventions or tasks, and whether or not there is a need to slow down robots.

Finally, some participants shared certain human-initiated strategies they  practised specifically within their applications: restoring lost SA about robots outside communication range through bridging communication by sending other robots (3/16), using filtering tools provided by the interface to hide unnecessary details on the map view depending on the task at hand (3/16), and referring to a cheatsheet to reduce the memory load of remembering controls of an interface during initial use (2/16).

\subsubsection{\textbf{Interface-initiated Strategies Practiced within Current Applications}\nopunct}\label{Sec:4.5.2}
\hfill

We identified one interface-initiated strategy, used across remote, distant and co-located HRTs, namely alerts. Alerts enabled fast attention switching to pre-configured important events with the use of alternative interaction modalities (e.g., audio or haptic vs. visual), supporting the maintenance of required SA.

\textit{\textbf{Bring selected information to human’s attention through audio alerts:}} Simple ``audio cues'' (P5) have been implemented into some HRT applications to draw human attention to ``when the robots were doing anything that was not standard operation'' (P3) or experiencing unusual states (P9), in addition to showing those visually through dedicated user-interface elements (11/16).  
For example, these alarm-like alerts notify the human if a robot is idle and waiting for human input, so the human can allocate some tasks to improve mission performance (P3-4, P14). Audio alerts were also issued when the ``robot failed to do a task'', so the human could look into why that happened and suggest a workaround (P1). Audio notifications were also generated ``when a robot regained communication'', so operators could update their awareness about the new information the robot collected while away from the network (P1). Moreover, alerts that notified when a drone's ``battery level goes over certain threshold'' (P9) made humans issue commands on time to safely land or return it to where it was launched (P16). 
One application communicated the ``level of warnings'' by changing the ``beep <type, its loudness>, and how long it go for'' (P11).

Audio alerts supported the operators to ``attend to the emergencies quicker'' (P5) when the human operator was on another screen or took their eyes off the screen (P2, P3); i.e., when they do not perceive important information on the graphical interface promptly. However, this strategy will not be useful if the human is in a noisy environment (P6, P9, P10, P14). In such scenarios, the use of haptic feedback was identified as useful (P11, P16). Several improvements for alerts were also suggested. One suggestion is to not design too many audio alerts within one application, as it can cause challenges for the operator to differentiate alerts generated for multiple events or make people disregard the alerts (P6, P10). Another suggestion is to replace beep-like alerts with short verbal messages (e.g., ``low battery going home''), so that the alert is more ``directive'' and ``explicit'' (P5, P11). Further, participants suggested personalising the alert, namely configuring audio or haptic alerts to match operator's preference (P9) and not generating audio alerts when the user has already noticed relevant information through other modalities (P2).

\subsubsection{\textbf{Human-initiated Strategies Suggested as Future Directions}\nopunct}
\hfill

Our participants envisioned two main human-initiated strategies to assist in the maintenance of desired dynamic SA level. These strategies relate to using natural language queries, which enable them to request explanations to resolve confusion arising from incomplete SA (9/16), or to request information about robot and task status while visually attending to other things and thus reducing SA losses and SA delays (4/16).

\textit{\textbf{Request explanations about robot actions and status to reach common ground:}} Being able to request speech or visual ``explanations'' (P13) or ``reasoning'' (P4) when the operator does not fully understand the robot’s choices, behaviour and status was identified to be vital (9/16: P1, P3-6, P8, P10-11, P13).

During HRT missions, when robots ``are not doing what <the operator> wanted to do'' (P1) and the operator ``does not know why'' (P3), being able to say and ask ``I do not think you are doing the right thing, what is the rationale? why you are doing this rather than that?'' (P1) would allow the human to understand whether their intuition is wrong because they missed or misinterpreted certain information due to SA inefficiencies, or whether they are correct and can thus provide robots with helpful guidance based on their knowledge and expertise. P5 and P10 provided examples where clarification queries can be used to better understand the robot status. When it is difficult to estimate the robot's risk (P5), being able to ask the robot questions such as ``can you describe the type of surface you are on?'' (P5), ``what's the environment like?'' (P10), ``are you nearby <something>?'' (P5) and follow-up questions to understand how they arrive at those estimates was identified as a way to tune the operator's understanding and minimise their gap in SA, so they could confidently decide ``when to pull a robot out of a particular situation'' (P5). P13 and P8 provided example queries that they would like to use to clarify robot actions and interactions. P13 was interested in asking about the ``level of uncertainty'' that a robot experiences in conducting a task, as it can be used to decide ``whether the human should take over or not''. If the robot is ``very confident and <predicts> that it can make it'', then the human can ``leave the task to the robot''; if the robot says ``I cannot do this, because of these reasons'', that indicates the robot needs more help (P13). In contrast, P8 thought that being able to clarify at the beginning, ``what can the robot do for me? and what do I need to signal to the robot?'', could help novice users avoid unnecessary actions (e.g., attempting to stand up and reach the items carried by the robot without knowing that the robot can stretch out its arm, trying to keep the centre of their desks clear without knowing that the robot can also handover to right or left sides).

Our participants shared the importance of being able to receive clarifications after a failed or unsuccessful task or a mission. For example, in their fruit-picking application, being able to ask the robot ``why did you stop?'' can help operators understand whether it has happened ``due to excessive force in a particular joint'' as a result of ``too tight path planning constraints'' or due to something else (P6).  To analyse and reflect on poor mission performance or certain failed events, P1 and P5 found that ``it would be quite useful to visualise''  ``the short-term choices that the autonomy system makes'' in multi-robot HRTs to receive high-level reasoning for ``why a robot navigated certain paths'' and ``why certain task allocations have happened?''. 

Given that both the robot and the human understanding might be wrong, this strategy of requesting clarifications and receiving explanations is useful, as it allows the human to consider both human and robot perspectives and arrive at a more complete and accurate understanding. Hence, such capability can be used to decide whether an intervention is required or unnecessary during HRT missions before jumping in and overriding the robot's behaviours. The nature of the explanations needs to be designed to suit the time criticality of the situation, given that time being lost while a clarification discussion is taking place during a mission can potentially risk task failure (e.g., short vs. detailed messaged or representations, at a speed and time that allow easy comprehension) (P5, P11). Moreover, the ``transparency of the robot actions'' provided by this capability can support reaching accurate and complete mental models (P13), a key human factor influencing actual SA.

\textit{\textbf{Check the robot and task status using natural language-based queries without changing visual attention:}} 
Our participants suggested incorporating ``Alexa and Cortana style natural language interfaces'' to help with attending to multiple robots and tasks ``without changing <eye> gaze'' (P5) (4/16: P4, P5, P11, P14).

This strategy was identified to reduce the time spent on switching and restoring visual attention on interface elements when there is a demand to multitask. Further, it enables designing cognitively less-overloaded visual interfaces focusing on the most important information, rather than ``displaying everything all over time'' (P14). This leaves room for the human to ``uncover the world model'' by asking for further information beyond what is being displayed (P14). Also, this strategy provides the opportunity to identify the conflicts or discrepancies between the described robot's state and what actually is observed, which may arise due to system malfunctions (P11). Further, it would support ``creating a shared mental model through being aware of the robot's mental model'' by asking a series of questions, such as ``What does the robot think it has done? What is it currently doing? 
What does it think it needs to do next?'' (P14).

In addition, participants shared other directions for human-initiated strategies specific to their applications, such as using augmented reality when awareness of  the 3D details of a remote robot's environment is needed (2/16), listening to robots in a remote environment or feeling haptic patterns modelled to communicate terrain characteristics when it is difficult to understand underlying terrain (2/16), and checking whether human predictions about the task outcome align with robot predictions through speech or visual simulation before instructing a robot to carry out a task (4/16).

\subsubsection{\textbf{Interface-initiated Strategies Suggested as Future Directions}\nopunct}
\hfill

Our participants had two suggestions for future interface-initiated strategies to bring salient information to users' attention, namely developing interfaces capable of predicting the need of human intervention (11/16) and tracking user attention (9/16). Another suggestion is modifying robot behaviour to reduce human workload, given the ability to track user states  (7/16).

\textit{\textbf{Predict incidents requiring human intervention and bring priority information to human's attention:}} 
In addition to current implementations bringing pre-configured events (e.g., errors, idling states, battery warnings) to the user's attention through system-initiated audio alerts (see Section \ref{Sec:4.5.2}), our participants suggested that providing the robot with the ability to detect the likeliness of hazards, novelty, uncertainties and inefficiencies can be used to bring high priority safety-, time- and performance-critical information to human attention (11/16: P1, P3, P5-6, P8-10, P13-16).

Designing a robotic ``system to have knowledge of what is important for the user to know'' is an ``open research problem'' (P1, P9). However, ``being able to recognise unusual'' or ``uncertain'' situations in which the robots do not ``have a policy'' (e.g., in new environment or ``operate near a negative obstacle'' that they are not trained for), and being able to notify the human can reduce the SA demands on humans (P1, P5, P16). P8 related this capability to requesting assistance from AI-based detection of ``anomaly activity''. P10 related it to ``predicting whether the mission is likely to be unsuccessful or high risk'' through assessing ``how hostile the environment is''. This ``reactive'' capability of robots, i.e., being able to ``ask the operator for help'' or input and ``if there is no immediate response, doing something else until the operator provides guidance'', was viewed as a powerful strategy (P1, P6, P14). Our participants shared several examples of robots asking for help, e.g., ``I have seen something that I have not seen before, so I do not know what to do with it''
(P3), ``there is something that looks like hazardous, so I do not need to go there without you confirming if that is safe or not'' (P1), ``I have failed at this traversability task once. Is this still something that I should try or not?'' (P1), ``I am reaching an area that is not very well traversable. So you better take control'' (P13), ``a collision is about to happen, instruments are too close, make the necessary adjustments'' (P15), ``You asked me to go in this direction, but I think that there is a high concentration of gases in this particular direction, do you still want me to go there?'' (P9), ``I noticed your instruments are moving from the center, do you want me to move the camera focus to that side?'' (P15). P3 further suggested sorting the signs of casualties that the ``human needs to verify, based on uncertainty scores'', so that human can plan to spend more time on potential false positives and quickly verify objects that the system is more certain about.

This strategy will reduce SA demands on humans by reducing the ``cognitive load'' associated with the need to ``constantly watch''  and understand ``when and whether to intervene and override robots' action'' (P5-6). It can also be a tool to gain better SA by robots when robots are ``not adequately confident to deal with'' certain situations and have the ability to ``learn and reduce the number of future queries'' (P1, P5).

\textit{\textbf{Track user's attention to assist in dynamically updating big-picture awareness:}}
As a means to overcome cognitive and memory loads associated with the currently practised strategy of \textit{frequent visual scans to update the big-picture} (see Section \ref{sec:human-initiated-current}), some participants suggested how eye gaze tracking capability can be integrated and used to remind humans when no recent visual scan was performed (9/16: P1-3, P7, P11-12, P14, P16).

    For example, P1 and P12 suggested running a background process ``to observe where the operator's time is being spent'', so when it identifies that they spend most of the time ``on a very narrow thing'', the system can generate a prompt to ``remind to check the wider surrounding''. Here, the narrow aspect can be ``fixating on <a single> agent, when there are other agents'' (P1), ``fixating on a particular area'' on the screen (P1) ``spending too much time on one panel'' when there are multiple panels to toggle between (P2, P11), or not fixating on HRT mission related aspects indicating that the operator is ``distracted'' (P14). This strategy was identified to assist the human operators, especially ``novice operators'' (P2), to focus on their role in HRTs without ``being very conscious about the amount of time that they would stay on <something> and reminding themselves to take the broader view'' (P1). This strategy can also assist in mitigating delayed and missed actions arising from SA latency and SA losses (P7).

While these system-generated reminders to update the big picture were identified as useful, our participants also suggested ideas for further assistance to ease the efficient information extraction and restoration of awareness on previously attended tasks. One suggestion was to ``highlight certain panels and blurring out certain areas'' (P3), considering information useful to achieve big-picture awareness.  
P1 recognised the reorientation process following a visual scan, which restores the awareness on tasks that the human was previously focusing on, as a hurdle. Hence, when the system identifies that the human has not updated the big picture recently, ``if the operator is being drilled in very close to one thing'', rather than requesting to do a visual scan, ``informing human about the surrounding'' (P12) was identified to be useful. They suggested doing this by posting ``a small simplified representation of big-picture <summary>, like a postage stamp'' (P1) or ``increasing the information density where the user is looking at'' (P5). Information to highlight to support efficient visual scans or include in such summaries needs to depend on the application context; e.g., in rescue and harvest inspection applications, details such as ``object detections unreviewed for a certain amount of time'', and ``a robot in a particular state that has not been viewed for a certain amount of time'' (P2). When to ask to update big-picture awareness is also context dependent; e.g., when collaborating with ``a robot with higher payload, operator's attention needs to be higher on the robot'' (P16).

\textit{\textbf{Track user's state to modify robot behaviours or autonomy:}} Our participants suggested if the HRT system can estimate users' behavioural states (e.g., attention and unusual movements, such as falling) and cognitive states (e.g., stress, workload and fatigue), those estimates can be used to modify robot behaviour or autonomy to take over some load from the human (8/16: P3, P5-6, P10, P12-14).

By making the human collaborators' cognitive states and ``behaviours legible to the robot'', it was expected that robots could ``make better decisions'' and ``understand if the human needs help or not'' (P13, P14). P3 and P5 suggested to ``stop the robots'' in less time-critical scenarios, ``if the operator is not paying attention'' or ``if the operator falls over'' to ``safeguard against accidents''. P4 and P14 suggested to ``give the operator a break'' upon ``tiredness detection''. P8 envisioned that it would be beneficial to design the robots to undertake ``a more conservative strategy''; e.g., ``follow the safest option, rather than trying to adapt risk'', when the operator is experiencing higher cognitive loads, stress or fatigue levels. Some high-level ideas such as customisation of explanation by understanding how the collaborator feels were suggested (P6).

Relating to HRT systems that are or will be implemented with varying levels of autonomy, P7, P10, P12 and P13 highlighted that robots would be able to ``alleviate the burden of the operator'' related to SA by ``taking more responsibility'' when the operator is experiencing high ``mental fatigue, physical fatigue,'' ``stress or workload''. However, it was identified that this strategy would not be feasible for HRT ``application with very well-defined tasks between the robot and the human'' because it is not possible to ``make the robot take over some of the tasks that <human> have to do'' (P4). Further, our participants emphasised that there might be certain events that  humans ``do not like to hand over to the robot even if the workload is high,'' (P12). This indicates the need to facilitate configurations and provide authority to humans ``when negotiating the decisions of control switch'' (P13). Having access to estimates of the robot's ``confidence level in completing a certain task'' was identified to help this human decision of ``which tasks to delegate and which tasks not to'' (P10). Also, the reason for high stress or workload needs to be considered; e.g., is the operator ``overloaded because the robot has gone haywire or because there is a lot going on''; in the former case, it is obvious that the solution should not be delegating more tasks to the robot (P10). P13 highlighted that there might be unanticipated negative consequences associated with this strategy, e.g., this strategy may indicate that the human is ``not performing well, and then that might be a discouragement''. Further investigation is needed to determine the effect on the operator and implications for designing this strategy.

Additionally, our participants suggested some context-specific interface-initiated strategies relating to remote HRTs. They proposed designing communication quality-aware information streams so that critical information can be prioritised when robots have limited communication with the human operator (1/16). In addition, communication quality-aware robot behaviour was suggested to allow the robot to resume communication and ask the operator on how to proceed when communication was lost temporarily (e.g., ``I just went beyond the comms, should I proceed?'') (1/16). 

In addition to human- and interface-initiated strategies elaborated throughout this section, some other strategies shared by our participants could support the overall efficiency of dynamic SA. 
These are standard human-computer/ human-machine interface (HCI/HMI) design strategies. Examples include simplifying the interface and formatting information within the interface by considering information useful for the HRT mission (e.g., remove unnecessary information, make useful information more prominent, use succinct presentations, colour schemes to differentiate controls, adjust the positioning of different views to reduce mental load of information processing) (6/16), undertaking prior training to familiarise themselves with the interface and improve their expertise and contextual knowledge (5/16), distribute the cognitive load by having multiple operators responsible for different tasks or robots (2/16), and following pre-mission routines where they confirm the robot is functioning as it should be (2/16). Some other strategies that were mentioned tend to overload the operator with information without dynamic adjustments, e.g., improving the view range of the robot's environment by changing camera locations or mounting 360-view or additional cameras additional cameras (4/16), and increasing the frequency and redundancy of certain information (3/16). Further, some participants referred to automating robots further by improving robots' awareness about their surroundings and other robots so robots can conduct specific tasks by themselves, therefore reducing operators' SA demands (7/16). This strategy may hinder the benefits of human-robot collaboration.

Overall, the strategies detailed in this Section could reduce the majority of SA gaps reported in Section {\ref{sec:SAgap}}, by improving the operator's attentional division skills, concentration skills, expertise and contextual knowledge over time, and distributing the workload. These strategies address the human and robotic factors and some contextual factors of actual SA. However, improved undisrupted communication is needed to  address SA inefficiencies due to the contextual factor of communication lags or temporary communication loss between the robot and human interface.

\section{Discussion}

The findings of our interview analysis demonstrate that the proposed framework of dynamic SA is applicable across diverse HRT contexts. It provides novel directions for assessing SA and designing intelligent user interfaces to assist with the maintenance of required levels of SA during HRT missions.

\subsection{Generalisability of The Dynamic Situational Awareness Framework and Associated Concepts}

All of our participants supported the idea that the required SA levels as well as SA levels that they experience (actual SA) during HRT missions vary over time, and lead to gaps between required and actual SA. Given that our  participants (1) represented diverse HRT applications, including co-located, distant and remote HRTs, high-to-low time-critical and high-to-low safety-critical HRT settings, and single to multiple robot HRTs within diverse domains (e.g., disaster response, medical, aesthetic); (2) represented both technical and non-technical collaborators; and (3) the data they shared included both novice and expert users' experiences\footnote{Although all our participants were expert users, they shared their own experiences as novices as well as observations of other novices' experiences}, we postulate that the presented dynamic situational awareness framework is generalisable across diverse HRT applications and diverse human collaborators.

The required SA was found to fluctuate due to 5 contextual factors, 3 robot-related factors and one human factor, as detailed in Section~\ref{sec:requiredSA}. According to our findings, required SA increases with increased level of task criticality and challenge (contextual), safety criticality of tasks (contextual), the number of robots per human (contextual), time criticality of the mission (contextual), and severity level of the robot’s status (robot);  required SA decreases with increased time spent on a mission (contextual) and software and hardware robustness (robot). As the capability of robot autonomy (robot) increases, required SA decreases.  However, this trend is reversed for mixed-initiative variable-autonomy applications~\cite{chiou2021mixed} if there is a disagreement between decisions made by robots and humans on who should be in control. Furthermore, required SA was found to vary given the role of the human, which depends on the HRT context; e.g., changes to the role when the robots run in different autonomy levels and when working physically closer to or remote to robots. 

We also found that actual SA levels are moderated by 5 human, 3 contextual and 2 robot-related factors, as detailed in Section~\ref{sec:actualSA}. Lack of expertise and contextual understanding (human), poor mental models about robots (human), under- or over-trust (human), lack of cognitive capacity and ability to multitask (human), lack of willingness to delegate tasks when robots have the capability to handle those tasks (human), distractions (contextual), higher distance to the robot (contextual) and inefficient information formats in the robotic interface (robot) decrease the actual SA. Moreover, we identified that some factors influence not only actual SA but also required SA. While communication-related interruptions to information availability (contextual) and missing information in the robotic interface (robot) lower a human's actual levels of SA, as they occur, the required SA levels are also identified to be changing. For example, when robots explore an area outside the communication range with the human collaborator, the human's actual SA about the robots degrades, and at the same time, the required SA may be increasing if the robots are in danger.  Similarly, humans will not be aware of information not provided through the robotic interface. Therefore, these factors act as confounding factors on the association between required and actual SA. 

SA latencies, SA losses, SA inaccuracies, incomplete SA and excess SA were identified to arise due to gaps in required vs. actual SA across diverse HRT contexts. Across diverse settings, the first four inefficiencies were found to contribute to delayed actions, missed actions, faulty actions and confusion (eventually leading to delayed, unnecessary or faulty actions), respectively. Evidence supporting the hypothesis that excess SA leads to unnecessary cognitive load and fatigue was limited, which may be due to self-identification challenges. Overall, SA inefficiencies were found to occur at a higher frequency in remote HRT settings as opposed to co-located HRT, with novice operators compared to expert operators, when the operators are experiencing higher workloads (e.g., when the number of robots assigned for an operator is high, task and safe criticality is high, level of autonomy is low), when there is a lack of trust towards robot autonomy and lack of willingness to delegate tasks, as well as when the underlying systems are not efficient or consistent (e.g., issues with hardware and software robustness, inefficiencies in information presentations, intermittent network latencies). This suggests that assistance to maintain SA may be more beneficial for certain user groups and in specific HRT contexts.

\begin{figure}[h!]
  \centering
  \includegraphics[width=0.74\linewidth]{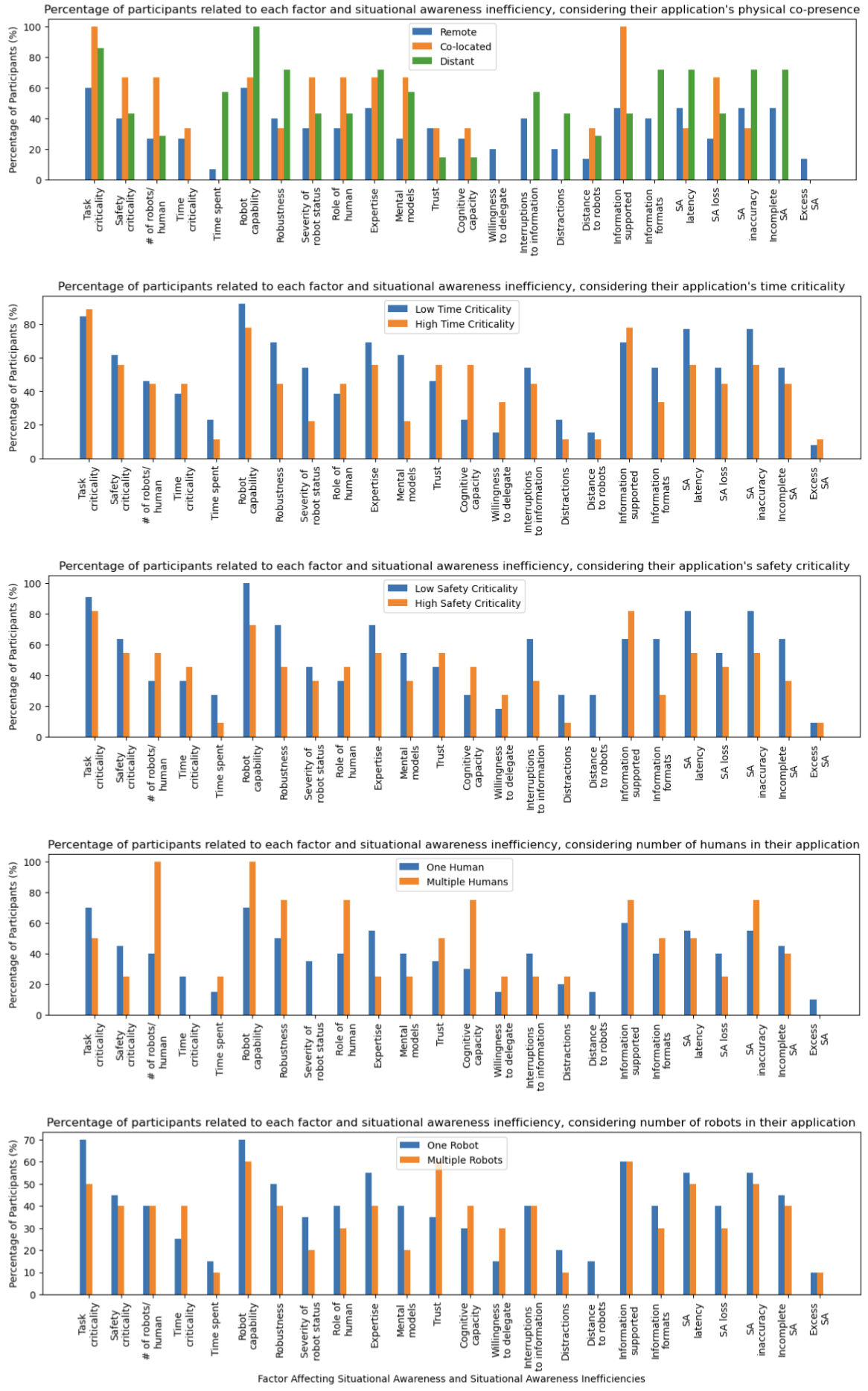}
  \caption{A detailed breakdown of percentage of participants who relate to each contextual, human and robot factor affecting required and actual SA and each SA inefficiencies, considering their HRT application context}
  \label{fig:Robotic-SAinefficiency}
\end{figure}

The factors contributing to required and actual SA variations, their trends and four SA inefficiencies are generalisable across diverse HRT contexts, given the agreement between the majority of our participants. Figure{~\ref{fig:Robotic-SAinefficiency}} presents an overview of the percentage of participants who reported on each factor and SA inefficiency across different HRT contexts (Table 24 in Appendix C lists the number of participants).  Distance to the robot (a contextual factor affecting actual SA) and excess SA (an SA inefficiency) were the least supported by our participants; e.g., participants who represented HRTs with multiple humans did not relate to this factor or SA inefficiency.

\subsection{Connections to Related Work}

Our dynamic situational framework proposes one way to address a critical research gap highlighted in a recent review on transparency and explainability in variable autonomy robotic systems{~\cite{10.1145/3645090}}. In their review, Mathnani et al. stress that changing levels of autonomy will require changing levels of information to deliver different levels of transparency.  They highlight that adequate transparency does not mean exposing everything, but using transparency to induce comprehensibility appropriate to the stakeholder in a time-appropriate manner. This resonates with our finding that the required SA level varies given the level of the robot's autonomy. Further, they highlight the need to identifythe most opportune moment to present information, while considering the challenges of SA. Our framework and findings propose that the occurrence of gaps in actual and required SA along the mission timeline can be used as a temporal adaptation criterion, which is dependent on a range of contextual, human and robot-related factors. Implications of this novel contribution are further discussed in Section{~\ref{sec:discusionEstimates}}.

Previous literature indirectly supports the argument that SA is a dynamic trade-off and depends on certain user, robot and context-specific factors. For example, the number of robots a human can monitor or control simultaneously (i.e., fan-out) is considered a measure of human attention{~\cite{10.1145/985692.985722}} due to the need for switching attention between agents{~\cite{5530413}}. Previous work has also identified the need for effective techniques to make operators aware of pertinent information regarding a robot and its environment during safety- and time-critical missions that impose requirements for error-free and efficient operations, respectively{~\cite{drury2003awareness}}. Existing works also identify that SA is more challenging during the teleoperation of low-autonomy robots, compared to guiding robots with reliable and robust capabilities {~\cite{10.3389/frobt.2022.704225}}. These have been studied as individual problems, and have not been collectively identified as influencing the dynamic nature of required SA. Similarly, factors such as distraction, inability to multitask and increased distance to robots have been recognised to impede SA{~\cite{chen2022situation,chen2007human}}. Factors such as calibrated trust, accurate mental models, and visualisations summarising key relevant information  have been identified as necessary for better SA {~\cite{MATTHEWS202159,Daniel2021HRI}}. However, what constitutes better SA is not well-documented. We propose that better SA can be achieved by reducing the gap between actual and required SA. Furthermore, we provide a synthesis view, unifying literature into a cohesive framework through the lens of dynamic situational awareness through a systematic investigation.

Furthermore, our factors share similarities to the work presented by Riley et al. ~\cite{riley2010situation}, who analyse the impact of a range of HRI factors on (actual) SA, by categorising those factors into:  Task / Mission Factors, System Factors, Environmental Factors, Individual and Team Factors, and External World and User Interface Design Factors. While Riley et al. do not differentiate between required and actual SA, some of the factors identified in their work (task difficulty, controlling multiple robots, distractions, operator's lack of knowledge, cognitive overload, communication latencies and robot's autonomy) contribute to degrading the actual SA. However, different from Riley et al.'s hypothesis that ``operators need to maintain a {\emph{high}} level of SA with regard to changes in the environment'', our work suggests that operators need to maintain a {\emph{required}} level of SA with regard to the dynamic nature of HRT missions without needing to maintain high SA at all times. We reinforce this argument and work by Mathnani et al.{~\cite{10.1145/3645090}} by providing evidence from interviews with human operators (Section{~\ref{sec:SAgap}}) and synthesise it through the proposed framework (Figure{~\ref{fig:DSAFramework}}).

Regarding the SA inefficiency classification provided in Section{~\ref{sec:SAgap}}, a few of these concepts are well-defined in the literature (e.g., SA latency~\cite{senaratne2023roman}), and some of the concepts have been vaguely defined (e.g., SA loss~\cite{onnasch2014human}). SA latency can be considered as a re-framing of a well-known problem, inattentional blindness{~\cite{nakayama1999inattentional, simons1999gorillas}}. Inattentional blindness is a phenomenon where individuals fail to notice obvious but unexpected objects or events in their visual field when their attention is engaged with another task, even if the unexpected stimulus is within their spatial focus of attention. SA loss can be regarded as infinite SA latency.
To the best of our knowledge, our work is the first to unify these concepts as a spectrum of outcomes resulting from differences in actual and required SA with regards to content (which information \emph{needs} to be vs. \emph{was} perceived, comprehended and predicted) and the time when the awareness \emph{should} vs. \emph{did}  occur. 

\subsection{Implications for Deriving Useful SA Estimates and Implementing Adaptive Interface Assistance}\label{sec:discusionEstimates}

One of the first steps towards providing user- and time-adaptive assistance to maintain required SA levels is to enable HRT systems to detect the gaps in SA by estimating the actual and required SA levels dynamically. Our findings suggest that actual SA and required SA can be viewed as functions of factors. Estimating each factor using objective measures can be a potential direction for future work. As objective measures, our participants suggested that eye gaze tracking and interfaces that accommodate natural conversations with robots can be helpful. This is in contrast to conventional SA estimates used within HRT applications (e.g., post-study questionnaires ~\cite{hopko2021effect,dini2017measurement} and freeze-probe questionnaires~\cite{dini2017measurement,endsley1988situation}), which cannot provide fine-grained assessments to estimate SA continuously without disturbing the human-robot collaboration flow. Using eye gaze measurements to estimate SA latency has been proposed~\cite{senaratne2023roman}; however, defining actual and required SA as a function of identified factors and the use of other natural interaction modalities such as speech for this purpose is yet to be explored. Future work in this field needs to continuously estimate other SA inefficiencies (SA loss, SA inaccuracy, incomplete SA and excess SA) without interrupting HRT interactions or burdening the operator.

Our participants emphasised the importance of maintaining ``high level'' or ``abstract'' SA vs. ``low level'' or ``detailed'' SA given different scenarios, indicating the need to detect these two SA levels differentially. This conceptualisation of SA was also related to the notion of narrow vs. big-picture focus. According to our framework, the high-level SA was found to be required at all times, and low-level SA was recognised to be useful only when assisting robots when they are not acting as expected. However, in the literature, the definitions of high-level and low-level SA are inconsistent. Some researchers claim that one can acquire high-level SA only by acquiring low-level situational awareness~\cite{lijing2022research}, because low-level SA is defined as basic perception capabilities and a limited understanding of the environment, and high-level SA is defined as a sophisticated understanding with advanced perception, comprehension and prediction capabilities that support handling a broad array of complex scenarios and adapting to dynamic changes in the surroundings. There is also work that suggests a concept called Situation Overview (SO) in relation to high-level SA, which consists of mission SO and Robot SO~\cite{petersen2014general}. These definitions differ from each other and what our participants suggested. In this paper, we refer to detailed SA when the human has complete 3-level SA on one aspect and high-level SA when the human has SA critical for the overall mission success. Given these discrepancies in definitions, first, ``abstract'' vs. ``detailed'' SA concepts must be carefully defined, then estimated and evaluated for their impact on maintaining dynamic SA requirements in HRT contexts.

In contrast to existing approaches that mainly suggest providing most of the information or additional control to human collaborators, increasing their cognitive load~\cite{tanke2013improving, sumigray2021improving, hong2020effect, fink2023expanded} or superficial modifications to the user interface, reducing the cognitive load~\cite{pitman2007picture} (also see Section~\ref{relatedwork:strategies} for a brief review of current approaches to improve SA in HRT), most of our participants expressed a preference for intelligent interfaces that can be adaptive to their actual SA levels and support maintaining it closer to required SA levels. This finding augments and extends existing literature as it argues that SA assistance is not simply about providing more information{~\cite{10.3389/frobt.2022.704225,yanco2004beyond}}. Examples include assisting the operator to update big-picture awareness based on the system's ability to track their attention, bringing priority-based information to their attention based on the system's ability to predict safety critical incidents or robot inefficiencies, suggesting to delegate certain tasks based on the system's ability to track user cognitive load and fatigue, facilitating to check the status of robots and tasks and request explanations to reduce confusion (i.e., fill the gap of incomplete SA) using natural language-based queries. These directions are broad, and some  have also been echoed in related literature; e.g., use of explainable AI techniques to establish shared mental models in human-machine teams to reach common ground{~\cite{andrews2023role}}, and cueing operator attention to the high priority information identified through automatic change detection capabilities using salient visual cues{~\cite{cook2015displays}}). Incorporating these suggestions may also require overcoming other technical challenges, e.g., using efficient natural language processing architectures to minimise response delay. Our interviews suggest more context-specific approaches, given that priority information, safety-critical incidents, robot inefficiencies, levels of autonomy, and underlying implementations causing confusion could vary from one HRT to another. Therefore, future work needs to consider how some of these strategies can be designed and implemented to be easily integrated and configured into different HRT systems.

\subsection{Limitations} 

Our findings are constrained by the sample size of this study (n=16) and the variability of participants' experiences. For instance, examples we received on good mental models about robots were mainly from roboticists who had been exposed to underlying algorithms of robots, given that we had limited participants who were practitioners without a robotic background (2/16). Similarly, potential interface and knowledge transfer limitations revealing the robots’ decision-making processes were addressed mostly from expert users' point of view. These examples could wrongly imply that operators need to be robot experts. Hence, future work needs to target a more comprehensive list of factors influencing actual and required SA in HRTs involving humans with diverse expertise and backgrounds, such as end-users or domain experts with limited technical knowledge of the robots. This will allow us to generate further insights into whether certain contexts or confounds affect the general trends found regarding whether SA demand rises or falls with respect to lower or higher values of each factor; confirming that the reported SA inefficiencies (SA loss, SA latency, SA inaccuracies, incompleted SA and excess SA) cover the full spectrum of SA gaps; generating further insights into how SA inefficiencies relate to adverse outcomes (missed actions, delayed actions, unnecessary actions, faulty actions, confusions and cognitive overload); and revealing detailed and generalisable methods to assess fine-grained estimates of SA gaps and to implement assistive technology to support the maintenance of dynamic required SA.

\section{Conclusion}

Situational awareness (SA) of human collaborators within human-robot teams (HRTs) influences the success and efficiency of the collaboration. We aimed to investigate how the required SA levels of human collaborators can be conceptualised and maintained within the context of diverse HRTs. We propose a framework for dynamic situational awareness (DSA), capturing how the required and actual levels of SA fluctuate within HRT missions, creating five types of SA inefficiencies based on evidence collected from an interview study with 16 people with repeated and diverse experience with HRTs.   

The dynamic SA framework, as shown in Figure{~\ref{fig:DSAFramework}}, illustrates that the gaps in actual and required SA result in delayed, missed, wrong, partial or unnecessary perception, comprehension, and predictions, leading to SA latency, SA loss, SA inaccuracies, incomplete SA and excess SA. These SA inefficiencies were found to contribute to delayed actions, missed actions, faulty or unnecessary actions, confusion and unnecessary cognitive load or fatigue by human collaborators within diverse HRTs, making the human-robot collaboration and mission performance ineffective. 

Our analysis of interviews reveals a list of contextual, human and robot factors affecting the changes in actual and required SA levels. It also indicates currently-used as well as proposed operator and interface-initiated strategies that can assist in maintaining SA at required levels. The dynamic SA framework and most of the supplementary findings are supported by participants representing high-to-low time critical, high-to-low safe critical, co-located and remote, and single to multiple robot HRTs within diverse domains (e.g., disaster response, medical, aesthetic). Some of the strategies suggested to assist in maintaining required SA are context-specific. Further research is needed to better understand the generalisability of the results and the opportunity to extend the proposed framework further. 

The scientific importance of the proposed framework lies in the spectrum of SA inefficiencies that it suggests, indicating the need for more fine-grained SA measures, such that the estimates of the occurrence of these SA gaps could be used not only to evaluate the success of resulting human-robot collaboration but also to implement user- and time-adaptive assistance to maintain required SA levels. Therefore, this framework motivates and encourages future research in objective SA assessment to achieve the above envisioned impacts. In parallel, the designers of collaborative HRTs can utilise the suggested adaptive interface strategies to create next-generation intelligent and effective interfaces for HRTs.

The focus of the proposed dynamic SA framework and this study is on individual human SA in HRT. During our study, we received limited insights related to the robot's SA, which we did not report in this paper. Furthermore, although it was not a focus, we provided a couple of examples of how required SA relates to team SA (shared and distributed SA). In future research, we are interested to extend our framework to define required and actual SA for robot agents and associated concepts, as well as to investigate the interplay of required and actual SA of humans, robots, and team SA.

\begin{acks}
We thank our participants for their intellectual contributions to this research and Ronal Singh for brainstorming certain ideas around the framework. 
The first author also acknowledges the funding support received from the CSIRO Collaborative Intelligence Future Science Platform during this research.

\end{acks}

\bibliographystyle{ACM-Reference-Format}
\bibliography{sample-base}


\begin{thebibliography}{76}


\ifx \showCODEN    \undefined \def \showCODEN     #1{\unskip}     \fi
\ifx \showDOI      \undefined \def \showDOI       #1{#1}\fi
\ifx \showISBNx    \undefined \def \showISBNx     #1{\unskip}     \fi
\ifx \showISBNxiii \undefined \def \showISBNxiii  #1{\unskip}     \fi
\ifx \showISSN     \undefined \def \showISSN      #1{\unskip}     \fi
\ifx \showLCCN     \undefined \def \showLCCN      #1{\unskip}     \fi
\ifx \shownote     \undefined \def \shownote      #1{#1}          \fi
\ifx \showarticletitle \undefined \def \showarticletitle #1{#1}   \fi
\ifx \showURL      \undefined \def \showURL       {\relax}        \fi
\providecommand\bibfield[2]{#2}
\providecommand\bibinfo[2]{#2}
\providecommand\natexlab[1]{#1}
\providecommand\showeprint[2][]{arXiv:#2}

\bibitem[Adriaensen et~al\mbox{.}(2022)]%
        {ADRIAENSEN2022103320}
\bibfield{author}{\bibinfo{person}{Arie Adriaensen}, \bibinfo{person}{Nicole Berx}, \bibinfo{person}{Liliane Pintelon}, \bibinfo{person}{Francesco Costantino}, \bibinfo{person}{Giulio {Di Gravio}}, {and} \bibinfo{person}{Riccardo Patriarca}.} \bibinfo{year}{2022}\natexlab{}.
\newblock \showarticletitle{Interdependence Analysis in collaborative robot applications from a joint cognitive functional perspective}.
\newblock \bibinfo{journal}{\emph{International Journal of Industrial Ergonomics}}  \bibinfo{volume}{90} (\bibinfo{year}{2022}), \bibinfo{pages}{103320}.
\newblock
\showISSN{0169-8141}


\bibitem[Ajoudani et~al\mbox{.}(2018)]%
        {ajoudani2018progress}
\bibfield{author}{\bibinfo{person}{Arash Ajoudani}, \bibinfo{person}{Andrea~Maria Zanchettin}, \bibinfo{person}{Serena Ivaldi}, \bibinfo{person}{Alin Albu-Sch{\"a}ffer}, \bibinfo{person}{Kazuhiro Kosuge}, {and} \bibinfo{person}{Oussama Khatib}.} \bibinfo{year}{2018}\natexlab{}.
\newblock \showarticletitle{Progress and prospects of the human--robot collaboration}.
\newblock \bibinfo{journal}{\emph{Autonomous Robots}} \bibinfo{volume}{42}, \bibinfo{number}{5} (\bibinfo{year}{2018}), \bibinfo{pages}{957--975}.
\newblock


\bibitem[Allenspach et~al\mbox{.}(2023)]%
        {allenspach2023design}
\bibfield{author}{\bibinfo{person}{Mike Allenspach}, \bibinfo{person}{Till K{\"o}tter}, \bibinfo{person}{Rik B{\"a}hnemann}, \bibinfo{person}{Marco Tognon}, {and} \bibinfo{person}{Roland Siegwart}.} \bibinfo{year}{2023}\natexlab{}.
\newblock \showarticletitle{Design and Evaluation of a Mixed Reality-based Human-Robot Interface for Teleoperation of Omnidirectional Aerial Vehicles}. In \bibinfo{booktitle}{\emph{2023 International Conference on Unmanned Aircraft Systems (ICUAS)}}. IEEE, \bibinfo{pages}{1168--1174}.
\newblock


\bibitem[Andrews et~al\mbox{.}(2023)]%
        {andrews2023role}
\bibfield{author}{\bibinfo{person}{Robert~W Andrews}, \bibinfo{person}{J~Mason Lilly}, \bibinfo{person}{Divya Srivastava}, {and} \bibinfo{person}{Karen~M Feigh}.} \bibinfo{year}{2023}\natexlab{}.
\newblock \showarticletitle{The role of shared mental models in human-AI teams: a theoretical review}.
\newblock \bibinfo{journal}{\emph{Theoretical Issues in Ergonomics Science}} \bibinfo{volume}{24}, \bibinfo{number}{2} (\bibinfo{year}{2023}), \bibinfo{pages}{129--175}.
\newblock


\bibitem[Arnold et~al\mbox{.}(2018)]%
        {arnold2018search}
\bibfield{author}{\bibinfo{person}{Ross~D Arnold}, \bibinfo{person}{Hiroyuki Yamaguchi}, {and} \bibinfo{person}{Toshiyuki Tanaka}.} \bibinfo{year}{2018}\natexlab{}.
\newblock \showarticletitle{Search and rescue with autonomous flying robots through behavior-based cooperative intelligence}.
\newblock \bibinfo{journal}{\emph{Journal of International Humanitarian Action}} \bibinfo{volume}{3}, \bibinfo{number}{1} (\bibinfo{year}{2018}), \bibinfo{pages}{1--18}.
\newblock


\bibitem[Bauer et~al\mbox{.}(2008)]%
        {bauer2008human}
\bibfield{author}{\bibinfo{person}{Andrea Bauer}, \bibinfo{person}{Dirk Wollherr}, {and} \bibinfo{person}{Martin Buss}.} \bibinfo{year}{2008}\natexlab{}.
\newblock \showarticletitle{Human--robot collaboration: a survey}.
\newblock \bibinfo{journal}{\emph{International Journal of Humanoid Robotics}} \bibinfo{volume}{5}, \bibinfo{number}{01} (\bibinfo{year}{2008}), \bibinfo{pages}{47--66}.
\newblock


\bibitem[Bedny and Meister(1999)]%
        {bedny1999theory}
\bibfield{author}{\bibinfo{person}{Gregory Bedny} {and} \bibinfo{person}{David Meister}.} \bibinfo{year}{1999}\natexlab{}.
\newblock \showarticletitle{Theory of activity and situation awareness}.
\newblock \bibinfo{journal}{\emph{International Journal of cognitive ergonomics}} \bibinfo{volume}{3}, \bibinfo{number}{1} (\bibinfo{year}{1999}), \bibinfo{pages}{63--72}.
\newblock


\bibitem[Begum et~al\mbox{.}(2015)]%
        {begum2015collaboration}
\bibfield{author}{\bibinfo{person}{M Begum}, \bibinfo{person}{R Huq}, \bibinfo{person}{R Wang}, {and} \bibinfo{person}{A Mihailidis}.} \bibinfo{year}{2015}\natexlab{}.
\newblock \showarticletitle{Collaboration of an assistive robot and older adults with dementia}.
\newblock \bibinfo{journal}{\emph{Gerontechnology}} \bibinfo{volume}{13}, \bibinfo{number}{4} (\bibinfo{year}{2015}), \bibinfo{pages}{405--419}.
\newblock


\bibitem[Chen and Terken(2022)]%
        {chen2022situation}
\bibfield{author}{\bibinfo{person}{Fang Chen} {and} \bibinfo{person}{Jacques Terken}.} \bibinfo{year}{2022}\natexlab{}.
\newblock \showarticletitle{Situation Awareness, Multi-tasking and Distraction}.
\newblock In \bibinfo{booktitle}{\emph{Automotive Interaction Design: From Theory to Practice}}. \bibinfo{publisher}{Springer}, \bibinfo{pages}{83--100}.
\newblock


\bibitem[Chen et~al\mbox{.}(2007)]%
        {chen2007human}
\bibfield{author}{\bibinfo{person}{Jessie~YC Chen}, \bibinfo{person}{Ellen~C Haas}, {and} \bibinfo{person}{Michael~J Barnes}.} \bibinfo{year}{2007}\natexlab{}.
\newblock \showarticletitle{Human performance issues and user interface design for teleoperated robots}.
\newblock \bibinfo{journal}{\emph{IEEE Transactions on Systems, Man, and Cybernetics, Part C (Applications and Reviews)}} \bibinfo{volume}{37}, \bibinfo{number}{6} (\bibinfo{year}{2007}), \bibinfo{pages}{1231--1245}.
\newblock


\bibitem[Chen et~al\mbox{.}(2018)]%
        {Chen2018}
\bibfield{author}{\bibinfo{person}{Jessie~YC Chen}, \bibinfo{person}{Shan~G Lakhmani}, \bibinfo{person}{Kimberly Stowers}, \bibinfo{person}{Anthony~R Selkowitz}, \bibinfo{person}{Julia~L Wright}, {and} \bibinfo{person}{Michael Barnes}.} \bibinfo{year}{2018}\natexlab{}.
\newblock \showarticletitle{Situation awareness-based agent transparency and human-autonomy teaming effectiveness}.
\newblock \bibinfo{journal}{\emph{Theoretical issues in ergonomics science}} \bibinfo{volume}{19}, \bibinfo{number}{3} (\bibinfo{year}{2018}), \bibinfo{pages}{259--282}.
\newblock


\bibitem[Chen et~al\mbox{.}(2014)]%
        {chen2014situation}
\bibfield{author}{\bibinfo{person}{Jessie~Y Chen}, \bibinfo{person}{Katelyn Procci}, \bibinfo{person}{Michael Boyce}, \bibinfo{person}{Julia Wright}, \bibinfo{person}{Andre Garcia}, {and} \bibinfo{person}{Michael Barnes}.} \bibinfo{year}{2014}\natexlab{}.
\newblock \showarticletitle{Situation awareness-based agent transparency}.
\newblock \bibinfo{journal}{\emph{US Army Research Laboratory}} \bibinfo{number}{April} (\bibinfo{year}{2014}), \bibinfo{pages}{1--29}.
\newblock


\bibitem[Chen et~al\mbox{.}(2011)]%
        {5530413}
\bibfield{author}{\bibinfo{person}{Jessie Y.~C. Chen}, \bibinfo{person}{Michael~J. Barnes}, {and} \bibinfo{person}{Michelle Harper-Sciarini}.} \bibinfo{year}{2011}\natexlab{}.
\newblock \showarticletitle{Supervisory Control of Multiple Robots: Human-Performance Issues and User-Interface Design}.
\newblock \bibinfo{journal}{\emph{IEEE Transactions on Systems, Man, and Cybernetics, Part C (Applications and Reviews)}} \bibinfo{volume}{41}, \bibinfo{number}{4} (\bibinfo{year}{2011}), \bibinfo{pages}{435--454}.
\newblock


\bibitem[Chen et~al\mbox{.}(2022)]%
        {9981165}
\bibfield{author}{\bibinfo{person}{Shengkang Chen}, \bibinfo{person}{Matthew~J. O'Brien}, \bibinfo{person}{Fletcher Talbot}, \bibinfo{person}{Jason Williams}, \bibinfo{person}{Brendan Tidd}, \bibinfo{person}{Alex Pitt}, {and} \bibinfo{person}{Ronald~C. Arkin}.} \bibinfo{year}{2022}\natexlab{}.
\newblock \showarticletitle{Multi-modal User Interface for Multi-robot Control in Underground Environments}. In \bibinfo{booktitle}{\emph{2022 IEEE/RSJ International Conference on Intelligent Robots and Systems (IROS)}}. \bibinfo{pages}{9995--10002}.
\newblock


\bibitem[Chiou et~al\mbox{.}(2021a)]%
        {chiou2021mixed}
\bibfield{author}{\bibinfo{person}{Manolis Chiou}, \bibinfo{person}{Nick Hawes}, {and} \bibinfo{person}{Rustam Stolkin}.} \bibinfo{year}{2021}\natexlab{a}.
\newblock \showarticletitle{Mixed-Initiative variable autonomy for remotely operated mobile robots}.
\newblock \bibinfo{journal}{\emph{ACM Transactions on Human-Robot Interaction (THRI)}} \bibinfo{volume}{10}, \bibinfo{number}{4} (\bibinfo{year}{2021}), \bibinfo{pages}{1--34}.
\newblock


\bibitem[Chiou et~al\mbox{.}(2021b)]%
        {chiou2021trust}
\bibfield{author}{\bibinfo{person}{Manolis Chiou}, \bibinfo{person}{Faye McCabe}, \bibinfo{person}{Markella Grigoriou}, {and} \bibinfo{person}{Rustam Stolkin}.} \bibinfo{year}{2021}\natexlab{b}.
\newblock \showarticletitle{Trust, shared understanding and locus of control in mixed-initiative robotic systems}. In \bibinfo{booktitle}{\emph{2021 30th IEEE International Conference on Robot \& Human Interactive Communication (RO-MAN)}}. IEEE, \bibinfo{pages}{684--691}.
\newblock


\bibitem[Cook et~al\mbox{.}(2015)]%
        {cook2015displays}
\bibfield{author}{\bibinfo{person}{Maia~B Cook}, \bibinfo{person}{Cory~A Rieth}, {and} \bibinfo{person}{Mary~K Ngo}.} \bibinfo{year}{2015}\natexlab{}.
\newblock \showarticletitle{Displays for effective human-agent teaming: the role of information availability and attention management}. In \bibinfo{booktitle}{\emph{Virtual, Augmented and Mixed Reality: 7th International Conference, VAMR 2015, Held as Part of HCI International 2015, Los Angeles, CA, USA, August 2-7, 2015, Proceedings 7}}. Springer, \bibinfo{pages}{174--185}.
\newblock


\bibitem[Cormier et~al\mbox{.}(2015)]%
        {cormier2015situational}
\bibfield{author}{\bibinfo{person}{Olivier St-Martin Cormier}, \bibinfo{person}{Andrew Phan}, {and} \bibinfo{person}{Frank~P Ferrie}.} \bibinfo{year}{2015}\natexlab{}.
\newblock \showarticletitle{Situational awareness for manufacturing applications}. In \bibinfo{booktitle}{\emph{2015 12th Conference on Computer and Robot Vision}}. IEEE, \bibinfo{pages}{320--327}.
\newblock


\bibitem[Dini et~al\mbox{.}(2017)]%
        {dini2017measurement}
\bibfield{author}{\bibinfo{person}{Amir Dini}, \bibinfo{person}{Cornelia Murko}, \bibinfo{person}{Saeed Yahyanejad}, \bibinfo{person}{Ursula Augsd{\"o}rfer}, \bibinfo{person}{Michael Hofbaur}, {and} \bibinfo{person}{Lucas Paletta}.} \bibinfo{year}{2017}\natexlab{}.
\newblock \showarticletitle{Measurement and prediction of situation awareness in human-robot interaction based on a framework of probabilistic attention}. In \bibinfo{booktitle}{\emph{International Conference on Intelligent Robots and Systems}}. IEEE, \bibinfo{pages}{4354--4361}.
\newblock


\bibitem[Domova et~al\mbox{.}(2020)]%
        {domova2020improving}
\bibfield{author}{\bibinfo{person}{Veronika Domova}, \bibinfo{person}{Erik G{\"a}rtner}, \bibinfo{person}{Fredrik Pr{\"a}ntare}, \bibinfo{person}{Martin Pallin}, \bibinfo{person}{Johan K{\"a}llstr{\"o}m}, {and} \bibinfo{person}{Nikita Korzhitskii}.} \bibinfo{year}{2020}\natexlab{}.
\newblock \showarticletitle{Improving Usability of Search and Rescue Decision Support Systems: WARA-PS Case Study}. In \bibinfo{booktitle}{\emph{2020 25th IEEE International Conference on Emerging Technologies and Factory Automation (ETFA)}}, Vol.~\bibinfo{volume}{1}. IEEE, \bibinfo{pages}{1251--1254}.
\newblock


\bibitem[Drury et~al\mbox{.}(2007)]%
        {drury2007lassoing}
\bibfield{author}{\bibinfo{person}{Jill~L Drury}, \bibinfo{person}{Brenden Keyes}, {and} \bibinfo{person}{Holly~A Yanco}.} \bibinfo{year}{2007}\natexlab{}.
\newblock \showarticletitle{LASSOing HRI: analyzing situation awareness in map-centric and video-centric interfaces}. In \bibinfo{booktitle}{\emph{international conference on Human-robot interaction}}. \bibinfo{pages}{279--286}.
\newblock


\bibitem[Drury et~al\mbox{.}(2003)]%
        {drury2003awareness}
\bibfield{author}{\bibinfo{person}{Jill~L Drury}, \bibinfo{person}{Jean Scholtz}, {and} \bibinfo{person}{Holly~A Yanco}.} \bibinfo{year}{2003}\natexlab{}.
\newblock \showarticletitle{Awareness in human-robot interactions}. In \bibinfo{booktitle}{\emph{SMC'03 Conference Proceedings. 2003 IEEE International Conference on Systems, Man and Cybernetics. Conference Theme-System Security and Assurance (Cat. No. 03CH37483)}}, Vol.~\bibinfo{volume}{1}. IEEE, \bibinfo{pages}{912--918}.
\newblock


\bibitem[Endsley(1988a)]%
        {endsley1988design}
\bibfield{author}{\bibinfo{person}{Mica~R Endsley}.} \bibinfo{year}{1988}\natexlab{a}.
\newblock \showarticletitle{Design and evaluation for situation awareness enhancement}. In \bibinfo{booktitle}{\emph{Proceedings of the Human Factors Society annual meeting}}, Vol.~\bibinfo{volume}{32}. Sage Publications Sage CA: Los Angeles, CA, \bibinfo{pages}{97--101}.
\newblock


\bibitem[Endsley(1988b)]%
        {endsley1988situation}
\bibfield{author}{\bibinfo{person}{Mica~R Endsley}.} \bibinfo{year}{1988}\natexlab{b}.
\newblock \showarticletitle{Situation awareness global assessment technique}. In \bibinfo{booktitle}{\emph{National aerospace and electronics conference}}. IEEE.
\newblock


\bibitem[Endsley(1995)]%
        {endsley1995measurement}
\bibfield{author}{\bibinfo{person}{Mica~R Endsley}.} \bibinfo{year}{1995}\natexlab{}.
\newblock \showarticletitle{Measurement of situation awareness in dynamic systems}.
\newblock \bibinfo{journal}{\emph{Human factors}} \bibinfo{volume}{37}, \bibinfo{number}{1} (\bibinfo{year}{1995}), \bibinfo{pages}{65--84}.
\newblock


\bibitem[Endsley and Garland(2000)]%
        {endsley2000situation}
\bibfield{author}{\bibinfo{person}{Mica~R Endsley} {and} \bibinfo{person}{Daniel~J Garland}.} \bibinfo{year}{2000}\natexlab{}.
\newblock \bibinfo{booktitle}{\emph{Situation awareness analysis and measurement}}.
\newblock \bibinfo{publisher}{CRC press}.
\newblock


\bibitem[Fink et~al\mbox{.}(2023)]%
        {fink2023expanded}
\bibfield{author}{\bibinfo{person}{Paul~DS Fink}, \bibinfo{person}{Anas Abou~Allaban}, \bibinfo{person}{Omoruyi~E Atekha}, \bibinfo{person}{Raymond~J Perry}, \bibinfo{person}{Emily~S Sumner}, \bibinfo{person}{Richard~R Corey}, \bibinfo{person}{Velin Dimitrov}, {and} \bibinfo{person}{Nicholas~A Giudice}.} \bibinfo{year}{2023}\natexlab{}.
\newblock \showarticletitle{Expanded Situational Awareness Without Vision: A Novel Haptic Interface for Use in Fully Autonomous Vehicles}. In \bibinfo{booktitle}{\emph{Proceedings of the 2023 ACM/IEEE International Conference on Human-Robot Interaction}}. \bibinfo{pages}{54--62}.
\newblock


\bibitem[Fischer and Lohse(2007)]%
        {fischer2007shaping}
\bibfield{author}{\bibinfo{person}{Kerstin Fischer} {and} \bibinfo{person}{Manja Lohse}.} \bibinfo{year}{2007}\natexlab{}.
\newblock \showarticletitle{Shaping naive users' models of robots' situation awareness}. In \bibinfo{booktitle}{\emph{RO-MAN 2007-The 16th IEEE International Symposium on Robot and Human Interactive Communication}}. IEEE, \bibinfo{pages}{534--539}.
\newblock


\bibitem[Green et~al\mbox{.}(2008)]%
        {green2008human}
\bibfield{author}{\bibinfo{person}{Scott~A Green}, \bibinfo{person}{Mark Billinghurst}, \bibinfo{person}{XiaoQi Chen}, {and} \bibinfo{person}{J~Geoffrey Chase}.} \bibinfo{year}{2008}\natexlab{}.
\newblock \showarticletitle{Human-robot collaboration: A literature review and augmented reality approach in design}.
\newblock \bibinfo{journal}{\emph{International journal of advanced robotic systems}} \bibinfo{volume}{5}, \bibinfo{number}{1} (\bibinfo{year}{2008}), \bibinfo{pages}{1}.
\newblock


\bibitem[Grimm et~al\mbox{.}(2018)]%
        {doi:10.1177/1541931218621034}
\bibfield{author}{\bibinfo{person}{David~A. Grimm}, \bibinfo{person}{Mustafa Demir}, \bibinfo{person}{Jamie~C. Gorman}, {and} \bibinfo{person}{Nancy~J. Cooke}.} \bibinfo{year}{2018}\natexlab{}.
\newblock \showarticletitle{Team Situation Awareness in Human-Autonomy Teaming: A Systems Level Approach}.
\newblock \bibinfo{journal}{\emph{Proceedings of the Human Factors and Ergonomics Society Annual Meeting}} \bibinfo{volume}{62}, \bibinfo{number}{1} (\bibinfo{year}{2018}), \bibinfo{pages}{149--149}.
\newblock


\bibitem[Hong et~al\mbox{.}(2020)]%
        {hong2020effect}
\bibfield{author}{\bibinfo{person}{Zirui Hong}, \bibinfo{person}{Qiqi Zhang}, \bibinfo{person}{Xing Su}, {and} \bibinfo{person}{Hong Zhang}.} \bibinfo{year}{2020}\natexlab{}.
\newblock \showarticletitle{Effect of virtual annotation on performance of construction equipment teleoperation under adverse visual conditions}.
\newblock \bibinfo{journal}{\emph{Automation in Construction}}  \bibinfo{volume}{118} (\bibinfo{year}{2020}), \bibinfo{pages}{103296}.
\newblock


\bibitem[Hopko et~al\mbox{.}(2021)]%
        {hopko2021effect}
\bibfield{author}{\bibinfo{person}{Sarah~K Hopko}, \bibinfo{person}{Riya Khurana}, \bibinfo{person}{Ranjana~K Mehta}, {and} \bibinfo{person}{Prabhakar~R Pagilla}.} \bibinfo{year}{2021}\natexlab{}.
\newblock \showarticletitle{Effect of cognitive fatigue, operator sex, and robot assistance on task performance metrics, workload, and situation awareness in human-robot collaboration}.
\newblock \bibinfo{journal}{\emph{IEEE Robotics and Automation Letters}} \bibinfo{volume}{6}, \bibinfo{number}{2} (\bibinfo{year}{2021}), \bibinfo{pages}{3049--3056}.
\newblock


\bibitem[Huuskonen and Oksanen(2019)]%
        {huuskonen2019augmented}
\bibfield{author}{\bibinfo{person}{Janna Huuskonen} {and} \bibinfo{person}{Timo Oksanen}.} \bibinfo{year}{2019}\natexlab{}.
\newblock \showarticletitle{Augmented reality for supervising multirobot system in agricultural field operation}.
\newblock \bibinfo{journal}{\emph{IFAC-PapersOnLine}} \bibinfo{volume}{52}, \bibinfo{number}{30} (\bibinfo{year}{2019}), \bibinfo{pages}{367--372}.
\newblock


\bibitem[Jalaleddine(2023)]%
        {jalaleddine2023evaluating}
\bibfield{author}{\bibinfo{person}{Reem Jalaleddine}.} \bibinfo{year}{2023}\natexlab{}.
\newblock \emph{\bibinfo{title}{Evaluating the Effect of Automated Driving Systems on Drivers' Visual Attention}}.
\newblock \bibinfo{thesistype}{Master's\ thesis}. \bibinfo{school}{University of Windsor (Canada)}.
\newblock


\bibitem[Kaber et~al\mbox{.}(2013)]%
        {doi:10.1037/h0095998}
\bibfield{author}{\bibinfo{person}{David~B. Kaber}, \bibinfo{person}{Jennifer~M. Riley}, \bibinfo{person}{Mica~R. Endsley}, \bibinfo{person}{Mohamed Sheik-Nainar}, \bibinfo{person}{Tao Zhang}, {and} \bibinfo{person}{Donald~R. Lampton}.} \bibinfo{year}{2013}\natexlab{}.
\newblock \showarticletitle{Measuring Situation Awareness in Virtual Environment-Based Training}.
\newblock \bibinfo{journal}{\emph{Military Psychology}} \bibinfo{volume}{25}, \bibinfo{number}{4} (\bibinfo{year}{2013}).
\newblock


\bibitem[Kanyok et~al\mbox{.}(2022)]%
        {kanyok2022novel}
\bibfield{author}{\bibinfo{person}{Nathan Kanyok}, \bibinfo{person}{Alfred Shaker}, {and} \bibinfo{person}{Jong-Hoon Kim}.} \bibinfo{year}{2022}\natexlab{}.
\newblock \showarticletitle{A Novel Metric of Continuous Situational Awareness Monitoring for Multi-telepresence Coordination System}. In \bibinfo{booktitle}{\emph{Intelligent Human Computer Interaction}}. Springer, \bibinfo{pages}{520--533}.
\newblock


\bibitem[Kidwell et~al\mbox{.}(2012)]%
        {kidwell2012adaptable}
\bibfield{author}{\bibinfo{person}{Brian Kidwell}, \bibinfo{person}{Gloria~L Calhoun}, \bibinfo{person}{Heath~A Ruff}, {and} \bibinfo{person}{Raja Parasuraman}.} \bibinfo{year}{2012}\natexlab{}.
\newblock \showarticletitle{Adaptable and adaptive automation for supervisory control of multiple autonomous vehicles}. In \bibinfo{booktitle}{\emph{Proceedings of the Human Factors and Ergonomics Society Annual Meeting}}, Vol.~\bibinfo{volume}{56}. Sage Publications Sage CA: Los Angeles, CA, \bibinfo{pages}{428--432}.
\newblock


\bibitem[Kunkes et~al\mbox{.}(2022)]%
        {kunkes2022influence}
\bibfield{author}{\bibinfo{person}{Taylor Kunkes}, \bibinfo{person}{Lora Cavuoto}, \bibinfo{person}{Jeff Higginbotham}, \bibinfo{person}{Ann Bisantz}, \bibinfo{person}{Ahmed~S Elsayed}, \bibinfo{person}{Naif~A Aldhaam}, \bibinfo{person}{Ahmed~A Hussein}, {and} \bibinfo{person}{Khurshid~A Guru}.} \bibinfo{year}{2022}\natexlab{}.
\newblock \showarticletitle{Influence of hierarchy on risk communication during robot-assisted surgery: a preliminary study}.
\newblock \bibinfo{journal}{\emph{Surgical Endoscopy}} \bibinfo{volume}{36}, \bibinfo{number}{5} (\bibinfo{year}{2022}), \bibinfo{pages}{3087--3093}.
\newblock


\bibitem[Lijing et~al\mbox{.}(2022)]%
        {lijing2022research}
\bibfield{author}{\bibinfo{person}{Wang Lijing}, \bibinfo{person}{Xiaonan Shi}, \bibinfo{person}{Yu Zhu}, {and} \bibinfo{person}{Yanzeng Zhao}.} \bibinfo{year}{2022}\natexlab{}.
\newblock \showarticletitle{Research on Situational Awareness of Crew Warnings Based on GDTA and FCM}. In \bibinfo{booktitle}{\emph{Man-Machine-Environment System Engineering: Proceedings of the 21st International Conference on MMESE: Commemorative Conference for the 110th Anniversary of Xuesen Qian’s Birth and the 40th Anniversary of Founding of Man-Machine-Environment System Engineering 21}}. Springer, \bibinfo{pages}{260--267}.
\newblock


\bibitem[Lynch et~al\mbox{.}(2023)]%
        {lynch2023maritime}
\bibfield{author}{\bibinfo{person}{Kirsty~M Lynch}, \bibinfo{person}{Victoria~A Banks}, \bibinfo{person}{Aaron~PJ Roberts}, \bibinfo{person}{Stewart Radcliffe}, {and} \bibinfo{person}{Katherine~L Plant}.} \bibinfo{year}{2023}\natexlab{}.
\newblock \showarticletitle{Maritime autonomous surface ships: can we learn from unmanned aerial vehicle incidents using the perceptual cycle model?}
\newblock \bibinfo{journal}{\emph{Ergonomics}} \bibinfo{volume}{66}, \bibinfo{number}{6} (\bibinfo{year}{2023}), \bibinfo{pages}{772--790}.
\newblock


\bibitem[Maddikunta et~al\mbox{.}(2022)]%
        {MADDIKUNTA2022100257}
\bibfield{author}{\bibinfo{person}{Praveen Kumar~Reddy Maddikunta}, \bibinfo{person}{Quoc-Viet Pham}, \bibinfo{person}{Prabadevi B}, \bibinfo{person}{N Deepa}, \bibinfo{person}{Kapal Dev}, \bibinfo{person}{Thippa~Reddy Gadekallu}, \bibinfo{person}{Rukhsana Ruby}, {and} \bibinfo{person}{Madhusanka Liyanage}.} \bibinfo{year}{2022}\natexlab{}.
\newblock \showarticletitle{Industry 5.0: A survey on enabling technologies and potential applications}.
\newblock \bibinfo{journal}{\emph{Journal of Industrial Information Integration}}  \bibinfo{volume}{26} (\bibinfo{year}{2022}), \bibinfo{pages}{100257}.
\newblock
\showISSN{2452-414X}


\bibitem[Mansikka et~al\mbox{.}(2021)]%
        {MANSIKKA2021103473}
\bibfield{author}{\bibinfo{person}{Heikki Mansikka}, \bibinfo{person}{Kai Virtanen}, \bibinfo{person}{Ville Uggeldahl}, {and} \bibinfo{person}{Don Harris}.} \bibinfo{year}{2021}\natexlab{}.
\newblock \showarticletitle{Team situation awareness accuracy measurement technique for simulated air combat - Curvilinear relationship between awareness and performance}.
\newblock \bibinfo{journal}{\emph{Applied Ergonomics}}  \bibinfo{volume}{96} (\bibinfo{year}{2021}), \bibinfo{pages}{103473}.
\newblock
\showISSN{0003-6870}


\bibitem[Matthews et~al\mbox{.}(2021)]%
        {MATTHEWS202159}
\bibfield{author}{\bibinfo{person}{Gerald Matthews}, \bibinfo{person}{April~Rose Panganiban}, \bibinfo{person}{Jinchao Lin}, \bibinfo{person}{Michael Long}, {and} \bibinfo{person}{Michaela Schwing}.} \bibinfo{year}{2021}\natexlab{}.
\newblock \showarticletitle{Chapter 3 - Super-machines or sub-humans: Mental models and trust in intelligent autonomous systems}.
\newblock In \bibinfo{booktitle}{\emph{Trust in Human-Robot Interaction}}, \bibfield{editor}{\bibinfo{person}{Chang~S. Nam} {and} \bibinfo{person}{Joseph~B. Lyons}} (Eds.). \bibinfo{publisher}{Academic Press}, \bibinfo{pages}{59--82}.
\newblock
\showISBNx{978-0-12-819472-0}


\bibitem[Methnani et~al\mbox{.}(2024)]%
        {10.1145/3645090}
\bibfield{author}{\bibinfo{person}{Leila Methnani}, \bibinfo{person}{Manolis Chiou}, \bibinfo{person}{Virginia Dignum}, {and} \bibinfo{person}{Andreas Theodorou}.} \bibinfo{year}{2024}\natexlab{}.
\newblock \showarticletitle{Who’s in Charge Here? A Survey on Trustworthy AI in Variable Autonomy Robotic Systems}.
\newblock \bibinfo{journal}{\emph{ACM Comput. Surv.}} \bibinfo{volume}{56}, \bibinfo{number}{7}, Article \bibinfo{articleno}{184} (\bibinfo{date}{April} \bibinfo{year}{2024}), \bibinfo{numpages}{32}~pages.
\newblock
\showISSN{0360-0300}


\bibitem[M{\"u}ller et~al\mbox{.}(2023)]%
        {muller2023self}
\bibfield{author}{\bibinfo{person}{Manuel M{\"u}ller}, \bibinfo{person}{Tam{\'a}s Ruppert}, \bibinfo{person}{Nasser Jazdi}, {and} \bibinfo{person}{Michael Weyrich}.} \bibinfo{year}{2023}\natexlab{}.
\newblock \showarticletitle{Self-improving situation awareness for human--robot-collaboration using intelligent Digital Twin}.
\newblock \bibinfo{journal}{\emph{Journal of Intelligent Manufacturing}} (\bibinfo{year}{2023}), \bibinfo{pages}{1--19}.
\newblock


\bibitem[Nakayama(1999)]%
        {nakayama1999inattentional}
\bibfield{author}{\bibinfo{person}{Ken Nakayama}.} \bibinfo{year}{1999}\natexlab{}.
\newblock \showarticletitle{Inattentional Blindness: by Arien Mack and Irvine Rock}.
\newblock \bibinfo{journal}{\emph{Trends in Cognitive Sciences}} \bibinfo{volume}{3}, \bibinfo{number}{1} (\bibinfo{year}{1999}), \bibinfo{pages}{39}.
\newblock


\bibitem[Norman(1986)]%
        {norman1986cognitive}
\bibfield{author}{\bibinfo{person}{Donald~A Norman}.} \bibinfo{year}{1986}\natexlab{}.
\newblock \showarticletitle{Cognitive engineering}.
\newblock \bibinfo{journal}{\emph{User centered system design}} \bibinfo{volume}{31}, \bibinfo{number}{61} (\bibinfo{year}{1986}), \bibinfo{pages}{2}.
\newblock


\bibitem[Norton et~al\mbox{.}(2017)]%
        {doi:10.1177/0278364916688254}
\bibfield{author}{\bibinfo{person}{Adam Norton}, \bibinfo{person}{Willard Ober}, \bibinfo{person}{Lisa Baraniecki}, \bibinfo{person}{Eric McCann}, \bibinfo{person}{Jean Scholtz}, \bibinfo{person}{David Shane}, \bibinfo{person}{Anna Skinner}, \bibinfo{person}{Robert Watson}, {and} \bibinfo{person}{Holly Yanco}.} \bibinfo{year}{2017}\natexlab{}.
\newblock \showarticletitle{Analysis of human–robot interaction at the DARPA Robotics Challenge Finals}.
\newblock \bibinfo{journal}{\emph{The International Journal of Robotics Research}} \bibinfo{volume}{36}, \bibinfo{number}{5-7} (\bibinfo{year}{2017}), \bibinfo{pages}{483--513}.
\newblock


\bibitem[Nowell et~al\mbox{.}(2017)]%
        {nowell2017thematic}
\bibfield{author}{\bibinfo{person}{Lorelli~S Nowell}, \bibinfo{person}{Jill~M Norris}, \bibinfo{person}{Deborah~E White}, {and} \bibinfo{person}{Nancy~J Moules}.} \bibinfo{year}{2017}\natexlab{}.
\newblock \showarticletitle{Thematic analysis: Striving to meet the trustworthiness criteria}.
\newblock \bibinfo{journal}{\emph{International journal of qualitative methods}} \bibinfo{volume}{16}, \bibinfo{number}{1} (\bibinfo{year}{2017}), \bibinfo{pages}{1609406917733847}.
\newblock


\bibitem[Olsen and Wood(2004)]%
        {10.1145/985692.985722}
\bibfield{author}{\bibinfo{person}{Dan~R. Olsen} {and} \bibinfo{person}{Stephen~Bart Wood}.} \bibinfo{year}{2004}\natexlab{}.
\newblock \showarticletitle{Fan-out: measuring human control of multiple robots}. In \bibinfo{booktitle}{\emph{Proceedings of the SIGCHI Conference on Human Factors in Computing Systems}} (Vienna, Austria) \emph{(\bibinfo{series}{CHI '04})}. \bibinfo{publisher}{Association for Computing Machinery}, \bibinfo{address}{New York, NY, USA}, \bibinfo{pages}{231–238}.
\newblock
\showISBNx{1581137028}


\bibitem[Onnasch et~al\mbox{.}(2014)]%
        {onnasch2014human}
\bibfield{author}{\bibinfo{person}{Linda Onnasch}, \bibinfo{person}{Christopher~D Wickens}, \bibinfo{person}{Huiyang Li}, {and} \bibinfo{person}{Dietrich Manzey}.} \bibinfo{year}{2014}\natexlab{}.
\newblock \showarticletitle{Human performance consequences of stages and levels of automation: An integrated meta-analysis}.
\newblock \bibinfo{journal}{\emph{Human factors}} \bibinfo{volume}{56}, \bibinfo{number}{3} (\bibinfo{year}{2014}), \bibinfo{pages}{476--488}.
\newblock


\bibitem[Paris and Reeson(2024)]%
        {paris2021s}
\bibfield{author}{\bibinfo{person}{Cécile Paris} {and} \bibinfo{person}{Andrew Reeson}.} \bibinfo{year}{2024}\natexlab{}.
\newblock \showarticletitle{What’s the secret to making sure AI doesn’t steal your job? Work with it, not against it}.
\newblock \bibinfo{journal}{\emph{The Conversation on Work}} (\bibinfo{year}{2024}).
\newblock
\urldef\tempurl%
\url{https://theconversation.com/whats-the-secret-to-making-sure-ai-doesnt-steal-your-job-work-with-it-not-against-it-172691}
\showURL{%
\tempurl}
\newblock
\shownote{Article first published in The Conversation in 2021}.


\bibitem[Petersen(2014)]%
        {petersen2014general}
\bibfield{author}{\bibinfo{person}{Karen Petersen}.} \bibinfo{year}{2014}\natexlab{}.
\newblock \showarticletitle{General concepts for human supervision of autonomous robot teams}.
\newblock  (\bibinfo{year}{2014}).
\newblock


\bibitem[Pitman et~al\mbox{.}(2007)]%
        {pitman2007picture}
\bibfield{author}{\bibinfo{person}{A Pitman}, \bibinfo{person}{Curtis~M Humphrey}, {and} \bibinfo{person}{Julie~A Adams}.} \bibinfo{year}{2007}\natexlab{}.
\newblock \bibinfo{title}{A picture-in-picture interface for a multiple robot system}.
\newblock
\newblock


\bibitem[Ratwani et~al\mbox{.}(2010)]%
        {ratwani2010single}
\bibfield{author}{\bibinfo{person}{Raj Ratwani}, \bibinfo{person}{J~Malcolm McCurry}, {and} \bibinfo{person}{J~Gregory Trafton}.} \bibinfo{year}{2010}\natexlab{}.
\newblock \showarticletitle{Single operator, multiple robots: an eye movement based theoretic model of operator situation awareness}. In \bibinfo{booktitle}{\emph{International Conference on Human-Robot Interaction}}. IEEE, \bibinfo{pages}{235--242}.
\newblock


\bibitem[Rea and Seo(2022)]%
        {10.3389/frobt.2022.704225}
\bibfield{author}{\bibinfo{person}{Daniel~J. Rea} {and} \bibinfo{person}{Stela~H. Seo}.} \bibinfo{year}{2022}\natexlab{}.
\newblock \showarticletitle{Still Not Solved: A Call for Renewed Focus on User-Centered Teleoperation Interfaces}.
\newblock \bibinfo{journal}{\emph{Frontiers in Robotics and AI}}  \bibinfo{volume}{9} (\bibinfo{year}{2022}).
\newblock
\showISSN{2296-9144}


\bibitem[Riek(2012)]%
        {riek2012wizard}
\bibfield{author}{\bibinfo{person}{Laurel~D Riek}.} \bibinfo{year}{2012}\natexlab{}.
\newblock \showarticletitle{Wizard of oz studies in hri: a systematic review and new reporting guidelines}.
\newblock \bibinfo{journal}{\emph{Journal of Human-Robot Interaction}} \bibinfo{volume}{1}, \bibinfo{number}{1} (\bibinfo{year}{2012}), \bibinfo{pages}{119--136}.
\newblock


\bibitem[Riley et~al\mbox{.}(2010)]%
        {riley2010situation}
\bibfield{author}{\bibinfo{person}{Jennifer~M Riley}, \bibinfo{person}{Laura~D Strater}, \bibinfo{person}{Sheryl~L Chappell}, \bibinfo{person}{Erik~S Connors}, {and} \bibinfo{person}{Mica~R Endsley}.} \bibinfo{year}{2010}\natexlab{}.
\newblock \showarticletitle{Situation awareness in human-robot interaction: Challenges and user interface requirements}.
\newblock \bibinfo{journal}{\emph{Human-Robot Interactions in Future Military Operations}} (\bibinfo{year}{2010}), \bibinfo{pages}{171--192}.
\newblock


\bibitem[Rold{\'a}n~G{\'o}mez(2018)]%
        {roldan2018adaptive}
\bibfield{author}{\bibinfo{person}{Juan~Jes{\'u}s Rold{\'a}n~G{\'o}mez}.} \bibinfo{year}{2018}\natexlab{}.
\newblock \emph{\bibinfo{title}{Adaptive and immersive interfaces to improve situational awareness in multi-robot missions}}.
\newblock \bibinfo{thesistype}{Ph.\,D. Dissertation}. \bibinfo{school}{Industriales}.
\newblock


\bibitem[Salmon et~al\mbox{.}(2008a)]%
        {salmon2008really}
\bibfield{author}{\bibinfo{person}{Paul~M Salmon}, \bibinfo{person}{Neville~A Stanton}, \bibinfo{person}{Guy~H Walker}, \bibinfo{person}{Chris Baber}, \bibinfo{person}{Daniel~P Jenkins}, \bibinfo{person}{Richard McMaster}, {and} \bibinfo{person}{Mark~S Young}.} \bibinfo{year}{2008}\natexlab{a}.
\newblock \showarticletitle{What really is going on? Review of situation awareness models for individuals and teams}.
\newblock \bibinfo{journal}{\emph{Theoretical Issues in Ergonomics Science}} \bibinfo{volume}{9}, \bibinfo{number}{4} (\bibinfo{year}{2008}), \bibinfo{pages}{297--323}.
\newblock


\bibitem[Salmon et~al\mbox{.}(2008b)]%
        {salmon2008representing}
\bibfield{author}{\bibinfo{person}{Paul~M Salmon}, \bibinfo{person}{Neville~A Stanton}, \bibinfo{person}{Guy~H Walker}, \bibinfo{person}{Daniel Jenkins}, \bibinfo{person}{Christopher Baber}, {and} \bibinfo{person}{Richard McMaster}.} \bibinfo{year}{2008}\natexlab{b}.
\newblock \showarticletitle{Representing situation awareness in collaborative systems: A case study in the energy distribution domain}.
\newblock \bibinfo{journal}{\emph{Ergonomics}} \bibinfo{volume}{51}, \bibinfo{number}{3} (\bibinfo{year}{2008}), \bibinfo{pages}{367--384}.
\newblock


\bibitem[Schleiger et~al\mbox{.}(2023)]%
        {schleiger2023collaborative}
\bibfield{author}{\bibinfo{person}{Emma Schleiger}, \bibinfo{person}{Claire Mason}, \bibinfo{person}{Claire Naughtin}, \bibinfo{person}{Andrew Reeson}, {and} \bibinfo{person}{Cecile Paris}.} \bibinfo{year}{2023}\natexlab{}.
\newblock \showarticletitle{Collaborative Intelligence: A scoping review of current applications}.
\newblock \bibinfo{journal}{\emph{Qeios}} (\bibinfo{year}{2023}).
\newblock


\bibitem[Schuster and Jentsch(2011)]%
        {schuster2011measurement}
\bibfield{author}{\bibinfo{person}{David Schuster} {and} \bibinfo{person}{Florian Jentsch}.} \bibinfo{year}{2011}\natexlab{}.
\newblock \showarticletitle{Measurement of situation awareness in human-robot teams}. In \bibinfo{booktitle}{\emph{Proceedings of the Human Factors and Ergonomics Society Annual Meeting}}, Vol.~\bibinfo{volume}{55}. SAGE Publications Sage CA: Los Angeles, CA, \bibinfo{pages}{1496--1500}.
\newblock


\bibitem[Sen et~al\mbox{.}(2020)]%
        {sen2020effects}
\bibfield{author}{\bibinfo{person}{Sumona Sen}, \bibinfo{person}{Hans-J{\"u}rgen Buxbaum}, {and} \bibinfo{person}{Lisanne Kremer}.} \bibinfo{year}{2020}\natexlab{}.
\newblock \showarticletitle{The Effects of Different Robot Trajectories on Situational Awareness in Human-Robot Collaboration}. In \bibinfo{booktitle}{\emph{Human-Computer Interaction. Multimodal and Natural Interaction: Thematic Area, HCI 2020, Held as Part of the 22nd International Conference, HCII 2020, Copenhagen, Denmark, July 19--24, 2020, Proceedings, Part II 22}}. Springer, \bibinfo{pages}{719--729}.
\newblock


\bibitem[Senaratne et~al\mbox{.}(2023)]%
        {senaratne2023roman}
\bibfield{author}{\bibinfo{person}{Hashini Senaratne}, \bibinfo{person}{Alex Pitt}, \bibinfo{person}{Fletcher Talbot}, \bibinfo{person}{Peyman Moghadam}, \bibinfo{person}{Pavan Sikka}, \bibinfo{person}{David Howard}, \bibinfo{person}{Jason Williams}, \bibinfo{person}{Dana Kulić}, {and} \bibinfo{person}{Cécile Paris}.} \bibinfo{year}{2023}\natexlab{}.
\newblock \showarticletitle{Measuring Situational Awareness Latency in Human-Robot Teaming Experiments}. In \bibinfo{booktitle}{\emph{2023 32nd IEEE International Conference on Robot and Human Interactive Communication (RO-MAN)}}. \bibinfo{pages}{2624--2631}.
\newblock


\bibitem[Shayesteh and Jebelli(2022)]%
        {shayesteh2022enhanced}
\bibfield{author}{\bibinfo{person}{Shayan Shayesteh} {and} \bibinfo{person}{Houtan Jebelli}.} \bibinfo{year}{2022}\natexlab{}.
\newblock \showarticletitle{Enhanced Situational Awareness in Worker-Robot Interaction in Construction: Assessing the Role of Visual Cues}. In \bibinfo{booktitle}{\emph{Construction Research Congress 2022}}. \bibinfo{pages}{422--430}.
\newblock


\bibitem[Simons and Chabris(1999)]%
        {simons1999gorillas}
\bibfield{author}{\bibinfo{person}{Daniel~J Simons} {and} \bibinfo{person}{Christopher~F Chabris}.} \bibinfo{year}{1999}\natexlab{}.
\newblock \showarticletitle{Gorillas in our midst: Sustained inattentional blindness for dynamic events}.
\newblock \bibinfo{journal}{\emph{perception}} \bibinfo{volume}{28}, \bibinfo{number}{9} (\bibinfo{year}{1999}), \bibinfo{pages}{1059--1074}.
\newblock


\bibitem[Smith and Hancock(1995)]%
        {smith1995situation}
\bibfield{author}{\bibinfo{person}{Kip Smith} {and} \bibinfo{person}{Peter~A Hancock}.} \bibinfo{year}{1995}\natexlab{}.
\newblock \showarticletitle{Situation awareness is adaptive, externally directed consciousness}.
\newblock \bibinfo{journal}{\emph{Human factors}} \bibinfo{volume}{37}, \bibinfo{number}{1} (\bibinfo{year}{1995}), \bibinfo{pages}{137--148}.
\newblock


\bibitem[Stanton et~al\mbox{.}(2001)]%
        {STANTON2001189}
\bibfield{author}{\bibinfo{person}{N.A Stanton}, \bibinfo{person}{P.R.G Chambers}, {and} \bibinfo{person}{J Piggott}.} \bibinfo{year}{2001}\natexlab{}.
\newblock \showarticletitle{Situational awareness and safety}.
\newblock \bibinfo{journal}{\emph{Safety Science}} \bibinfo{volume}{39}, \bibinfo{number}{3} (\bibinfo{year}{2001}), \bibinfo{pages}{189--204}.
\newblock
\showISSN{0925-7535}


\bibitem[Sumigray et~al\mbox{.}(2021)]%
        {sumigray2021improving}
\bibfield{author}{\bibinfo{person}{Austin Sumigray}, \bibinfo{person}{Eliot Laidlaw}, \bibinfo{person}{James Tompkin}, {and} \bibinfo{person}{Stefanie Tellex}.} \bibinfo{year}{2021}\natexlab{}.
\newblock \showarticletitle{Improving Remote Environment Visualization through 360 6DoF Multi-sensor Fusion for VR Telerobotics}. In \bibinfo{booktitle}{\emph{Companion of the 2021 ACM/IEEE International Conference on Human-Robot Interaction}}. \bibinfo{pages}{387--391}.
\newblock


\bibitem[Szafir and Szafir(2021)]%
        {Daniel2021HRI}
\bibfield{author}{\bibinfo{person}{Daniel Szafir} {and} \bibinfo{person}{Danielle~Albers Szafir}.} \bibinfo{year}{2021}\natexlab{}.
\newblock \showarticletitle{Connecting Human-Robot Interaction and Data Visualization}. In \bibinfo{booktitle}{\emph{Proceedings of the 2021 ACM/IEEE International Conference on Human-Robot Interaction}} (Boulder, CO, USA) \emph{(\bibinfo{series}{HRI '21})}. \bibinfo{publisher}{Association for Computing Machinery}, \bibinfo{address}{New York, NY, USA}, \bibinfo{pages}{281–292}.
\newblock
\showISBNx{9781450382892}


\bibitem[Tanke(2013)]%
        {tanke2013improving}
\bibfield{author}{\bibinfo{person}{Niels Tanke}.} \bibinfo{year}{2013}\natexlab{}.
\newblock \showarticletitle{Improving Situational Awareness during Human-Robot Interaction using a Head-Mounted Display}.
\newblock  (\bibinfo{year}{2013}).
\newblock


\bibitem[Thomas and Kellgren(2017)]%
        {thomas2017benner}
\bibfield{author}{\bibinfo{person}{Christine~M Thomas} {and} \bibinfo{person}{Molly Kellgren}.} \bibinfo{year}{2017}\natexlab{}.
\newblock \showarticletitle{Benner’s novice to expert model: An application for simulation facilitators}.
\newblock \bibinfo{journal}{\emph{Nursing science quarterly}} \bibinfo{volume}{30}, \bibinfo{number}{3} (\bibinfo{year}{2017}), \bibinfo{pages}{227--234}.
\newblock


\bibitem[Wickens and Alexander(2009)]%
        {wickens2009attentional}
\bibfield{author}{\bibinfo{person}{Christopher~D Wickens} {and} \bibinfo{person}{Amy~L Alexander}.} \bibinfo{year}{2009}\natexlab{}.
\newblock \showarticletitle{Attentional tunneling and task management in synthetic vision displays}.
\newblock \bibinfo{journal}{\emph{The international journal of aviation psychology}} \bibinfo{volume}{19}, \bibinfo{number}{2} (\bibinfo{year}{2009}), \bibinfo{pages}{182--199}.
\newblock


\bibitem[Yanco et~al\mbox{.}(2004)]%
        {yanco2004beyond}
\bibfield{author}{\bibinfo{person}{Holly~A Yanco}, \bibinfo{person}{Jill~L Drury}, {and} \bibinfo{person}{Jean Scholtz}.} \bibinfo{year}{2004}\natexlab{}.
\newblock \showarticletitle{Beyond usability evaluation: Analysis of human-robot interaction at a major robotics competition}.
\newblock \bibinfo{journal}{\emph{Human--Computer Interaction}} \bibinfo{volume}{19}, \bibinfo{number}{1-2} (\bibinfo{year}{2004}), \bibinfo{pages}{117--149}.
\newblock


\bibitem[Yonga~Chuengwa et~al\mbox{.}(2023)]%
        {yonga2023research}
\bibfield{author}{\bibinfo{person}{Thierry Yonga~Chuengwa}, \bibinfo{person}{Jan~Adriaan Swanepoel}, \bibinfo{person}{Anish~Matthew Kurien}, \bibinfo{person}{Mukondeleli~Grace Kanakana-Katumba}, {and} \bibinfo{person}{Karim Djouani}.} \bibinfo{year}{2023}\natexlab{}.
\newblock \showarticletitle{Research Perspectives in Collaborative Assembly: A Review}.
\newblock \bibinfo{journal}{\emph{Robotics}} \bibinfo{volume}{12}, \bibinfo{number}{2} (\bibinfo{year}{2023}), \bibinfo{pages}{37}.
\newblock


\end{thebibliography}

\newpage
\appendix

\section{Appendix: Questionnaire and Interview Questions}

\subsection{Online Questionnaire}\label{sec:questionnaire}

\begin{enumerate}
    \item Demographics:
    \begin{enumerate}
    \item Participant ID:
    \item What does best describe your gender?\\
        (Women, Man, Non-binary/ gender diverse, Not listed and I identify as:<>, Prefer not to say)
    \item What is your age?
    \item What is your profession?
    \end{enumerate}
    
    \item Human-Robot Team Experience:
    \begin{enumerate}
    \item Provide a brief description of human-robot team applications that you have worked with:
    \item For how long have you used human-robot interfaces? (in days, weeks, months or years)
    \item How frequently have you used human-robot interfaces? (x per year)
    \item Where are you on a scale from a novice to an expert in performing the above role in that application? (Novice, Advanced Beginner, Competent, Proficient, Expert)
    \end{enumerate}
\end{enumerate}

\subsection{Questions Used to Guide Interviews}\label{sec:interviewQuestions}

\begin{enumerate}
    
   \item Human-Robot Team Context:\\
   In your human-robot team application(s),
   \begin{enumerate}
     \item What is the \textbf{team mission}?
     \item What is the \textbf{team structure}? 
     \item Which tasks are \textbf{performed well by different robots}?
     \item Which tasks do you think \textbf{you are good at performing}? 
     \item Why do you think those tasks cannot be carried out by a robot or an AI agent? 
   \end{enumerate}
   
   \item Challenges and Failures:\\
   Considering your overall time using human-robot team applications,
   \begin{enumerate}
     \item Have you experienced situations where you \textbf{failed or found it challenging to perform a required action on time}? 
     \begin{enumerate}
     \item Can you share a \textbf{couple of examples}?
     \item \textbf{Why} do you think you could not act on time? 
     \item What were the \textbf{consequences} of such failures or delays in performing actions?
     \end{enumerate}
     
     \item Have you experienced situations where you \textbf{performed actions that were not necessary}?
     \begin{enumerate}
     \item Can you share a \textbf{couple of examples}?
     \item \textbf{Why} do you think that happened?
     \item What were the \textbf{consequences} of such unnecessary actions?
     \end{enumerate}
     
     \item Have you experienced situations where you \textbf{missed, misinterpreted or were confused} about information presented? 
     \begin{enumerate}
     \item Can you share a \textbf{couple of examples}?
     \item \textbf{Why} do you think you that happened?
     \item What were the \textbf{consequences}?
     \end{enumerate}
     
     \item Have you experienced situations where you \textbf{were presented with information that were not necessary}?
     \begin{enumerate}
     \item Can you share a \textbf{couple of examples}?
     \item \textbf{Why} do you think that the system presented with such unnecessary information?
     \item What were the \textbf{consequences} of seeing such unnecessary information?
     \end{enumerate}
     
     \end{enumerate}
     
  \item Suitable Level of Situational Awareness:\\
  <Given a definition of situational awareness,>
   \begin{enumerate}
     \item Do you think that you need to maintain the same level of situational awareness through out the mission context? 
        \begin{enumerate}
        \item If yes: Explain why.
        \item If no: In your application, when do you think that you need a higher-level of situational awareness and a lower-level of situational awareness?
     \end{enumerate}
     \item Let's consider $<$x scenario - identified as requiring higher level of situational awareness or challenging to act on time$>$. Which information do you think you need to perceive during x?
     \item During $<$x$>$, how do you think that you need to synthesis perceived information?
     \item During $<$x$>$, hat projections do you think that you need to make?
     \item Which \textbf{strategies} do you use to maintain a suitable level of situational awareness with the changes in mission context? \textit{(explore both areas: when taxed with multiple competing demands and when everything goes smooth autonomously)}
     \end{enumerate}
     
  \item Design Thoughts for Adaptations:
   \begin{enumerate}
     \item \textbf{How well the user-interface of your application supports} to maintain a suitable level of situational awareness?
     \item Do you have ideas for \textbf{how the user interface of your application could support improving awareness}? 
     \item Do you have ideas for \textbf{how the the underlying AI capabilities could support improving awareness}?
     \item In the event of high workload or incapacitation, do you \textbf{prefer an AI agent to take over some of your tasks} within their capacity? If yes: 
     \begin{enumerate}
     \item What kind of tasks you prefer to take over? 
     \item How might the other tasks be handled?
     \end{enumerate}
     \item Imagine that you can maintain a \textbf{collaborative dialogue} with the robots or an AI agent integrated to the system. 
     \begin{enumerate}
     \item What kind of questions that you would like to ask to maintain a suitable level of situational awareness?
     \item What kind of information that you think robots can benefit from asking you, to improve the collaboration?
     \end{enumerate}
     \end{enumerate}

 \end{enumerate}

\newpage
\section{Appendix: Demographic Details of Participants and Background Details of Human-Robot Teaming Applications}\label{appendixB}

\begin{table}[h!]
    \centering\small
    \begin{tabular}{|l|c||l|c|c|}
        \hline
         & n &  & Mean & Standard deviation\\
 \hline
        Gender  &  & Age & 34.59  & 7.10 \\
        \hspace{15pt} Male  & 15   & Years of experience with their HRT & 3.27 &  2.70\\
        \hspace{15pt} Female  & 1   & Frequency of usage of HRT per year    & 75 &  87.97\\

        Highest level of education &  &  & & \\
        \hspace{15pt} High school diploma & 2 & & & \\
        \hspace{15pt} Bachelor's degree & 7 & & & \\
        \hspace{15pt} Master's degree & 1 & & & \\
        \hspace{15pt} Doctoral degree & 7 & & &\\
        \hline
    \end{tabular}
    \caption{Participants' Demographics and Human-Robot Teaming Application Related Details} 
    \label{tab:demographics}
    \vspace{-16pt}
\end{table}

\subsection{Search and Rescue Application 1}

\begin{table}[h!]
    \centering\small
    \begin{tabular} {|p{.15\linewidth}|p{.75\linewidth}|} \hline    
        Participants & P1-3, P5 \\
        \hline

         Environment & A physical made-up disaster site with underground tunnels and urban areas.\\
         \hline

         HRT Interface & A visual interface with a view to verify detected objects and a map view with robot status panels, which also facilitates high-level waypoint -like commands. \par A joysick controller to teleoperate the robot. \\
         \hline

         Team Mission & Perform a human-in-the-loop exploration of an unknown disaster site to search for signs of survivors and detect and localize them. \\
         \hline

         {\# of Humans} & 1 collaborator \\
         \hline
         
         {\# of Robots} & 1 - 6 (1-2 quadrupeds, 1-2 all-terrain robots, 1-2 drones) \\
         \hline
         
         {Role of Human} & Monitor robots through a remote interface and provide high-level guidance by issuing waypoints or allocating tasks, or low-level guidance through teleoperation, as required. Verify signs of casualties detected by robots. Initiate dropping WiFi nodes carried on all-terrain robots to maintain robots' connectivity to base station. \\
         \hline
         
         {Role of Robot} & Explore the map by navigating the disaster site. Detect signs of casualties and report back to the human. Operate under the selected level of autonomy: teleoperation (low), follow waypoints / prioritise or deprioritise an area (mid), act autonomously (high).\\
         \hline
    \end{tabular}
    \caption[]{Environment, Interface, Mission, Team  Structure and Role Details of Search and Rescue Application 1}
    \label{tab:sr1details} 
\end{table}

\begin{figure}[h!]
  \centering
  \includegraphics[width=\linewidth]{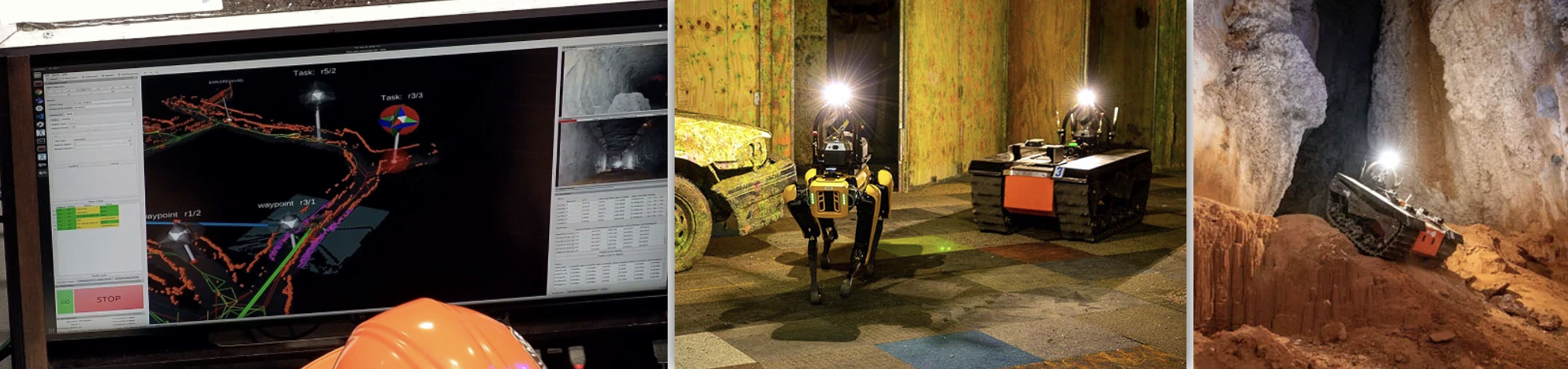}
  \caption{Search \& Rescue Application 1: user interface and robots in the field}
  \label{fig:rescue_real}
      \vspace{-10pt}
\end{figure}

\clearpage

\subsection{Search and Rescue Application 2}

\begin{table}[h!]
    \centering\small
    \begin{tabular} {|p{.15\linewidth}|p{.75\linewidth}|} \hline
    
        Participants & P12-13 \\
        \hline

         Environment & A virtual area simulating a disaster. \\
         \hline

         HRT Interface & A graphical simulation interface with a map view, camera view, and robot's status, which also facilitates waypoint-like high-level controls. \par Audio alters, informing autonomy level change. \par A joystick controller to teleoperate the robot. \\
         \hline

         Team Mission & Navigate the disaster area, explore and find the victims as fast as possible while avoiding collisions.\\
         \hline

         {\# of Humans} & 1 collaborator \\
         \hline
         
         {\# of Robots} & 1 \\
         \hline
         
         {Role of Human} &  See for marks of casualties, and note down where a victim  is located. Make sure the robot is not going over the casualties. Give or take control from robot by judging the robot's performance. \\
         \hline
         
         {Role of Robot} & Explore the disaster site by navigating under the selected level of autonomy: teleportation with (high) or  without (low) autonomous collision avoidance. Give or take control from the human by evaluating whether the task is performed close to the ideal. \\
         \hline
    \end{tabular}
    \caption[]{Environment, Interface, Mission, Team  Structure and Role Details of Search and Rescue Application 2}
    \label{tab:sr2details} 
\end{table}

\begin{figure}[h!]
  \centering
  \includegraphics[width=\linewidth]{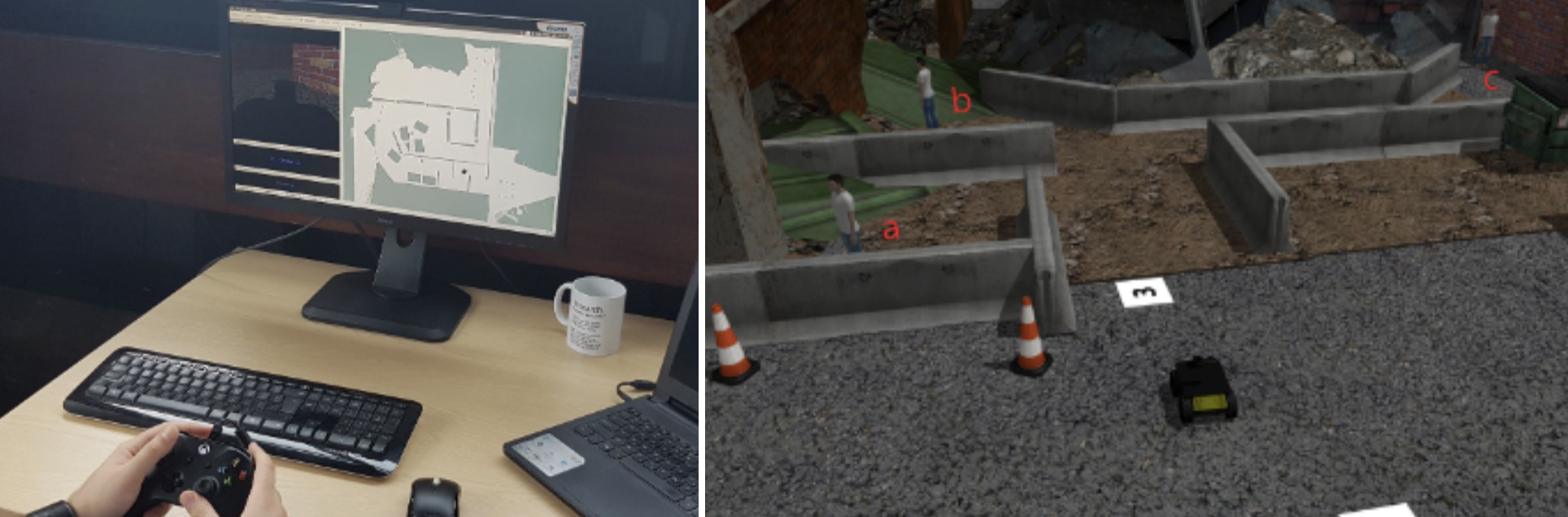}
  \caption{Search \& Rescue Application 2: user interface with simulated environment}
  \label{fig:rescue_simulated}
\end{figure}

\clearpage

\subsection{Nuclear Disaster Response Application}

\begin{table}[h!]
    \centering\small
    \begin{tabular} {|p{.15\linewidth}|p{.75\linewidth}|} \hline
    
        Participants & P12-13 \\
        \hline

         Environment & A physical made-up nuclear disaster site. \\
         \hline

         HRT Interface & A visual user interface with integrated physical joysticks and designated buttons for controlling robot and arm movements. \par A big screen showing the video feeds from the robot's onboard cameras and the secondary robot. \\
         \hline

         Team Mission & Use a robot to navigate a nuclear disaster site and locate and dispose of a contaminated piece.\\
         \hline

         {\# of Humans} & 2 collaborators (main operator, second operator)\\
         \hline
         
         {\# of Robots} & 2 (large track robot with an arm and a gripper, small track robot) \\
         \hline
         
         {Role of Human} &  Main operator: teleoperate the large robot to open a door with a lock and enter a nuclear disaster site, locate a contaminated piece on the floor, pick it and place it  inside a container.  \par  Second operator: teleoperate  the small robot to follow the large robot to get a secondary  view of the scene to the  main operator.\\
         \hline
         
         {Role of Robot} & No automated functions. Navigate to risky area and secure a contaminated item as instructed by the main  human operator.\\
         \hline
\end{tabular}
    \caption[]{Environment, Interface, Mission, Team  Structure and Role Details of Nuclear Disaster Response Application}
    \label{tab:nucleardetails} 
\end{table}

\begin{figure}[h!]
  \centering
  \includegraphics[width=\linewidth]{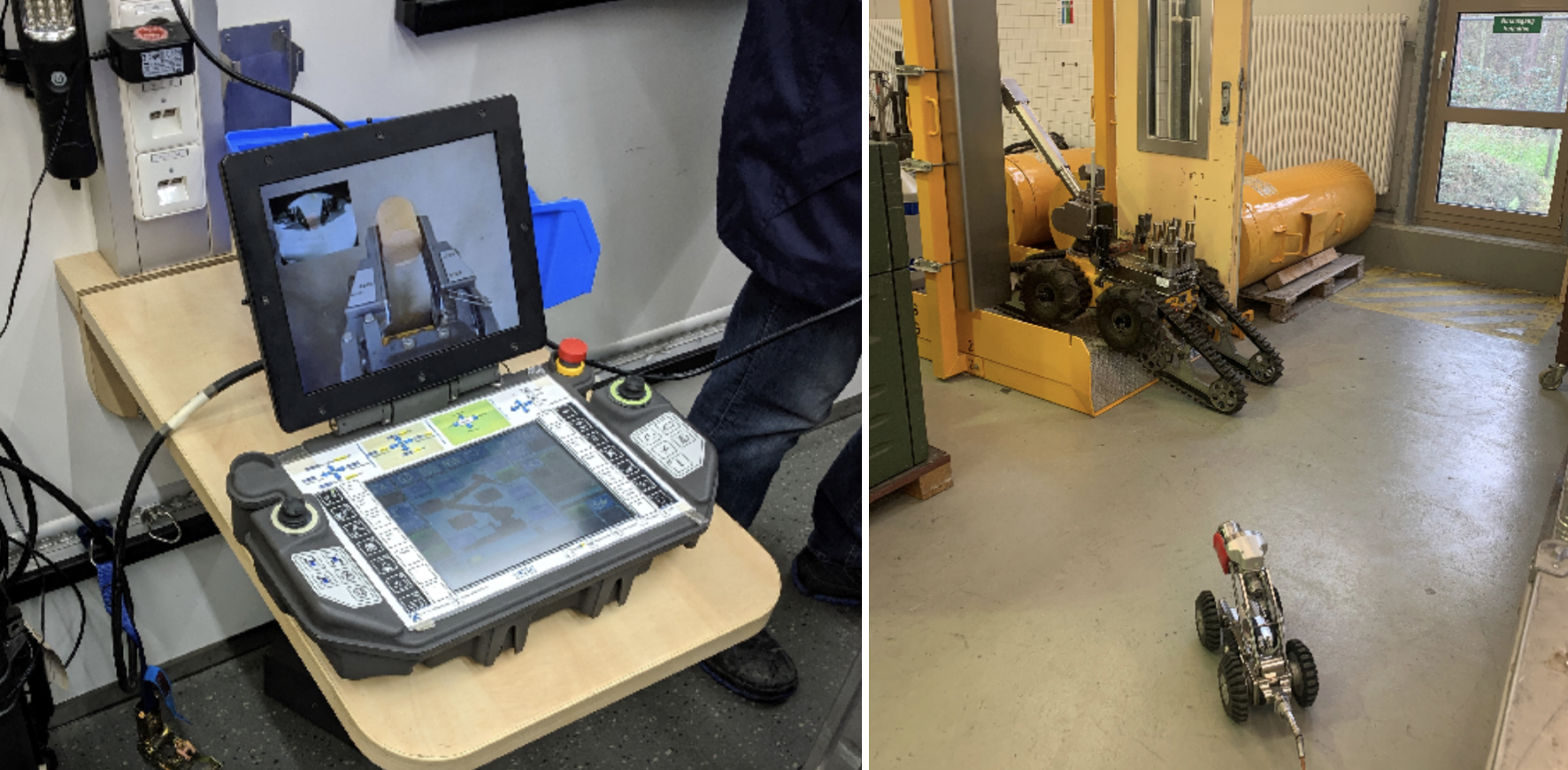}
  \caption{Nuclear disaster response application: user interface and robots in the field}
  \label{fig:nuclear}
\end{figure}
\clearpage

\subsection{Fruit Inspection Application}

\begin{table}[h!]
    \centering\small
    \begin{tabular} {|p{.15\linewidth}|p{.75\linewidth}|} \hline
        Participants & P1-3, P5 \\
        \hline

         Environment &  A mosaic farm site. \\
         \hline

         HRT Interface &  A visual interface with a view to verify detected fruits and a map view with robot status panels, which also facilitates high-level waypoint -like commands. \par A joysick controller to teleoperate the robot. \\
         \hline

         Team Mission & Perform a human-in-the-loop exploration to detect and locate fruits on a farm and inspect  them closely. \\
         \hline

         {\# of Humans} & 1 collaborator \\
         \hline
         
         {\# of Robots} & 2 (1 quadruped robot, 1 all-terrain robot)\\
         \hline
         
         {Role of Human} &  Monitor robots through a remote interface and provide high-level guidance by issuing waypoints or allocating tasks, or low-level guidance through teleoperation, as required. Verify fruits and other objects detected by robots, and select fruits for close inspection.  \\
         \hline
         
         {Role of Robot} & Explore the map by navigating the farm area. Detect fruits and other objects of interest and report back to the human. Perform close inspections of fruits marked by the human. Operate under the selected level of autonomy: teleoperation (low), follow waypoints/ prioritise  or deprioritise an area (mid), act autonomously (high). \\
         \hline
\end{tabular}
    \caption[]{Environment, Interface, Mission, Team  Structure and Role Details of Fruit Inspection Application}
    \label{tab:FruInsdetails} 
\end{table}

\begin{figure}[h!]
  \centering
  \includegraphics[width=\linewidth]{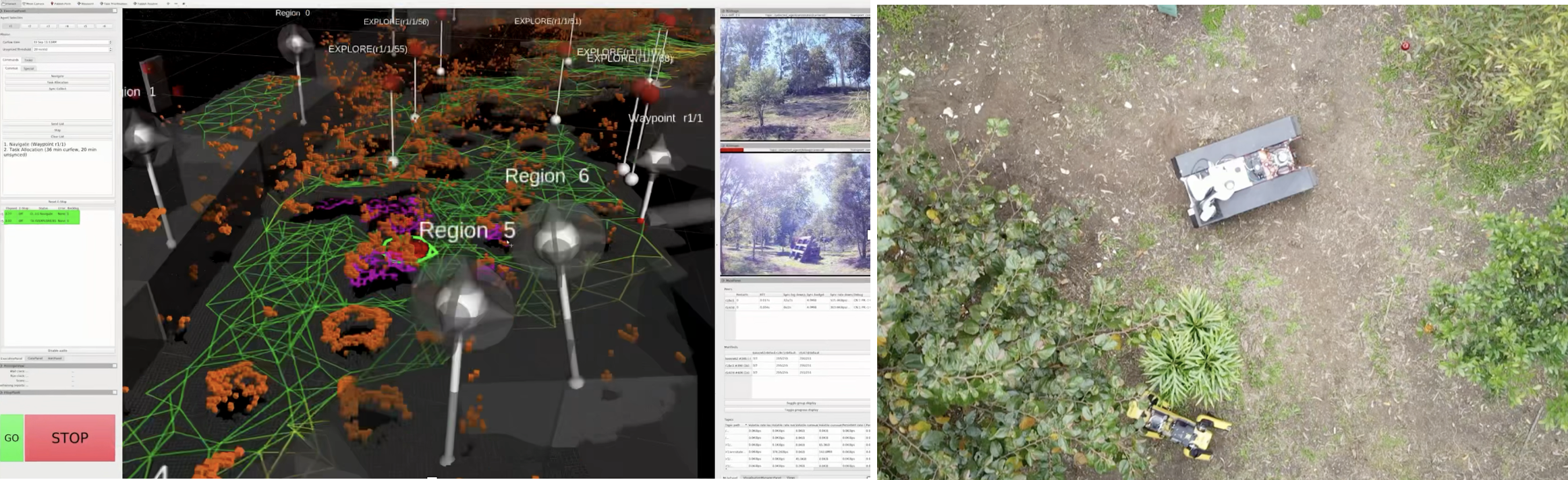}
  \caption{Fruit inspection application: user interface and robots in the field}
  \label{fig:fruit_inspection}
\end{figure} 
\clearpage

\subsection{Fruit Harvesting Application}

\begin{table}[h!]
    \centering\small
    \begin{tabular} {|p{.15\linewidth}|p{.75\linewidth}|} \hline    
        Participants & P4, P6  \\
        \hline

         Environment & An apple orchard. \\
         \hline

         HRT Interface & A graphical interface to view a visualisation of apples detected, picking order and progress of  picking, which also allows the operator to select or deselect apples to pick, change the picking order and stop the robot. \par A visual interface to  view the joint-wise movements of the arm. \par A joysick controller to teleoperate the robot. \\
         \hline

         Team Mission & Get the robot to harvest as many apples as feasible.\\
         \hline

         {\# of Humans} & 1 - 2 collaborators \\
         \hline
         
         {\# of Robots} & 1 (Arm with a gripper on a mobile base) \\
         \hline
         
         {Role of Human} & Drive the robot through the orchard and position it closer to the canopy for harvesting, using a remote controller (low autonomy). Override robot's decisions by filtering out apples that may cause damage to the arm when harvesting, adjusting the sequence  of harvesting, adjusting the approach angle or stopping the arm to avoid potential collisions (mid autonomy). Manual recovery of the arm and gripper when those cannot get back to planned trajectory (low autonomy).  \\
         \hline
         
         {Role of Robot} & Detect where the apples are, decide in which sequence to pick the apples, try to harvest all of the selected apples within the view through automatic path planning and trajectory following, and if the gripper or the arm hits something, activate  emergency stop mode (high autonomy).\\
         \hline
\end{tabular}
    \caption[]{Environment, Interface, Mission, Team  Structure and Role Details of Fruit Harvesting Application}
    \label{tab:FruHardetails} 
\end{table}

\begin{figure}[h!]
  \centering
  \includegraphics[width=\linewidth]{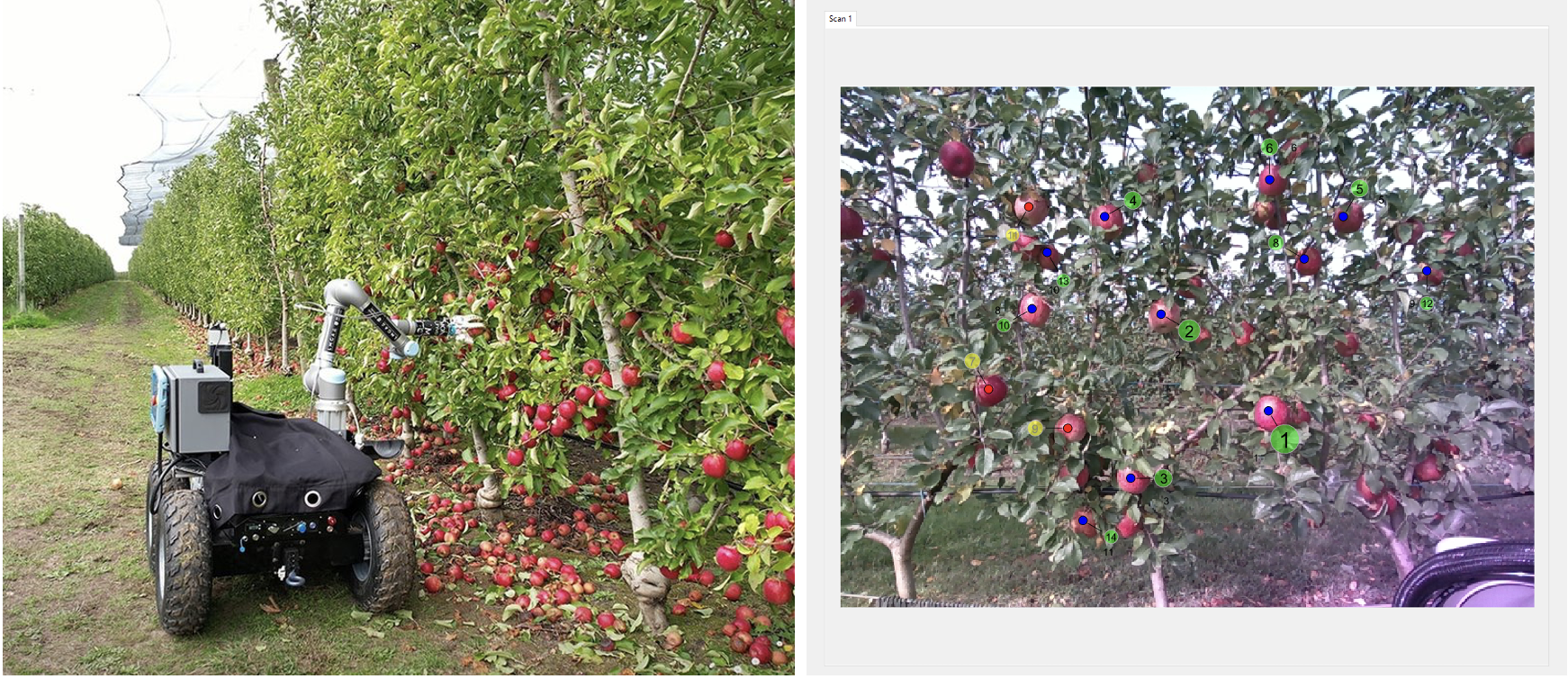}
  \caption{Fruit harvesting application: the robot in the field and part of user interface}
  \label{fig:fruit_picking}
\end{figure}
\clearpage

\subsection{Crafting Application}

\begin{table}[h!]
    \centering\small
    \begin{tabular} {|p{.15\linewidth}|p{.75\linewidth}|} \hline
    
        Participants & P7-8 \\
        \hline

         Environment & An experimental setting replicating a part of a shared art studio. \\
         \hline

         HRT Interface & A visual interface showing the crafter's front and side camera feeds, and a map view of the robot. \par A joystick controller to teleoperate the robot and the arm. \\
         \hline

         Team Mission & Make the robot a useful assistant to crafters, so they can share tools and materials during crafting activities. \\
         \hline

         {\# of Humans} & 1 collaborator, at least 1 crafter (who signals the robot when need crafting materials or need to return materials) \\
         \hline
         
         {\# of Robots} & 1 (Fetch robot)\\
         \hline
         
         {Role of Human} & Issue a set of pre-defined commands (go forward, turnaround, position to handover) to the robot using a joystick so it drives between crafters' desks, hands over the crafting materials and collects those back by observing crafters' signals, identifying when they need something, and deciding when the robot should initiate the handovers. \\
         \hline
         
         {Role of Robot} & Hold a basket from which the crafter can pick crafting materials and return them.  Drive between crafters' desks as instructed, stop automatically when reaching the desk, and hand over the materials or collect returning items at right, left or centre as instructed by moving the basket through motion planning. Face towards the basket location when handing over, as a social action.\\
         
         \hline
\end{tabular}
    \caption[]{Environment, Interface, Mission, Team  Structure and Role Details of Crafting Application}
    \label{tab:Craftdetails} 
\end{table}

\begin{figure}[h!]
  \centering
  \includegraphics[width=\linewidth]{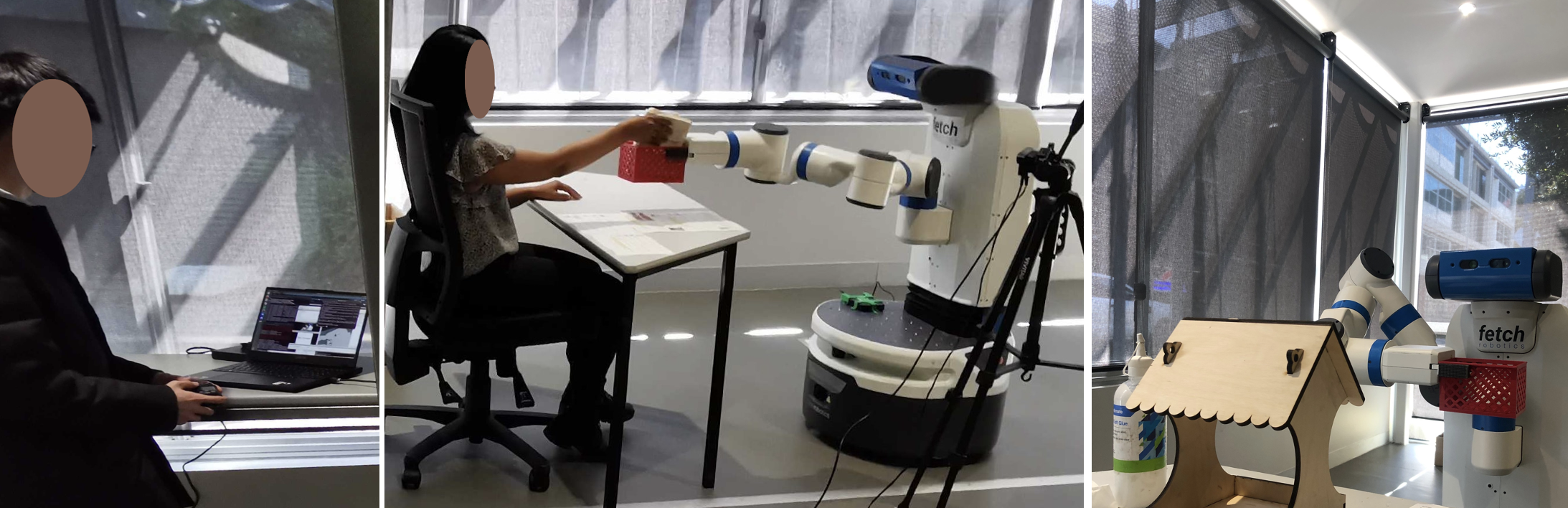}
  \caption{Crafting application: human collaborator, robot handing over material to crafter and crafter's view}
  \label{fig:crafting}
\end{figure}

\clearpage

\subsection{Mine Inspection Application}

\begin{table}[h!]
    \centering\small
    \begin{tabular} {|p{.15\linewidth}|p{.75\linewidth}|} \hline
        Participants & P9-11 \\
        \hline

         Environment & Underground mine site. \\
         \hline

         HRT Interface & An app on the tablet with a live camera feed and a real-time point cloud of the environment, which can also also used to send high level waypoint-like commands. \par A joystick controller to teleoperate the drone. \\
         \hline

         Team Mission & Explore a mine and create a 3D model of the environment to get a detailed picture (shape, volume of an area, unstable structures and other dangerous areas to send people) to make decisions around mining activities.\\
         \hline

         {\# of Humans} & 1 collaborator (surveyor/ mining operator/ geologist) \\
         \hline
         
         {\# of Robots} & 1 (drone) \\
         \hline
         
         {Role of Human} &  Monitor the drone and collected data (high autonomy), and decide whether the data collected is sufficient and  whether the drone requires intervention. Task the drone to go and collect missing data from specific areas to get a full picture or assist the drone in avoiding damage to the environment, surrounders or itself by issuing waypoints (mid autonomy) or teleoperating while the in-built collision avoidance is enabled (low autonomy) or disabled (lowest autonomy).  \\
         \hline
         
         {Role of Robot} & Fly around until it decides that it has mapped the area indicated by the human while avoiding obstacles. Autonomously operate in the event of a loss of  communication with the operator until a failsafe kicks in (e.g., low battery, SLAM error), initiating returning home or landing at a safe place.\\
         \hline
\end{tabular}
    \caption[]{Environment, Interface, Mission, Team  Structure and Role Details of Mine Inspection Application}
    \label{tab:Minedetails} 
\end{table}

\begin{figure}[h!]
  \centering
  \includegraphics[width=\linewidth]{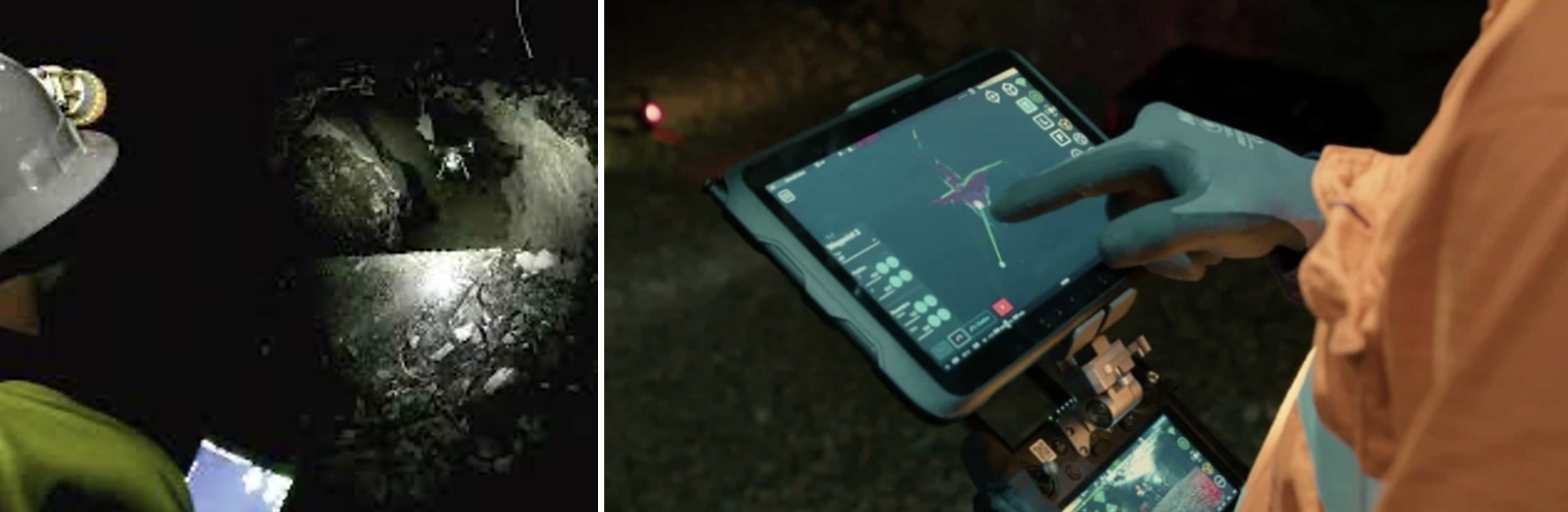}
  \caption{Mine inspection application: the drone in the field and the user interface}
  \label{fig:mining}
\end{figure}

\clearpage

\subsection{Aircraft Manufacturing Application}

\begin{table}[h!]
    \centering\small
    \begin{tabular} {|p{.15\linewidth}|p{.75\linewidth}|} \hline

        Participants & P14 \\
        \hline

         Environment &  Aircraft parts manufacturing facility. \\
         \hline

         HRT Interface &  A visual interface on the robot showing progress, errors and messages to  the operator. \par Lights on the robot with sounds indicating the robot's status. \\
         \hline

         Team Mission & Achieve a manufacturing outcome (e.g., sanding, drilling, fascinating, sealing) in the process of building an aircraft.\\
         \hline

         {\# of Humans} & 1 collaborator \\
         \hline
         
         {\# of Robots} & 1 (universal robot with a tool on the end) \\
         \hline
         
         {Role of Human} & Do the same job as the robot, do preparation and make sure that the task is possible for the robot, or finish parts of jobs cannot be completed by robots.  \\
         \hline
         
         {Role of Robot} & Execute the job plan provided by the human by carrying out tasks such as sanding, drilling, fastening (nuts and bolts), sealing (applying Silicon), or performing inspections. Display messages on the interface and wait when human intervention is needed.\\
         \hline
\end{tabular}
    \caption[]{Environment, Interface, Mission, Team  Structure and Role Details of Aircraft Manufacturing Application}
    \label{tab:AirMfgdetails} 
\end{table}

\begin{figure}[h!]
  \centering
  \includegraphics[width=\linewidth]{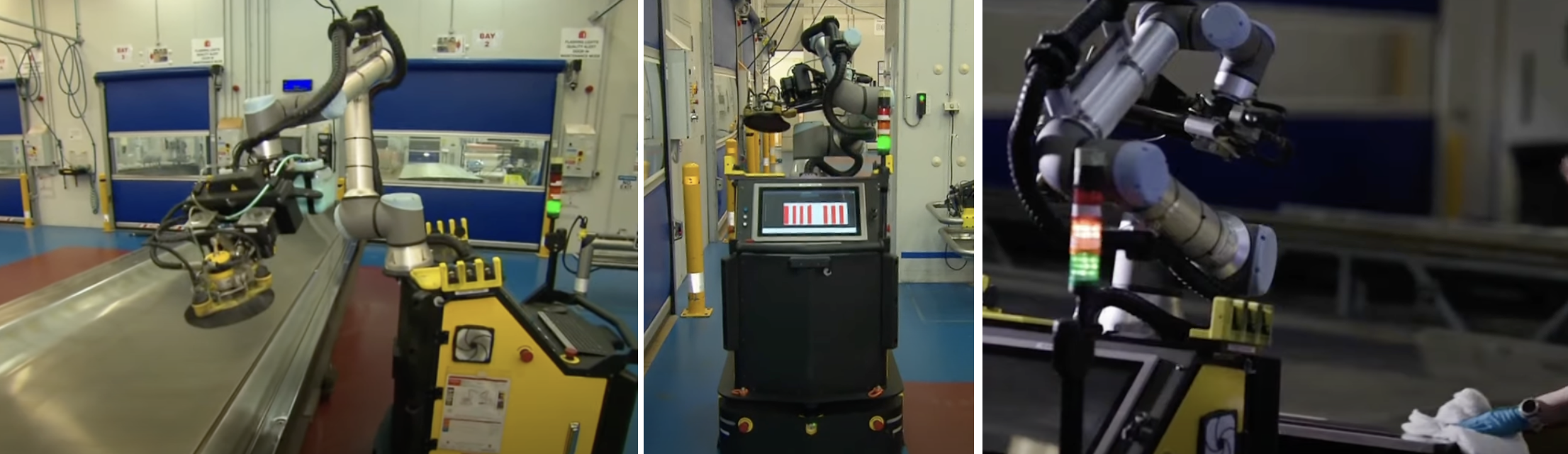}
  \caption{Aircraft manufacturing application: robot in action, visual interface and human working with a robot}
  \label{fig:aircraft}
\end{figure}

\clearpage

\subsection{Large 3D Artwork Manufacturing Application}

\begin{table}[h!]
    \centering\small
    \begin{tabular} {|p{.15\linewidth}|p{.75\linewidth}|} \hline

        Participants & P16 \\
        \hline

         Environment & Artwork manufacturing facility. \\
         \hline

         HRT Interface & A visual interface on the robot showing progress, errors and messages to the operator. \par Lights on the robot with sounds indicating the robot's status. \\
         \hline

         Team Mission & Collaborate with the robot to manufacture high-standard, large-scale, unique artwork while allowing humans  to focus on creativity rather than  repetition and tasks that the robot cannot be easily programmed to do. \\
         \hline

         {\# of Humans} & 1 collaborator (finisher)\\
         \hline
         
         {\# of Robots} & 1(universal robot)  \\
         \hline
         
         {Role of Human} &  Program the robot to complete certain parts of the  artwork. Work beside the robot to assess and perform whatever the robot was unable to do (e.g., in the plasma cut process, deciding the radius, moving the robot to a point to initiate cutting). Complete actions requested by the robot (e.g., change a sandpaper). Stop the robot in an emergency. \\
         \hline
         
         {Role of Robot} & Execute the jobs that it is programmed to do (e.g., plasma cutting, metal polishing - linishing/ sanding). Display messages on the interface and wait  when pending human input or action. \\
         \hline
\end{tabular}
    \caption[]{Environment, Interface, Mission, Team  Structure and Role Details of Large 3D Artwork Manufacturing Application}
    \label{tab:Artdetails} 
\end{table}

\begin{figure}[h!]
  \centering
  \includegraphics[width=\linewidth]{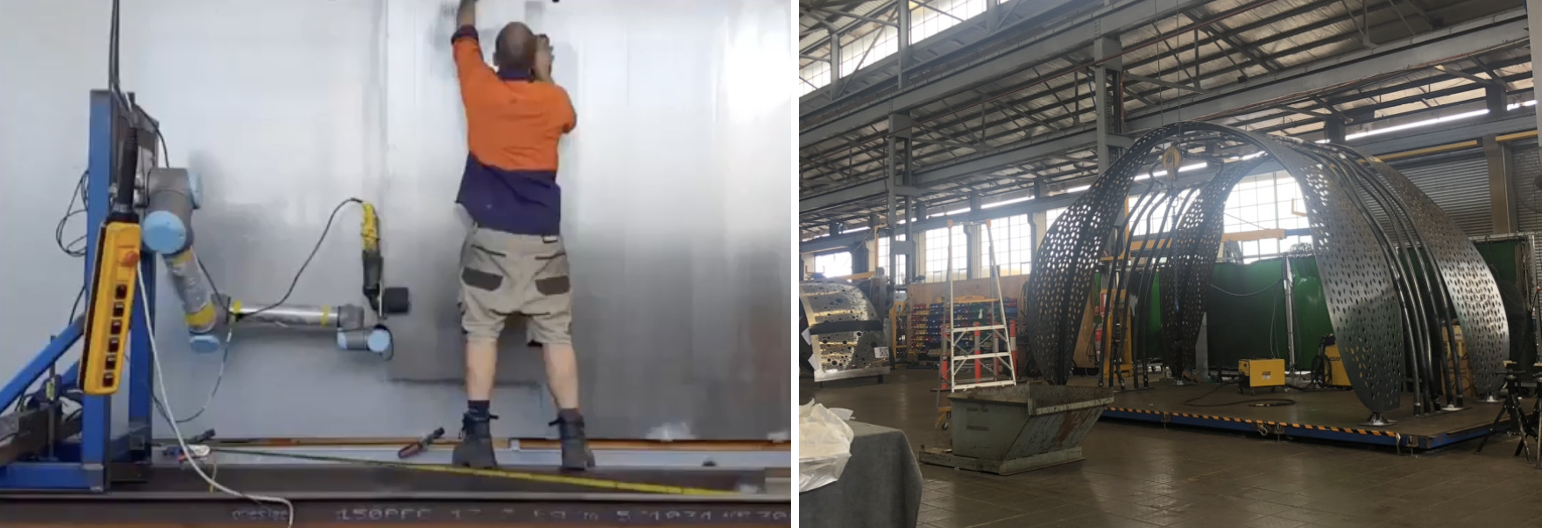}
  \caption{Large-scale art manufacturing application: human working with a robot, and sample large-scale arts}
  \label{fig:large_art}
\end{figure}

\clearpage

\subsection{Laparoscopy Surgery Application}

\begin{table}[h!]
    \centering\small
    \begin{tabular} {|p{.15\linewidth}|p{.75\linewidth}|} \hline
    
        Participants & P15 \\
        \hline

         Environment & A physical simulator replicating the abdomen to practice suturing-like skills. \\
         \hline

         HRT Interface & A foot interface to override auto zooming and switch between auto and manual modes. \par A visual interface to view the zoomed-in surgery area and the foot interface to adjust the foot position.\\
         \hline

         Team Mission & Zooming in to the surgical area during laparoscopic surgery training activities, so the surgeons can have a clear  vision of what is been done and improve accuracy.\\
         \hline

         {\# of Humans} & 1 collaborator (surgeon) \\
         \hline
         
         {\# of Robots} & 1 (an arm controlling a camera) \\
         \hline
         
         {Role of Human} &  Override the auto zoom-in function of the robot or instruct the robot to zoom out to get a perspective of the surrounding area, using a foot interface (low autonomy). \\
         \hline
         
         {Role of Robot} & Detect and auto zoom into the target area in the laparoscopic procedures, so it is clear for the surgeon (high autonomy).\\
         \hline
\end{tabular}
    \caption[]{Environment, Interface, Mission, Team  Structure and Role Details of Laparoscopy Surgery Application}
    \label{tab:Surgdetails} 
\end{table}
 
\begin{figure}[h!]
  \centering
  \includegraphics[width=\linewidth]{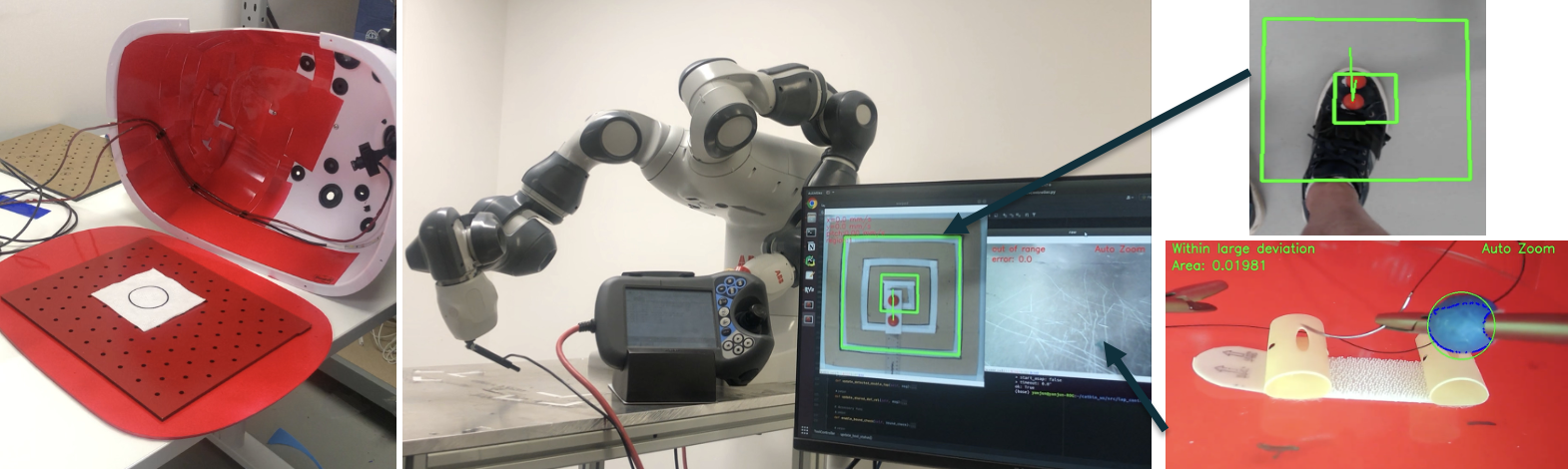}
  \caption{Laparoscopy surgery application: simulator box and user interface with a view of foot interface and surgical area}
  \label{fig:surgery}
\end{figure}

\clearpage

\clearpage
\newpage

\section{Appendix: Examples of Contextual, Robotic and Human Factors Influencing Required Situational Awareness}

\begin{center}
\small
\begin{longtblr} [caption = {Examples of Contextual Factors Influencing Required Situational Awareness},
label = {tab:Contexual_Required_SA} ]{|p{.1\linewidth}|p{.15\linewidth}|p{.66\linewidth}|}  \hline
        \textbf{Contextual Factor} & \textbf{Trend in} \par \textbf{Required SA} & \textbf{Example}\\
        \hline
        Task Criticality and Challenge & High when the environment is tricky for robots or when involved in critical tasks & When the robots are in ``complex environments'' (P2) and acting ``around novel objects or areas'' (P1), such as ``narrow spaces” (P5), areas with smoke where “autonomy would see smoke as a   barrier'' (P1), unusual terrains with ``railways'', ``big amounts of rubble'', ``arbitrary piles of rolls'',   or ``sheer cliffs'' (P1, P5), they are more likely to ``get in trouble'' and require frequent attention   and interventions by human to overcome associated traversability challenges and ``to stop them   from doing things that they should not necessarily be doing'' (P2).\\
        \cline{3-3}
        &&``When driving the robot through a narrow and confined doorway'' where it was difficult for the human operator to comprehend whether ``robot was way too big to fit in'' without using a secondary view, and ``when controlling the robot arm in bad lighting'' (P12).\\
        \cline{3-3}
        &&When ``the workspace for the robot is quite challenging'' (P4), e.g., when ``the apples are closely   clustered together'', the human is required to modify the robot-proposed order of picking. When   ``an entire branch moves to a different position'' as a result of extracting an apple or ``when   apples are unintentionally knocked off the canopy'', human operators needs to rectify these  dynamic changes by rescanning the entire canopy (P6).\\
        \cline{3-3}
        &&When the drone fails to differentiate ``very small wires <or> wall-<like structures in> dust <or>   humid areas that form condensation on the lens'' (P9, P10) or to detect ``changes to the <initially   scanned> environment, such as a vehicle has been moved on the way to <its> landing area'' (P11) \\
        \cline{3-3}
        &&``When there is bleeding from a vessel or any risk to an organ requiring tissue repair'', the surgeon’s   required SA increases (P15). \\
        \cline{3-3}
        
        &
        & When the robot is carrying out a step that “is really important <and> it is not possible to fix <the errors made in that step> by hand”, the humans usually “pay more attention” (P14). \\
        \cline{3-3}
        &&During the “most critical part of gallbladder surgery”; i.e., when the surgeon is differentiating the “gallbladder duct and artery before tissue dissection, <the robot-operated> cameras should be very still and need to be zooming at the right moment” for the surgeon to reach the highest level of SA about the surgical area and the robot's actions (P15).\\
        \cline{2-3}
        
        &Low when the environment is not tricky for robots or when not involved in crucial tasks & ``When the ground is quite flat and open'', the human operator ``does not need high SA to monitor  the robots'', as the ``robots are quite robust'' and ``capable of being autonomous without any  supervision'' in such ``easy terrains'' (P2, P3).\\
        \cline{3-3}
        
        &&``When the canopy is in perfect condition'', i.e., the ``canopy is completely flat'' and ``there is no chance of collision with wires'', humans can ``let <their> guard down a little bit (P4, P6).\\
        \cline{3-3}
        
        &
        & When the robots are progressing well with their tasks, and those tasks are not critical components of the mission, humans felt ``relaxed'' (P5) and could ``take a quick break'' (P6), while focusing on ``monitoring and comprehending the situation at a high level'' (P5).\\
        \cline{3-3}
        &&During the stages where the crafter is engaged in ``painting or assembling the birdhouse'' (P8), the human ``needs not put in much attention <and can> look away for a bit'' (P7) as the level of interactivity of the crafter with the robot is minimum and ``robot is just waiting for the <crafter> to finish and initiate a signal'' (P7).\\
        \hline

        Safety Criticality of tasks & High when robots are in danger or causing danger & When ``robots are about to fall off a cliff'' (P3) \par ``the robot tries to grasp an apple that is behind a trellis wire, <which the robot cannot perceive, risking> damaging its fingers'' (P6) \par flying a drone in ``heavy dust <that> risks damage to the system'' (P10) \\
        \cline{3-3}
        
        &
        & ``If a robot is   impacting a tree'' (P1) \par ``if a person comes closer without   understanding what is going on'' (P2) \par ``when there is not a lot of room for the robot   to move without colliding with the canopy'' (P6) \par ``when the drone starts moving out into areas   where people are working in the mine'' (P9) \par when working with ``robots with higher   payload <or working at> full speed'' (P16) \par when a robot-controlled ``tool is   going outside the plate'',   causing damage to the art and   risks to people around (P16) \par if an auto controlled zooming   ``camera collides with the   instruments <that a surgeon>   was holding'' (P15)\\
        \cline{2-3}
        
        & Low when robots are not in danger &
        ``When a robot is navigating through open space'' (P2), human collaborators would just focus on their usual roles (e.g., verifying robot-detected objects) rather than overriding the robot's actions.\\
        \hline
        The number of robots per human & High when there are multiple robots per human & when ``the number of agents is high'' (P5)\par
         when one human controls ``multiple robots'' (P6,P13,P14) \par
         when ``having 4 robots on the same piece'' (P16)\\
        \cline{2-3}
        & Low when there are limited robots per human& When having ``two operators'', each assigned to  the ``main robot'' and the ``secondary robot'' (P12)\\
        \cline{3-3}
        & & When having ``one operator to move the mobile base to a specific spot, and another operator to initialise the visualisation process at the same time'' on the same robot (P4)\\
        \cline{3-3}
        & & If ``everybody <around> can take care of'' single robot, reducing the waste of time associated with ``waiting for <one specific human> to change the sandpaper with a pop up on the screen'' (P16)\\
        \hline
        Time criticality of the mission & High when the time criticality is high & Because the rescue mission ``is time-constrained'', humans ``have to maintain higher SA levels'',  ``purely divide their  attention <among  robots, and intervene on time>'' (P1, P2, P3, P5)\\
        \cline{2-3}

        & Low when the time criticality is low & ``In an agricultural'' mission, there are ``allowances for delays''; ``if there are a few bits that robots do not know, it is fine to buffer those questions and let the human respond at their convenience'' (P1, P5)\\
        \cline{3-3}

        & & Since aircraft manufacturing systems are ``not necessarily time-critical, like if <the humans> do not intervene <as soon as the need arises> something very bad will happen'', robots are designed to ``wait for human input.'' (P14)\\
        \hline

        Time spent on the mission & High  when spent less time& ``At the beginning of a mission'', given that humans ``need to pay more attention'' (P4) ``to figure out what is going on'' (P8) and ``need to make predictions'' as they have less information <about the context>'' (P5) and ``do not know what to expect'' (P8).\\
        \cline{2-3}

        & Low when spent considerable time & ``Towards the end of the mission'' (P5), as there is no need to ``do that much with predictions'' (P5) as humans have developed a good understanding of what to expect (P7, P8) and ``become comfortable with the robot's ability to do particular things'' (P6).\\
        \hline
        
\end{longtblr}
\end{center}

\clearpage

\begin{center}
\small
\begin{longtblr} [caption = {Examples of Robot-related Factors Influencing Required Situational Awareness},
label = {tab:Robotic_Required_SA} ]{|p{.1\linewidth}|p{.15\linewidth}|p{.66\linewidth}|}  \hline
        \textbf{Robot-related Factor} & \textbf{Trend in} \par \textbf{Required SA} & \textbf{Example}\\
        \hline
        Capability of robot autonomy & High when the robot is operating at a lower level of autonomy & When switching to lower levels of autonomy, humans reported experiencing 
        a need to ``mentally switch into a high SA'' (P5), ``remain quite vigilant'' (P11) and ``pay more attention'' (P4), given the increase in SA demands (P9), reduction in the ``level of assistance'' to maintain SA at required levels (P11)  and ``interruption to focus on <primary> tasks'' that the human operators are responsible for (P15).\\
       \cline{3-3}

       & & When teleoperating a robot with no autonomous capabilities, humans need to maintain a higher level of awareness of the ``position of the robot, the position of the arm, whether a collision is likely to happen'' and ``be there  100\%'' most of the time (P12). \\
       \cline{3-3}

       & & During the ``early stages of <product research and> development'' process (P1), where: ``the autonomous capabilities are lower than in the end product'' (P1), e.g., ``robots are not able to handle particular kinds of hazard, <such as> water, railway tracks, and situations where a couple of robots can get stuck head to head in a narrow tunnel'' (P1) ``robot is not capable of avoiding collisions with items on the desk during handover'' or guessing ``when to initiate handover by analysing user signals'' (P7, P8)\\
       \cline{2-3}
        
        & Low when the robot is operating at a higher level of autonomy & ``Situational awareness'' demands can ``vary a lot'' as the level of autonomy changes, and when the robot is acting under ``autonomous capabilities, <humans> could be distracted a bit'' (P12). ``The autonomy made the drivability much easier and safer'' in most scenarios (P12), allowing humans to ``focus on <making important> predictions without <trying to> perceive everything. <Thus, they will be able to> preemptively take necessary actions'' to override the robot's behaviour (P13), and to perform necessary ``secondary tasks, <such as> communicating to the commanders and firefighters the position of  the victims or how the situation looks like.'' (P13). \\
        \cline{3-3}

        & & Towards the later stages of product research and development process, where robots are more capable (P1, P7, P8), so humans would be able to  ``reduce the attention paid'' to capability aspects (P7)  and ``focus on other'' aspects (P8)\\
        \hline

        Software and hardware robustness & High when robots are less robust & When there are errors resulting from software or hardware robustness issues (e.g., mapping errors 
        and localisation errors due to issues with SLAM (P1, P13), failing ``to park the robot at the correct time, due to wear and tear condition of robot's wheels'' (P4) ``robot not stopping <at the end of the path> due to programming mistake'' (P16))\\
        \cline{3-3}
        
        & & The humans need to put extra effort into comprehending errors and identifying workarounds to ``recover from mistakes'' (P7) (e.g., ``robot grasped a leaf rather than an apple'' (P6) due to a computer vision perception issue) and reduce the undesirable impacts (e.g., ``drone crashed <in a certain scenario> due to an issue with the firmware version'' (P11))\\
        \cline{3-3}
        
        & & In case of a lack of fail-safe mechanisms to overcome communication loss, humans need to focus on additional tasks such as ``bridging communication'' using other robots to bring them back and to retrieve collected data (P1).\\
        \cline{2-3}

        & Low when robots are robust &
        When the robots ``do what <they are> meant to do'' without making any errors (P6) \\
        \cline{3-3}
        & & ``When the robots seem to be consistently achieving their goals'' (P5) \\
        \cline{3-3}
        & & When ``the autonomy itself has some failsafes,'' (P3) implemented to overcome communication loss (e.g., ``there are pre-set curfew times for the robot to come back on time'' (P3) or drones are designed to ``returns on <their> own to a safe location or the home point'' (P11) if ``the battery is running low or <upon experiencing a> SLAM loss'' (P10)), allowing humans to focus on tasks other than monitoring the robots who are out of communication range, knowing that those robots will come back with data.\\
        \cline{2-3}
        The severity level of the robot's status & High when the severity level of status is high & A ``critical low-battery warning'' (P9) and an ``error indicating that the collision avoidance shield is no longer working'' (P10)  are more critical and require priority interventions when compared to a low-battery warning indicating a 10\% battery (P11). \\
        \cline{2-3}
        & Low when the severity level of status is low & When a warning is telling that the robot is ``idling'' (P2, P3) \\ 
        \cline{3-3}
        & & In ``scenarios where the robot is programmed to wait for human input'' (P14) (e.g., request and wait for a human to ``change a sandpaper on a tool'' (P16), ``make a clamping point'' (P14), analyse and ``replan when sensed something unexpected or that <the robot> does not understand'' (P14))\\
        \hline
\end{longtblr}
\end{center}

\clearpage

\begin{center}
\small
\begin{longtblr} [caption = {Examples of Human Factors Influencing Actual Situational Awareness},
label = {tab:Human_Actual_SA} ]{|p{.1\linewidth}|p{.15\linewidth}|p{.66\linewidth}|}  \hline
        \textbf{Human Factor} & \textbf{Trend in} \par \textbf{Actual SA} & \textbf{Example}\\
        \hline
        Expertise and contextual understanding & Close to the required SA when the expertise and contextual understanding is high & The learning process over time (P3, P12, P8, P16), ``training on the system'' (P9) and ``pre-knowledge of the workflow'' (P15) assisted humans in ``faster perceiving'' (P12) information that they previously struggled with, comprehending the correct time to do certain actions (P3, P8) (e.g., when to handover materials to crafters (P8),  when to drop off a WiFi node carried by a robot to bridge communication (P3)), making right predictions about robots' actions (P3, P15, P16) (e.g., ``be able to predict how <a particular tool would behave> on the robot'' (P16)), and understanding why robots behave in particular ways and make certain decisions (P9, P10, P11).\\
        \cline{2-3}
        & Lower than the required SA when the expertise and contextual  understanding is low &``Depth perception'' errors (P12) and wrong ``comprehension'' of the need to intervene (P2) occurred due to a lack of contextual knowledge about the environment. 
        \\
        \cline{3-3}
        
        & & Partly due to an operator's ``shortfall in <contextual> understanding'' with no prior knowledge about that mine, they failed to understand the drone’s  
        environment properly. This resulted in them choosing a planner waypoint, assuming that it would make the ``drone go down and take a 90-degree right turn to get into a void area <in a mine>'', which made the drone ``take off and fly backwards. ``It was not until after <they> looked at the scan'' that they realised ``the drive <they> were operating on was not a 90-degree corner''. Instead, ``it folded back, intersecting with the used planner waypoint.''  (P10). \\
        \cline{3-3}
        & & Humans made errors in robot control (P7, P8, P12) and forgot to monitor the mission effectively by frequently switching between different robots, tasks or information panels (P1, P2, P3) due to the lack of ``familiarity with the user interface'' when they were ``novices''; e.g., P7 ``diverted <their> attention a little bit'', causing a lack of SA on the primary task, due  to referring to ``a cheat sheet for instructions of how to use the controller with joysticks''.\\
        \hline        
        Mental Models about Robots & Close to required SA when the humans' mental models align with robots' actions and skills & Some of the human collaborators, who were also the ``developers'' (P5, P9) of ``certain parts of the <robots'> software stack'', identified themselves as having ``a very good mental model on how those parts work and what <they> should expect from robots'' (P5) and ``a <good> insight in terms of what <robots> can do'' (P9), therefore, being able to ``quickly identify cases where <robots are> not going to perform well and <when to> manually override <robot actions>'' (P5). 
        For example, ``as part of the team that developed the global navigation component'', P5 knew that ``when <multiple> robots enter the field near each other, if <robots> are switched immediately into exploration, they are probably going to waste some time'' because ``that component is going to be weakest at the beginning of the mission''. So, they usually ``provide waypoints to force robots to spread out before switching into exploration''.\\
        \cline{2-3}
        
        & Lower than required SA when the humans' mental models misalign with robots' actions and skills & During human collaborators' firsthand experiences, they experienced poor SA, demonstrated by ``failures to intervene at the right times, intervening too early, or not intervening when needed'' (P5). These experiences reflect deviations in actual SA from the required SA, which resulted from ``a clash between <operators'> mental model <and the reality>'' (P5) or when it is hard to develop mental models to explain what the robot is doing (P6, P9, P14) or why robots make certain decisions (P1, P9), given the lack of knowledge about (P5, P9, P10) or experience with (P10) the system, and lack of explainability of the system (P14, P1).\\
        \cline{3-3}
        & &         A person who came in close proximity to a robot while the robot and its human collaborator were in action failed to move away ``when the robot-controlled tool was coming towards him''. Although their attention was on the robot, they behaved this way because they ``had this expectation of <since the collaborator> was there, nothing could happen to them'' (P16).\\
        \cline{3-3}
        & & The crafters ``came up with different models <through> different explanations for the robot's behaviour,'' e.g., ``whether the robot is autonomous or being teleoperated'' and ``whether the robot is adaptive or not,'' contributing to correct or incorrect SA. For example, some crafters who noticed that the ``robot <had> done handover at different positions'' thought the ``robot <was> adaptive'' (which is correct) but related the robot's ``head tilting behaviour <to> looking at the person's hand to decide where to stop'' (which is incorrect) (P8).\\

        \hline

        Trust in automation & Higher than required SA when humans undertrust robots & Human collaborators often undertrust robots during their initial interactions (P1, P5) or when robots are operating in new environments (P3), making humans ``very hands-on'' (P6) and ``try to override a lot of the autonomy'' until they make a ``mental transition to trust the robots more'' through a wide range of experiences (P5).\\
        \cline{3-3}

        & & The farmers that P6 work with usually demonstrated a ``level of distrust and resistance in adopting robots at the beginning until the robot can prove that it is capable of decision making and <fruit harvesting>'', given the ``black box nature of robots''.\\
        \cline{2-3}
        & Lower than required SA when humans overtrust robots & Once, the operator failed to detect a ``depth perception error'' of the robot due to not paying much attention (i.e., low SA), which resulted in ``grasping a leaf rather than an apple'' (P6).\\
        \cline{2-3}
        & Closer to the required SA when human trust in robots is properly calibrated & ``In the early stage of development, the system is not that trustworthy,'' as the rate of errors by robots is high, so it is not appropriate to blindly trust the robot (P3). When ``the robots' capability <and robustness> evolve'', the humans need to ``gradually build trust'' (P1), develop the ``appropriate sense of confidence'' (P16), and achieve relatively ``low levels of SA'' compared to early development stages (P3), so they can ``back off from micro-managing <the robots and> let robots to act more unsupervised''  (P3).\\
        \cline{3-3}
        & & Humans ``realised that the <robot> navigation was much better'', while initially they ``had that frame of mind that the robot was not capable'', but later on ``when the robot was in a similar location'' they were ``about to grab the joystick, assuming that the robot was in a very precarious spot, but then stopped <intervention> and the robot was able to get out of that situation'' (P3).\\
        \cline{3-3}
        & & After learning ``how to set up safety areas'' around a robot through programming and ``how to quickly shut off the tool'' through experience, the human could trust the robot more and ``felt confident <to> stay aside <the robot when it is operating at> full speed'' (P16).\\

        \hline

        Cognitive capacity and ability to multitask & Lower than required SA when humans' capacity to multitask is low & If the human does ``not have the capacity to do more than one thing at once'' (P3) and is ``not good at multitasking'' (P5) when there are demands for multiple parallel tasks (e.g., ``using foot interface to override robot's zooming <while carrying out> the surgery'' (P15), ``two robots in very critical locations'' (P3), ``performing secondary tasks <of> communicating to the commanders the position of detected victims, <in addition to assisting the robots in exploring the site and detecting victims>'' (P13))\\
        \cline{3-3}
        
        & & During unpredictable, less-frequent tasks that arise during human-robot collaboration compared to predictable, repetitive tasks (P15), which may increase ``cognitive burden''\\
        
        \hline

        Willingness to delegate & Higher than required SA when humans are not willing to delegate & The tendency of not willing to delegate can arise from ``personality traits'' (P13), such as wanting to be in control of everything (P13) or execute exactly as they planned (P1). Humans with this trait ``like to take more control of the robot even in circumstances <when> the robot is going to perform better or ideally, and <operator> should not need to have some workload'' (P13), attempt to ``understand as much as they can'' about each robot and ``babysit'' each robot even in situations where robots could autonomously operate (P5), ``get the robot to execute exactly what they want, <while the robots could have> achieved those plans in a slightly different way (P1).\\
        
        \cline{3-3}

        & & The reason for unwillingness to delegate can also be related to context-based ``preference''; e.g., <when the operator is> bored just by monitoring, <they may prefer> to be in action'' (P13).\\
        
        \cline{2-3}

        & Lower than required SA when humans are unreasonably willing to delegate & When operators ``like to relax, <they may> prefer to give the robot the goal until something bad happens <without trying to optimise the mission>'' (P13).\\

    \hline
\end{longtblr}
\end{center}

\begin{center}
\small
\begin{longtblr} [caption = {Examples of Contextual Factors Influencing Actual Situational Awareness},
label = {tab:Contextual_Actual_SA} ]{|p{.1\linewidth}|p{.15\linewidth}|p{.66\linewidth}|}  \hline
        \textbf{Contextual Factor} & \textbf{Trend in} \par \textbf{Actual SA} & \textbf{Example}\\
        \hline
        
        Interruptions to information & Lower than required SA when communication with robots is interrupted & ``When <robots> lose comms <with humans at a remote station, humans'> SA of what is occurring with the robot is null and void; <they> do not know what's going on'' (P10). \\

        \cline{3-3}
         & & Interruptions to communication often take away <humans'> ability to assist robots in need (P1, P9) because humans cannot perceive or comprehend what exactly the robots are doing. During these periods, humans have to ``rely on <robots'> autonomy'' <if the robots are designed with> ``recovery behaviours'' (P1, P9) or slow down the primary tasks and work on to bridge or restore communication (P5, P7, P8) \\

        \cline{3-3}
        
        & & If the communication disruptions remain for ``short periods (e.g., due to radio frequency black spots)'', <humans would still be able to> ``make predictions to a certain extent'' (P5) (e.g., ``where <the robot> is or when it should be returning around'' (P9)) based on ``indicators on the map where it lost communications'' (P5).\\
        \cline{3-3}
        & & When it is not a complete communication interruption with a robot but a ``degradation'' (P2) or only losing one information channel among many others (P7), operators often experience limited access to information about the robot (e.g., losing ``video feedback'' <while having access to robots' locations on the map> (P2), losing ``footage from the robot's camera'' <while having access to> ``side camera footage'' (P7)) or  delayed access (P8, P13), creating challenges to make decisions because of incomplete SA.\\

        \cline{2-3}

        & Closer to the  required levels when interruptions to information is minimal & When the remote robots are ``within communication range and have steady communication <with the human>'', the operator's ``situational awareness <about robots> is gonna be much better'' (P2).\\

        \hline

        Distractions & Lower than required SA in the presence of distractions & When ``someone was walking behind <the operator>,'' that took their ``attention off a bit'' (P7).\\

        \cline{3-3}
        
        & & When acting as the main operator, ``there were other people inside <the control room> talking most of the time,'' which made it a ``bit difficult to concentrate on doing the job as <they> should'' (P12).\\

        \cline{3-3}

        & & Operators sometimes fail to ``pick up audio battery warnings'' implemented into their interface to indicate that the drone runs out of battery due to ``noisy'' surroundings in mines (P9, P10).\\

        \cline{3-3}

        & & ``Constantly being told about a notification that <humans have> already seen''; e.g., issuing multiple warnings to say that the drone is low in battery, are ``unwanted notifications''  that can take away their focus from what they should be doing by making them ``overwhelmed'' and ``stressed'' (P10). An error early on ``during the flight, <while it> needs action after the drone has landed,'' is also a distraction, as it is ``confusing for the operator or may prompt the <operator> to make a decision that's not necessary at the time'' (P9).\\

        \hline

        Distance to the robot & Lower than required levels when proximity is high  & ``But when <the drone> is going beyond the visual line of sight, <their SA is comparatively> low'' (P11), <given the challenges associated with> ``perceiving a 3D environment on a 2D display'' (P9).\\ 
        \cline{2-3}
        & Closer to required levels when proximity is low & ``When <human collaborators> can actually see the drone itself and it is quite close to <them>, <they are> well aware of what is happening'' (P9).\\
    \hline
\end{longtblr}
\end{center}

\clearpage

\begin{center}
\small
\begin{longtblr} [caption = {Examples of Robot Factors Influencing Actual Situational Awareness},
label = {tab:Robotic_Actual_SA} ]{|p{.1\linewidth}|p{.15\linewidth}|p{.66\linewidth}|}  \hline
        \textbf{Robot-related Factor} & \textbf{Trend in} \par \textbf{Actual SA} & \textbf{Example}\\
        \hline
        
        Information provided by robotic interface & Lower than the required SA when the interfaces lack information & `Minimalistic interfaces sometimes do not provide enough information of what is happening'' (P13) \\
        \cline{3-3}

        & & P14 received feedback from other users saying that there is some ``missing information'' (P14)\\
        \cline{3-3}
        & & Humans face difficulties in ``determining underlying terrain <on which robots act> based on video footage <and> point cloud data'' (P2), which made them arrive at incorrect assumptions and do inappropriate actions (P2, P3) or ``waste time to figure that out'' (P3).\\
        \cline{3-3}
        & & Failure to perceive ``a gesture <made by a crafter> requesting a handover'' (P8) \\
        \cline{3-3}
        & & Failure to comprehend that a branch ``is going to interfere with the apple picking process'', given the robot's camera position and orientation (P6) \\
        \cline{3-3}
        & & As the interface does not provide a view of ``<zooming> camera shaft in the workspace'', using extra cameras, surgeon cannot perceive information required to ``avoid potential collisions'' (P15).\\
        \cline{3-3}
        & & It was insufficient just to use ``a beep'' alert to indicate to the human that the robot was running out of battery and returning home; therefore, the initially used beep audio feedback was replaced by a ``voice recording'' that said ``low battery, going home'' to be ``more informative'' (P11).\\
        \cline{3-3}
        & & The large-scale art manufacturing application has not implemented ways to ``notify <the collaborator that> the robot is waiting for <some human action>'' or provide a ``WiFi camera for <the collaborator> to check <and perceive the robot's status when they are away>'' (P16).\\

        \cline{2-3}

        & Lower than the required SA when the interfaces provide unnecessary information & ``<Not all> the information that comes back to <them through the interface> is always necessary; in fact, it can be very noisy'' (P3), and visualising everything on the screen is at times overwhelming (P2, P5) ``to make sense of things'' (P2).\\
        \cline{3-3}
        
         & & Having ``a lot of button <controls>'' in their nuclear disaster response application, where ``some of them are not that necessary'' (P12)\\
         
        \cline{3-3}
         & & Showing the ``facial expression and pose status of the <crafter> on the interface'' as a duplicate of what the robot collaborator ``can directly perceive through the camera footage'' (P7, P8)\\
         
        \cline{3-3}
        & & Showing all structures of the environment, robots and system-generated topometric data in a map view at once makes it ``difficult to <perceive> what is going on'' (P5).\\
        
        \cline{2-3}
        
        & Closer to the required SA when the interfaces provide the necessary information & The interfaces that present ``only the very necessary things'' (P13) \\
        
        \cline{3-3}
        & & Providing ``a really simple pictograph-like'' representation of the map instead of ``a very detailed map'' when the use of the map is only to verify whether ``the robot is roughly in the place (P14)\\
        \hline

        Information formats in robotic interfaces & Lower than the required SA when the interfaces provide poor formats & Since the interface did not provide ``the ability to separate <3D visualisations> into individual layers'' (P2), operators found those presentations ``indecipherable'' (P2) and ``quite challenging to interpret'' (P1) and ``comprehend'' (P2).\\

        \cline{3-3}
        
        & & Operators experienced challenges due to ``mirror effect of cameras'' in their robotic interface (P12).\\
        \cline{3-3}
        
        & & Some errors communicated through interfaces ``confused'' the human collaborators (P10). This happened because of ``not telling the full story'' (P10) using ``informative'' language (P11), e.g., issuing a warning as ``disconnected <instead of saying that> the Wifi is disconnected'' (P10), and ``saying that collision avoidance is deactivated <while it> means that <the drone is> no longer using the <user-configured> collision avoidance parameters, but it still has its own collision avoidance'' (P11). Sometimes, human operators also have ``misinterpreted errors in the user interface because the wording <is quite technical and related to> inner workings <of the system>'' (P9).\\
        \cline{3-3}
        
        & & P12 shared their experience of using a controller interface that was ``way too complex'' because ``it was in German and <they> did not have the exact translation of everything''.\\
        \cline{3-3}

         & & The approach of ``fitting everything on the same screen'' makes it difficult to perceive certain information easily because it makes the panels within the screen ``pretty small'' (P8). \\
        \cline{3-3}
        
        & & In the ``early version of the interface, when there were a lot of apples <to visualise> on one screen, it was really difficult to say what was what'', and sometimes, the operator had to go through labeled details of every apple displayed on the screen to understand which apples could not be picked (P6).\\

        \cline{2-3}

        & Closer to the required SA when the interfaces provide efficient formats & The redesigned interface uses a ``sizing system, <where the> apples that are further away are smaller <and> apples that are closest are made more prominent'' (P6). This format to ``bring forward the most important information'' reduces the overload associated with gaining required SA.\\
        
    \hline
\end{longtblr}
\end{center}

\begin{table}[h!]
    \centering \small
    \setlength{\tabcolsep}{2pt}
    \begin{tabular}{|l|c|c|c|c|c|c|c|c|c|c|c|c|c|c|c|c|c|c|c|c|c|c|c|c|}
        \hline
    Application Context& \rotatebox{90}{Task Criticality \& Challenge} & \rotatebox{90}{Safety Criticality} & \rotatebox{90}{\# of Robots per Human} & \rotatebox{90}{Time Criticality} & \rotatebox{90}{Time Spent} & \rotatebox{90}{Capability of Autonomy} & \rotatebox{90}{Severity of Robot's Status} & \rotatebox{90}{Human Role} & & \rotatebox{90}{Expertise \& Contextual Understanding}  & \rotatebox{90}{Mental Models about Robots} & \rotatebox{90}{Trust in Automation} & \rotatebox{90}{Cognitive Capacity \& Ability to Multitask} & \rotatebox{90}{Willingness to Delegate} & \rotatebox{90}{Interruptions to Information} & \rotatebox{90}{Distractions} & \rotatebox{90}{Distance to Robot} & \rotatebox{90}{Information Provided by Interface} &  \rotatebox{90}{Information Formats} & \rotatebox{90}{SA Latency} & \rotatebox{90}{SA loss} & \rotatebox{90}{Inaccurate SA} & \rotatebox{90}{Incomplete SA} &  \rotatebox{90}{Excess SA}\\       
     \hline
        Physical Co-presence  & & & & & & & & & & & & & & & & & & & & & & & &\\
        \hspace{10pt} Remote  & 9 & 6& 4& 4& 1      & 9& 6& 5& 5    &7&4&5&4&3    &6&3&2  &7&6   &7&4&7&7 & 2\\
        \hspace{10pt} Co-located & 3 & 2& 2& 1& 0   & 2& 1& 2& 2    &2&2&1&1&0   &0&0&1  &3&0   &1&2&1&0 & 0\\
        \hspace{10pt} Distant  &6 & 3& 2& 0& 4     & 7& 5& 3& 3     &5&4&1&1&0   &4&3&2  &3&5   &5&3&5&5 & 0\\
        \hline

        Time Criticality  & & & & & & & & & & & & & & & & & & & & & & & &\\
        \hspace{10pt} Low & 11 & 8& 6& 5& 3        & 12& 9& 7& 5    & 9&8&6&3&2  &7&3&2  &9&7   &10&7&10&7 & 1 \\
        \hspace{10pt} High & 8 & 5& 4& 4& 1        & 7& 4& 2& 4     & 5&2&5&5&3  &4&1&1  &7&3   &5&4&5&4 & 1\\
        \hline

        Safety Criticality  & & & & & & & & & & & & & & & && & & & & & & & \\
        \hspace{10pt} Low & 10 & 7& 4& 4& 3       & 11& 8& 5& 4     &8&6&5&3&2   &7&3&3  &7&7   &9&6&9&7 & 1\\
        \hspace{10pt} High & 9 & 6& 6& 5& 1       & 8& 5& 4& 5      &6&4&6&5&3   &4&1&0  &9&3   &6&5&6&4 & 1\\
        \hline

        \# of Humans  & & & & & & & & & & & & & & & & & & & & & & & & \\
        \hspace{10pt} One & 14 & 9& 8& 5& 3        & 14& 10& 7& 8   &11&8&7&6&3  &8&4&3  &12&8  &11&8&11&9 & 2\\
        \hspace{10pt} Multiple & 2 & 1& 4& 0& 1  & 4& 3& 0& 3     & 1&1&2&3&1  &1&1&0  &3&2   &2&1&3&4 & 0\\
        \hline

        \# of Robots  & & & & & & & & & & & & & & & & & & & & & & & & \\
        \hspace{10pt} One & 14 & 9& 8& 5& 3        & 14& 10& 7& 8   &11&8&7&6&3  &8&4&3  &12&8  &11&8&11&9 & 2\\
        \hspace{10pt} Multiple & 5 & 4& 4& 4& 1  & 6& 4& 2& 3     &4&2&6&4&3   &4&1&0  &6&3   &5&3&5&4 & 2\\
        
        \hline
    \end{tabular}
    \caption{A detailed breakdown of number of participants who relate to each contextual (C), human (H) and robotic (R) factor affecting required (R.) and actual (A.) situational awareness and situational awareness inefficiencies (IN), considering their human-robot teaming application context. Note that participants were counted multiple times when their examples were applied to multiple applications. }
    \label{tab:generalisability}
\end{table}

\end{document}